\documentclass[lettersize,journal]{IEEEtran}
\usepackage{amsmath,amsfonts}
\usepackage{algorithmic}
\usepackage{algorithm}
\usepackage{array}
\usepackage{booktabs}
\usepackage{ragged2e}
\usepackage[caption=false,font=normalsize,labelfont=sf,textfont=sf]{subfig}
\usepackage{textcomp}
\usepackage{threeparttable}
\usepackage{multirow}
\usepackage{stfloats}
\usepackage{url}
\usepackage{verbatim}
\usepackage{graphicx}
\usepackage{cite}
\usepackage{xcolor}
\begin{document}
\onecolumn
\begingroup
\fontsize{20pt}{20pt}\selectfont
IEEE Copyright Notice\\
\endgroup
\\
© 2024 IEEE. Personal use of this material is permitted. Permission from IEEE must be obtained for all other uses, in any current or future media, including reprinting/republishing this material for advertising or promotional purposes, creating new collective works, for resale or redistribution to servers or lists, or reuse of any copyrighted component of this work in other works.
\twocolumn

\title{Model-Based Generation of Representative\\Rear-End Crash Scenarios Across the Full Severity Range Using Pre-Crash Data}

\author{Jian Wu, Carol Flannagan, Ulrich Sander, and Jonas Bärgman
\thanks{This work was supported by the FFI program sponsored by Vinnova, the Swedish governmental agency for innovation, as part of the project Improved quantitative driver behavior models and safety assessment methods for ADAS and AD (QUADRIS: nr. 2020-05156).}
\thanks{Jian Wu is with the Volvo Cars Safety Center, 41878 Gothenburg, Sweden, and the Department of Mechanics and Maritime Sciences, Chalmers University of Technology, 41756 Gothenburg, Sweden. (e-mail: jian.wu.2@volvocars.com)}
\thanks{Carol Flannagan is with the University of Michigan Transportation Research Institute (UMTRI), Ann Arbor, Michigan 48109, USA, and the Department of Mechanics and Maritime Sciences, Chalmers University, 41756 Gothenburg, Sweden. (email: cacf@umich.edu)}
\thanks{Ulrich Sander is with the Volvo Cars Safety Center, 41878 Gothenburg, Sweden. (e-mail: ulrich.sander@volvocars.com)}
\thanks{Jonas Bärgman is at the Department of Mechanics and Maritime Sciences, Chalmers University of Technology, 41756 Gothenburg, Sweden. (e-mail: jonas.bargman@chalmers.se)}
}



\maketitle

\begin{abstract}
Generating representative rear-end crash scenarios is crucial for safety assessments of Advanced Driver Assistance Systems (ADAS) and Automated Driving systems (ADS).
However, existing methods for scenario generation face challenges such as limited and biased in-depth crash data and difficulties in validation.
This study sought to overcome these challenges by combining naturalistic driving data and pre-crash kinematics data from rear-end crashes.
The combined dataset was weighted to create a representative dataset of rear-end crash characteristics across the full severity range in the United States.
Multivariate distribution models were built for the combined dataset, and a driver behavior model for the following vehicle was created by combining two existing models.
Simulations were conducted to generate a set of synthetic rear-end crash scenarios, which were then weighted to create a representative synthetic rear-end crash dataset.
Finally, the synthetic dataset was validated by comparing the distributions of parameters and the outcomes (Delta-v, the total change in vehicle velocity over the duration of the crash event) of the generated crashes with those in the original combined dataset.
The synthetic crash dataset can be used for the safety assessments of ADAS and ADS and as a benchmark when evaluating the representativeness of scenarios generated through other methods.
\end{abstract}


\section{Introduction} \label{section:intro}
\IEEEPARstart{A}{dvanced} Driver Assistance Systems (ADAS) and Automated Driving Systems (ADS) have the potential to improve traffic safety significantly \cite{pradhan2022impact}.
However, evaluating the safety performance of such systems is still a challenge.
Currently, virtual safety assessment is the primary procedure due to its low cost and high efficiency compared to conventional field tests \cite{feng2020safety, dona2022virtual, cai2022survey, szalay2023critical}.
Typically, in such a virtual assessment method, a comparison between the 'baseline' and 'treatment' is conducted to assess a technology.
The baseline is a set of scenarios to be analyzed without the technology under assessment, and these scenarios must match the assessment objective and include all relevant elements that may impact the performance of the technology under assessment \cite{wimmer2023harmonized}.
A large number of baseline scenarios is essential for making a statistically significant comparison between the baseline and treatment \cite{wimmer2023harmonized}.
There are two main approaches to creating a large number of baseline scenarios: traffic-simulation-based \cite{feng2021intelligent, baron2020repeatable, shah2018airsim} and in-depth-crash-data-based (referred to as IDC-based) \cite{scanlon2021waymo, bareiss2019crash, hamdane2015issues, haus2019potential, bargman2017counterfactual}.

The traffic-simulation-based approach aims to replicate daily driving activities to generate virtual crashes in a naturalistic driving environment \cite{feng2021intelligent, baron2020repeatable, shah2018airsim}.
Typically, this approach uses road-user behavior models created using naturalistic driving data (NDD) that contain a limited number of crashes, often of minor severity.
The simulations are carried out over an extended period, measured in millions of simulated driving hours.
Often, it is only the crash avoidance performance of the system that is assessed – by comparing the number of crashes generated through simulations in which the subject vehicle is equipped with the specific ADAS or ADS under assessment to the number of crashes from traffic simulations without the system \cite{ma2022verification}.

The IDC-based approach, on the other hand, uses detailed real-world crash information.
This information includes reconstructed or recorded data, such as the pre-crash kinematics of the involved road users.
Virtual crashes are generated by sampling from distributions of the parameterized pre-crash kinematics and/or other relevant crash characteristics.
The crashes generated serve as the baseline for assessing the safety performance of the ADAS or ADS.
Treatment simulations are then executed using the baseline crashes as a starting point but with the ADAS or ADS under assessment included.
The outcomes of the baseline and treatment simulations are then compared to assess (for example) the system's crash avoidance and injury mitigation performance \cite{scanlon2021waymo, bareiss2019crash, hamdane2015issues, haus2019potential, bargman2017counterfactual}.

Both approaches have their own set of challenges.
As noted, the traffic-simulation-based approach takes extensive time to simulate \cite{feng2021intelligent}.
Also, using NDD as the initial condition for generating crash scenarios may lead to crash characteristics that are significantly different than those in real-world crashes \cite{olleja2022can}.
For example, few higher-severity crashes are typically generated, biasing the assessment towards a baseline/treatment comparison of low-severity crashes (or even to crash surrogates \cite{arun2021systematic}).
In addition, crashes generated by the traffic-simulation-based approach rely heavily on multiple accurate models of road-user behaviors that can produce realistic crashes, representing the real world.
However, there is typically a lack of proof of similarity (i.e., validation) between the generated and real-world crashes regarding the characteristics of individual crashes and the characteristics' distributions.

In contrast, the IDC-based approach requires substantial in-depth pre-crash kinematics data.
This information is seldom available for most types of scenarios—and when it is, it is typically biased towards severe crashes due to the selection criteria of conventional crash databases.
As a result, relying solely on these databases to create synthetic crashes skews the crash generation models, potentially distorting the overall analysis \cite{leledakis2021method, gambi2019generating, wang2022autonomous}.

This study aims to address these challenges by combining both approaches, creating a dataset of synthetic, passenger-vehicle-involved, rear-end crash scenarios that are representative of the population of such crashes with respect to severity in the United States (referred to as the 'reference dataset' with notation $\Tilde{\Phi}$).
The dataset is intended for use in the safety assessment of ADAS and ADS and as a benchmark for evaluating the representativeness of scenarios generated through other scenario generation methods.

A synthetic rear-end crash scenario consists of three main components: a speed profile of the lead vehicle, a behavior model of the following vehicle (how it responds to the behavior of the lead vehicle), and the initial states of the scenario.
We used the speed profile of the lead vehicle, which is a vehicle kinematics model (instead of a behavior model) because the lead vehicle's behavior is mostly independent of the following vehicle's behavior \cite{wu2024modeling}.
The initial states include the speeds of both vehicles and the following distance at the beginning of the scenario.

For the first component, we turned to our previous study \cite{wu2024modeling} in which we modeled the pre-crash lead-vehicle kinematics in rear-end crash scenarios and produced a synthetic dataset of lead-vehicle speed profiles representative of crashes across all severity levels.

The second component, a following-vehicle behavior model, was created by merging two existing driver behavior models \cite{derbel2013modified, svard2021computational}.
For the third component, data from multiple rear-end crash datasets from various sources were combined and weighted to create a reference dataset of the initial states of rear-end crash scenarios and minimum accelerations of both vehicles.
Distributions were then built for this reference dataset.

Once the three components were complete, simulations were conducted to obtain a set of synthetic rear-end crash scenarios.
The scenarios were weighted to match the obtained reference datasets, creating a representative synthetic rear-end crash dataset.
Finally, this dataset was validated by comparing the parameter distributions of the generated crashes with the reference datasets, as well as by comparing the lead-vehicle Delta-v (i.e., the total change in vehicle velocity over the duration of the crash event) distributions of the two.

\section{Data} \label{section:datasets}
\subsection{Data Sources}
The datasets used are from four sources: the Crash Investigation Sampling System (CISS), the Second Strategic Highway Research Program (SHRP2) Naturalistic Driving Study (NDS), the German In-Depth Accident Study (GIDAS) Pre-Crash Matrix (PCM), and our prior study \cite{wu2024modeling}.

CISS is a nationally representative sample of crashes in the United States in which at least one light vehicle was towed away from the scene \cite{zhang2019crash, subramanian2020crash}. The data were obtained from comprehensive crash investigations, encompassing examinations of damaged vehicles and crash sites as well as assessments of crash kinematics.
CISS includes Event Data Recorder (EDR) data whenever available.

The SHRP2 NDS provides recorded pre-crash information originating from the United States.
Over 3,300 passenger vehicles were equipped with a data acquisition system (DAS) to capture four distinct video perspectives, coupled with information extracted from vehicle networks and sensors.
Between 2010 and 2013, naturalistic driving data were collected from participants' instrumented vehicles in six locations across the United States.
The SHRP2 dataset includes a continuum of conflicts, from near-crashes to (a few) high-severity crashes.
The incidents were identified by applying a set of event identification algorithms to the accumulated trip records.
In a subsequent manual annotation step, the identified instances were classified by severity level \cite{hankey2016description}.

The GIDAS dataset is a renowned German dataset that comprehensively investigates traffic accidents with personal injury in Germany.
The PCM subset of the GIDAS dataset contains reconstructed pre-crash time-series data that describes the pre-crash trajectories and provides digitized information about the road layout and potential sight obstructions.
Since its inception in 2011, the PCM dataset has provided time-series data of the phases leading up to a diverse array of crash scenarios, encompassing a temporal span of up to five seconds before the events \cite{schubert2017gidas}.
Note, however, that the reconstructions of the pre-crash phase are based on evidence from the accident site and eyewitnesses; detailed driver behavior is unknown.

Among the three datasets, the SHRP2 dataset (with a frequency of 10 Hz) covers incidents from near-crashes to severe crashes, while the GIDAS-PCM dataset (with a frequency of 100 Hz) only includes crashes resulting in personal injury, and the CISS dataset (with a frequency varying from 1 to 10 Hz) only contains accidents involving towed vehicles.
Consequently, the latter two datasets exclude low-severity crashes, and the censoring boundary is not obvious to quantify.

\subsection{Datasets}
\begin{table*}[!t]
\centering
\caption{Datasets in the study\label{tab:datasets}}
\begin{threeparttable}
\begin{tabular}{cccccc}
\hline
Type & Notation$^{*}$ & Source & Description & Signals & Sample size\\
\hline
\multirow{7}{*}{Input} & CISS\_m & CISS & General rear-end crash data & $m_{f}, m_{l}$ & 748\\
& CISS\_f & CISS & Rear-end pre-crash data (SV-striking) & $v_{f}, \Delta v_{l}$ & 408\\
& SHRP2\_f & SHRP2 & Rear-end pre-crash data (SV-striking) & $v_{f}, \Delta v_{f}$ & 116\\
& SHRP2\_b & SHRP2 & Further annotated rear-end pre-crash data & $v_{f}, v_{l}, d, \Delta v_{f}$ & 37\\
& PCM\_b & GIDAS-PCM & Reconstructed rear-end crash data & $v_{f}, v_{l}, d, \Delta v_{l}$ & 861\\
& REF\_l & Previous study & Combined rear-end crash lead-vehicle speed profile & $v_{c}, a_1, a_2, \tau_s, \tau_1, \tau_2, \Delta v_{l}$ & 132\\
& REF\_sl & Previous study & Synthetic lead-vehicle speed profiles & $v_{c}, a_1, a_2, \tau_s, \tau_1, \tau_2$ & 10,000\\
\hline
\multirow{4}{*}{Intermediate} & COM\_f & \multirow{4}{*}{-} & Combined following-vehicle-information dataset & \multirow{4}{*}{-} & 524\\
& COM\_b & & Combined both-vehicle-information dataset & & 913\\
& REF\_f & & Reference following-vehicle-information dataset & & 324\\
& REF\_i & & Reference intermediate dataset & & 10,000\\
\hline
\multirow{3}{*}{Output} & REF\_b & \multirow{3}{*}{-} & General rear-end crash data & \multirow{3}{*}{-} & 852\\
& REF\_sb &  & Synthetic both-vehicle-information dataset & & 10,000\\
& REF\_ss &  & Synthetic rear-end crash dataset & & 5,000\\
\hline
\end{tabular}
\begin{tablenotes}
\RaggedRight
\item $^{*}$ Each dataset is named according to the source type (i.e., REF: reference, COM: combined), with a postfix indicating the type of information that the dataset includes (m: mass information, l: lead-vehicle information, f: following-vehicle information, b: both-vehicle information, i: intermediate, sl: synthetic lead-vehicle information, sb: synthetic both-vehicle information, and ss: synthetic crash scenario).
\end{tablenotes}
\end{threeparttable}
\end{table*}

Table \ref{tab:datasets} shows the datasets used in this study.
Seven datasets derived from the four sources were used as input.
Four intermediate datasets were created from these to obtain three output datasets.
Datasets other than those used as input are introduced later (in Section \ref{section:methodology}).
Each dataset is named according to the source type, with a postfix indicating the kind of information included.
(See Table \ref{tab:datasets} for details).

CISS\_m (n = 748) includes all CISS rear-end crash records, which contain the curb weight of the lead and following vehicle, $m_{f}$ and $m_{l}$ (kg).
CISS\_f (n = 408) comprises EDR-based pre-crash data from CISS rear-end crashes in which the subject vehicle (SV) was the striking/following vehicle.
Signals extracted from the CISS\_f dataset were the following-vehicle speed $v_{f}$ (m/s) and Delta-v of the lead vehicle $\Delta v_{l}$ (m/s).
SHRP2\_f (n = 116) consists of SHRP2 rear-end pre-crash events recorded by the striking vehicles.
The signal $v_{f}$ and its derivative signal, the following vehicle's Delta-v $\Delta v_{f}$ (m/s), were the signals used in the dataset. 
SHRP2\_b (n = 37) was generated by further annotating a subset of the SHRP2 rear-end crashes included in SHRP2\_f. 
For each crash, the following distance $d$ (m) was estimated by Victor et al. \cite{victor2015analysis} using image processing, and the lead-vehicle speed $v_{l}$ (m/s) was deduced based on $v_{f}$ and $d$.
PCM\_b (n = 861) contains reconstructed rear-end crash data from the GIDAS-PCM dataset, including $v_{f}$, $v_{l}$, $d$, and $\Delta v_{l}$ signals.
(Note that we excluded one pedal misapplication case in which the driver of the following vehicle accidentally pushed the acceleration pedal when approaching the lead vehicle.)

\begin{figure}[!t]
    \centering
    \includegraphics[width=0.35\textwidth]{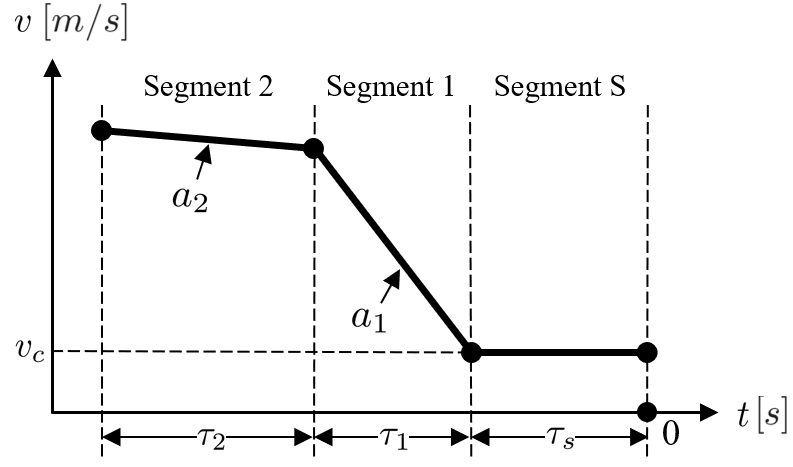}
    \caption{Three selected segments of the lead-vehicle speed profile in a rear-end crash.}
    \label{fig:segments}
\end{figure}

In our previous study \cite{wu2024modeling}, we fitted the lead-vehicle speed profiles five seconds before the impact in recorded rear-end crashes from the SHRP2 and CISS datasets into a piecewise linear model. 
This model simplifies a speed profile as a sequence of, at most, three straight lines; the slopes of the lines are the fitted accelerations.
(Note that there can be cases with fewer than three segments.)
Each lead-vehicle speed profile was parameterized as a six-dimensional vector: $[v_c, a_1, a_2, \tau_s, \tau_1, \tau_2]$. 
Fig. \ref{fig:segments} shows an example with three segments (going backward in time from time zero, the impact moment):
\begin{itemize}
    \item Segment S: The lead vehicle maintains a steady speed in this segment. $\tau_s$ is the segment duration, and $v_c$ is the lead vehicle's estimated speed at time zero.
    \item Segment 1: The lead vehicle keeps a non-zero constant acceleration in this segment. $\tau_1$ is the segment duration, and $a_1$ is the fitted constant acceleration.
    \item Segment 2: The lead vehicle keeps a constant acceleration in this segment. $\tau_2$ is the segment duration, and $a_2$ is the fitted constant acceleration.
\end{itemize}
This piecewise linear model will be reintroduced in Section \ref{section:datacombination}.

REF\_l and REF\_sl were also derived from our previous study \cite{wu2024modeling}. 
REF\_l (n = 132) contains 132 rear-end crash lead-vehicle speed profiles, of which 83 and 49 come from the SHRP2 and CISS datasets, respectively.
The samples in the dataset were weighted to create a reference dataset of the lead-vehicle kinematics covering the full range of crash severity (from non-severe to severe).
In this study, the lead vehicle's Delta-v ($\Delta v_l$) was obtained by either extracting the signal (for CISS crashes) or estimating it as the difference between the post- and pre-impact lead-vehicle speed (for SHRP2 crashes). 
REF\_sl (n = 10,000) is a dataset of 10,000 synthetic lead-vehicle speed profiles generated by the distribution models built on REF\_l. 
The two datasets can be considered reference datasets of lead-vehicle kinematics in rear-end crashes.

\subsection{Event Data Extraction}

CISS pre-crash data typically include five seconds before the impact, while SHRP2 and GIDAS-PCM data cover a longer duration.
The start time of all events was thus set to five seconds before the impact (defined as time zero) to make all events equivalent.

For each crash event in CISS\_f, only the following vehicle's initial speed (i.e., the speed at $t = -5$ s) was extracted.
In contrast, we extracted the whole events (all the time-series data) for SHRP2- and GIDAS-PCM-sourced datasets.
As in the previous study \cite{wu2024modeling}, the extracted event duration for datasets SHRP2\_f, SHRP2\_b, and PCM\_b spanned from -5 to -0.3 s, ending just before impact to avoid a possible sharp acceleration pulse.

\section{Methodology} \label{section:methodology}
\begin{figure}[!t]
    \centering
    \includegraphics[width=0.35\textwidth]{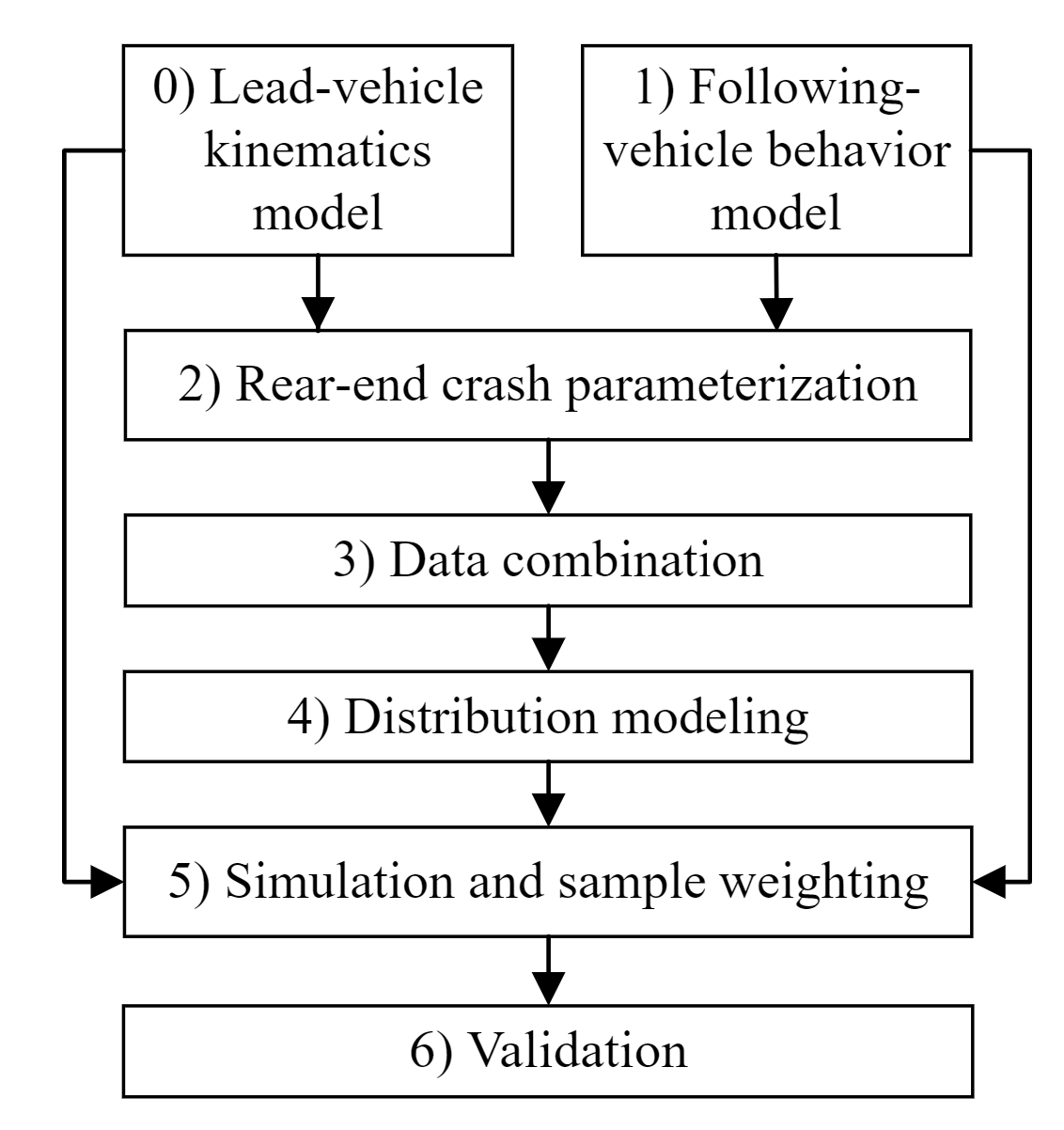}
    \caption{Flowchart of the methodology. Step 0 (the lead-vehicle kinematics model) was performed in our previous study \cite{wu2024modeling}.}
    \label{fig:methodology}
\end{figure}

In this study, the following six steps were performed in order:
\begin{enumerate}
    \item Following-vehicle behavior model creation
    \item Rear-end crash parameterization
    \item Data combination
    \item Distribution modeling
    \item Simulation and sample weighting
    \item Validation
\end{enumerate}
Fig. \ref{fig:methodology} shows how these steps are interconnected.
Step 0 (the lead-vehicle kinematics model) was performed in our previous study \cite{wu2024modeling}.
Steps 0 and 1 (the following-vehicle behavior model) provided the basis for Step 2, which simplified a rear-end crash event by representing it as a set of parameters.
Then, Step 3 created a reference dataset of a subset of parameters (initial states of rear-end crash scenarios and minimum accelerations of both vehicles), and Step 4 built a distribution model on the reference dataset.
The simulation and sample weighting step (Step 5) was carried out using the distribution model and the following-vehicle behavior model, as well as the lead-vehicle kinematics model from the previous study \cite{wu2024modeling}.
The step resulted in the synthetic rear-end crash dataset.
Finally, we validated the synthetic dataset in Step 6.

\subsection{Step 1: Following-vehicle Behavior Model Creation} \label{section:followingvehmodel}
The following-vehicle behavior model is a combination of two existing driver behavior models: 1) the modified intelligent driver model \cite{derbel2013modified}, which describes the longitudinal vehicle control behavior during car following, and 2) the driver pre-crash brake response model \cite{svard2021computational}, which predicts when and how the driver brakes in the rear-end pre-crash phase.
Additionally, the model includes the possibility of generating abnormal driver acceleration behavior under some conditions, which is introduced later in this sub-section.

\subsubsection{Modified intelligent driver model}
The modified intelligent driver model is a modification of the Intelligent Driver Model (IDM), a time-continuous car-following model frequently used in traffic flow modeling.
In this model, only longitudinal movement is considered. The acceleration of vehicle $\alpha$, $a_\alpha$ (m/s$^2$), is computed as
\begin{equation}
    a_\alpha = a \cdot [1 - (\frac{v_\alpha}{v_0})^4 - (\frac{d_\alpha^*}{d_\alpha})^2],
\end{equation}
where $v_\alpha$ ($m/s$) is the speed of vehicle $\alpha$, $a$ (m/s$^2$) is the maximum acceleration, $v_0$ (m/s) is the desired speed of vehicle $\alpha$ in free traffic, $d_\alpha$ (m) is the following distance, and $d_\alpha^*$ (m) is the desired minimum following distance, which is defined as
\begin{equation}
    d_\alpha^* = d^*(v_\alpha,\Delta v_\alpha) = d_0 + v_{\alpha}T + c \frac{v_\alpha^2}{b} - \frac{v_\alpha\Delta v_\alpha}{2\sqrt{ab}},
\end{equation}
where $\Delta v_\alpha$ (m/s) is the relative speed of the lead vehicle of vehicle $\alpha$, $d_0$ (m) is the jam distance, $T$ (s) is the minimum time headway to the vehicle in front, $b$ (m/s$^2$) is the comfortable braking deceleration, and $c$ is the coefficient added in the modified model to increase the desired minimum following distance.
$v_0$ was set as the road speed limit.
As suggested by Derbel et al. \cite{derbel2013modified}, in this work $a$, $b$, and $c$ were set to 3 m/$s^2$, 4 m/s$^2$, and 0.4, respectively, and $T \sim \mathcal{N}(1.5, 0.16)$ s.

\subsubsection{Driver pre-crash brake response model}
This model, proposed by Svärd et al. \cite{svard2017quantitative, svard2021computational} (denoted as model $BWL_{rc}$ in their paper), is a driver model that quantitatively predicts how and when the driver will initiate and modulate the pre-crash brake response.
The model uses the accumulation of the prediction error of looming (the relative expansion rate of the lead vehicle's image on the retina of the following vehicle \cite{lee1976theory}) as the basis for the driver’s braking response.
In addition, the model considers the driver's off-road glance behavior.
Specifically, the model applies an off-road glance looming weight parameter to account for the driver's partial perception of looming during off-road glances.
The driver's brake responses can thus occur quickly since the driver accumulates evidence even when not looking directly at the road.
The model parameters in this study were set the same as the fitting results calibrated for the 13 SHRP2 rear-end pre-crash events \cite{svard2021computational}.
The model's inputs are looming, the glance off-road signal, and the minimum acceleration (or maximum deceleration) of the following vehicle $a_{f,min}$ (m/s$^2$).
The model outputs a non-positive acceleration $a_b$ (m/s$^2$).

Research \cite{victor2015analysis, bargman2015does, markkula2016farewell} has shown that the role of distraction in rear-end crashes is influenced by situational urgency.
This influence is operationalized by emphasizing off-road glances after the time to collision ($TTC$) falls below a certain threshold instead of focusing on off-road glances throughout the event.
Therefore, to describe the glance-off-road behavior of the following vehicle’s driver, we used the parameter suggested in the reference study \cite{bargman2015does}: the glance off-road overshoot $t_g$ (s) after $TTC^{-1} = 0.2$ s$^{-1}$.
The overshoot is the off-road glance that occurs after $TTC^{-1} = 0.2$ s$^{-1}$ (hereafter called the \textit{anchor point}) and continues for a duration of $t_g$ seconds. $TTC^{-1}$ is the inverse time to collision (s$^{-1}$).
The same process in the reference study was followed to create the reference distribution for $t_g$, using glance behavior for normal driving from the SHRP2 dataset.

\subsubsection{Abnormal acceleration behavior} \label{section:abnormalaccel}
In the PCM\_b dataset, there are crashes in which both the lead and following vehicles were initially stationary.
Then, after a while, the following vehicle started accelerating until it hit the lead vehicle.
Unlike the excluded case of pedal misapplication, the driver of the following vehicle seemed to ignore the lead vehicle completely in these cases, possibly due to distraction.

Two parameters, $a_a$ and $t_a$, were added to the following-vehicle behavior model to account for these 'abnormal' acceleration behaviors.
When the behaviors occurred, the following vehicle was assumed to keep a constant acceleration $a_a$, which was set to 1.8 m/s$^2$ (the mean of the acceleration values in the abnormal acceleration cases in the PCM\_b dataset).
$t_a$ is the time duration from the event's start ($t = -5$ s) to the beginning of the abnormal behavior of the following vehicle's driver (s). 
If $t_a$ is equal to or greater than five seconds, then no abnormal acceleration behavior is present (i.e., the following vehicle does not have time to initiate acceleration before the crash).
However, if $t_a$ is less than five seconds, then the lead vehicle will not be taken into account (i.e., the model will act as if there is no lead vehicle) in the calculation of $a_{i,a}$ after $t_a$ from the event's start (i.e., $t = t_a - 5$).
The reference distribution of $t_a$ was obtained by fitting the data of $t_a$ in cases with abnormal acceleration behaviors into a normal distribution (see Section \ref{section:categorization} for further information regarding the fitting process).

\subsubsection{Combined following-vehicle behavior model}
We combined the modified IDM, the driver pre-crash brake response model, and the two abnormal acceleration behavior parameters to create a combined following-vehicle behavior model (referred to as 'the following-vehicle behavior model').
In the combined model, when abnormal acceleration behavior occurs, the output acceleration is the acceleration of the modified IDM, ignoring the lead vehicle.
Otherwise, the modified IDM only describes the driver's acceleration behavior, and the brake response model describes the braking behavior.
Hence, the acceleration of vehicle $i$ ($a_i$, m/s$^2$) is computed as
\begin{equation}
    a_i = \begin{cases}
        0,& \text{if}\ t < t_a - 5 < 0\\
        a_{a},& \text{if}\ t_a < 5\ \&\ t > t_a - 5\\
        {\text{max}(a_{i,a}, 0),}& \text{if}\ {a_{i,b} = 0\ \&\ t_a \geq 5}\\
        a_{i,b},&{\text{otherwise.}}
    \end{cases},
\end{equation}
where $a_{i,b}$ is the output acceleration of the driver pre-crash brake response model, $a_{i,a}$ is the output acceleration of the modified IDM, and $a_{a}$ is the constant acceleration of the following vehicle when the abnormal acceleration behavior occurs.
In summary, four parameters are used in the following-vehicle behavior model: $a_{f,min}$, $t_g$, $T$, and $t_a$.

\subsection{Step 2: Rear-end Crash Parameterization}
\begin{table*}[!t]
\centering
\caption{Definitions of the twelve parameters\label{tab:parameters}}
\begin{threeparttable}
\begin{tabular}{p{0.08\textwidth}p{0.5\textwidth}p{0.05\textwidth}p{0.2\textwidth}}
\hline
Parameter & Definition & Unit & Type\\
\hline
$d_{init}$ & Following distance at the initial$^*$ & m & Both-vehicle-related\\
$v_{f,init}$ & Following vehicle's speed at the initial & m/s & Following-vehicle-related\\
$a_{f,min}$ & Minimum fitted following-vehicle acceleration & m/s$^2$ & Following-vehicle-related\\
$T$ & Minimum time headway to the lead vehicle & s & Following-vehicle related\\
$t_g$ & Glance off-road overshoot of the following vehicle's driver & s & Following-vehicle-related\\
$t_a$ & Duration from the event's start to the beginning of the abnormal behavior of the following vehicle's driver & s & Following-vehicle-related\\
${v_{l,init}}^{**}$ & Lead vehicle's speed at the initial & m/s & Lead-vehicle-related\\
${a_1}$ & Fitted acceleration of Segment 1 of the lead-vehicle speed profile & m/s$^2$ & Lead-vehicle-related\\
${a_2}$ & Fitted acceleration of Segment 2 of the lead-vehicle speed profile & m/s$^2$ & Lead-vehicle-related\\
${\tau_s}$ & Duration of Segment S of the lead-vehicle speed profile & s & Lead-vehicle-related\\
${\tau_1}$ & Duration of Segment 1 of the lead-vehicle speed profile & s & Lead-vehicle-related\\
${\tau_2}$ & Duration of Segment 2 of the lead-vehicle speed profile & s & Lead-vehicle-related\\
\hline
\end{tabular}
\begin{tablenotes}
\RaggedRight
\item $^*$ The initial moment refers to $t = -5$ s for extracted crash events.
\item $^{**}$ $v_{l,init} = v_c - a_1 \tau_1 - a_2 \tau_2$.
\end{tablenotes}
\end{threeparttable}
\end{table*}

In this step, we parameterized a rear-end crash as a twelve-dimensional vector which considers the initial states of rear-end crash scenarios, the following-vehicle behavior model, and the parameterized lead-vehicle speed profile (see Table \ref{tab:parameters}).
The parameters can be divided into three types: 1) both-vehicle-related ($d_{init}$), 2) following-vehicle-related ($v_{f,init}, a_{f,min}, T, t_g, t_a,$), and 3) lead-vehicle-related ($v_{l,init}, a_1, a_2, \tau_s, \tau_1, \tau_2$).
The rationale is that, for instance, $a_{f,min}$ and $t_g$ affect the following vehicle's braking behavior, while $T$ and $t_a$ affect the following vehicle's acceleration behavior.

\subsection{Step 3: Data Combination} \label{section:datacombination}

It would be ideal to create a reference dataset for the parameterized rear-end crashes directly from a single source of crash data.
However, none of the available datasets alone could serve as a reference dataset since they all have limitations.
For instance, only SHRP2\_b and PCM\_b contain information about both vehicles, allowing for joint distribution of the twelve parameters.
The other datasets either contain information about only one vehicle's pre-crash dynamics (e.g., SHRP2\_f) or general crash data without detailed vehicle dynamics (CISS\_m). 
SHRP2\_b, for example, has a relatively small sample size, and the quality of the lead-vehicle speed signal in the dataset is limited since it was deduced by the following distance, which was estimated using image processing on relatively low-quality video. 
Further, as previously mentioned, PCM\_b is biased towards severe crashes.
Also, in the reconstruction process for the cases in PCM\_b, it is usually assumed that the lead vehicle was moving with constant acceleration or deceleration before the crash when the evidence for a detailed speed profile is lacking.
However, our previous study \cite{wu2024modeling} has shown that lead-vehicle speed profiles can take more forms in the pre-crash phase.
Since both SHRP2\_b and PCM\_b lack a high-quality lead-vehicle speed signal, the initial speed and minimum acceleration of the lead vehicle ($v_{l,init}$ and $a_{l,min}$) were selected to represent the lead-vehicle kinematics for the two datasets.

\begin{figure*}[!t]
    \centering
    \includegraphics[width=0.9\textwidth]{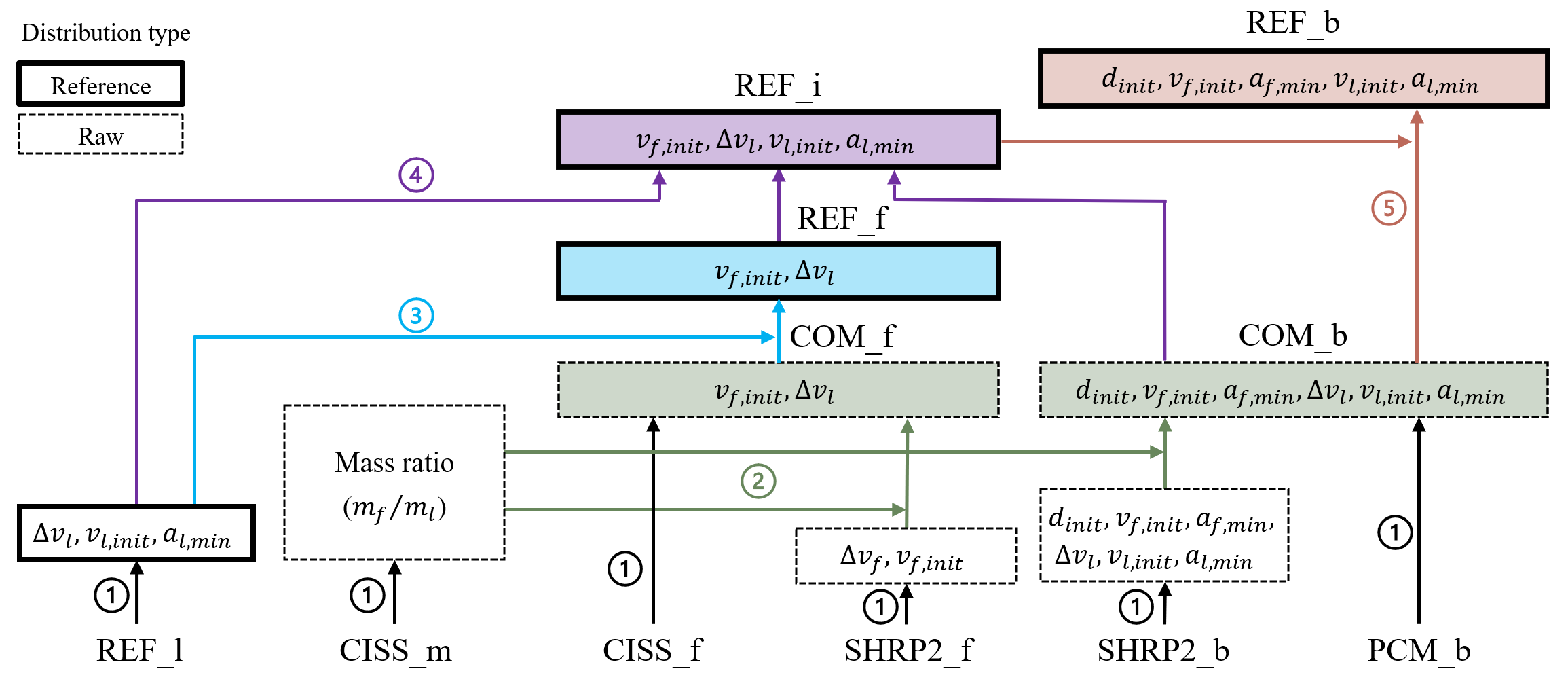}
    \caption{Flowchart of the data combination.
    Reference and raw datasets are marked by solid and dashed lines, respectively.
    Colored arrows and numbers indicate sub-steps.
    The outcome is the dataset REF\_b, the reference dataset of $d_{init}$, $v_{f,init}$, $a_{f,min}$, $v_{l,init}$, and $a_{l,min}$.}
    \label{fig:datacombination}
\end{figure*}

Consequently, it is not feasible to create a reference dataset of all twelve parameters directly from the available datasets.
We then created several reference datasets of subsets of parameters as an intermediate step in building the final reference database.
As mentioned in Section \ref{section:intro}, a synthetic rear-end crash scenario consists of three main components: the lead-vehicle speed profile, the following-vehicle behavior model, and the initial states of the scenario. 
The synthetic lead-vehicle speed profile dataset (REF\_sl) from our previous study \cite{wu2024modeling} serves as a reference dataset of lead-vehicle-related parameters.
Therefore, the challenge is to create another reference dataset containing the remaining parameters, which can be linked to REF\_sl so that we can combine them to create synthetic crash scenarios.
The final dataset, REF\_b, is a reference dataset of the initial states of the crash scenarios and minimum accelerations of both vehicles: $\Tilde{\Phi}(d_{init}, v_{f,init}, a_{f,min}, v_{l,init}, a_{l,min})$.
The parameters $d_{init}$, $v_{f,init}$, and $v_{l,init}$ are used for setting the initial states of the overall scenario; $a_{f,min}$ is required for the following-vehicle model. 
The parameters $v_{l,init}$ and $a_{l,min}$ are common to both REF\_b and REF\_sl, so they can be used to link any data point in the distribution to a set of synthetic lead-vehicle speed profiles in REF\_sl.

Fig. \ref{fig:datacombination} shows the data combination process applied to obtain REF\_b; the five sub-steps are numbered and color-coded.
Sub-step 1 extracts relevant signals from each dataset. 
Sub-step 2 deduces the Delta-v of the lead vehicle and combines the following-vehicle-related and both-vehicle-related datasets, respectively.
To finally obtain REF\_b, sub-steps 3-5 apply sample weight adjustments to reduce biases in the datasets.

\textbf{\textit{Sub-step 1 (Extract signals):}} Relevant signals from each dataset were extracted: specifically, the speed profiles of the lead and following vehicles in SHRP2\_b and PCM\_b were fitted into the six-parameter piecewise linear model and simplified as (at most) three consecutive straight lines (see Fig. \ref{fig:segments}).
The minimum fitted acceleration for all segments was also extracted for each speed profile.
Finally, the following parameters were extracted for each crash event in SHRP2\_b and PCM\_b: $d_{init}$, $v_{f,init}$, $a_{f,min}$, $v_{l,init}$, and $a_{l,min}$.
The lead vehicle's minimum fitted acceleration ($a_{l,min}$) was also extracted in REF\_l.

\textbf{\textit{Sub-step 2 (Deduce Delta-v):}} $\Delta v_{l}$ is used as the indicator of crash severity.
REF\_l contains the reference distribution of $\Delta v_{l}$, $\Tilde{\Phi}(\Delta v_l)$, which was further used to mitigate the severity level bias in other raw datasets (such as CISS\_f and SHRP2\_f) in later sub-steps.
However, SHRP2\_f and SHRP2\_b can only provide Delta-v of the following vehicle, $\Delta v_{f}$.
Therefore, in this sub-step, the mass ratio ($m_f/m_l$) data extracted from CISS\_m was fitted into a generalized gamma distribution, which was then used to transform the $\Delta v_{f}$ signal in the two datasets (SHRP2\_f and SHRP2\_b) to $\Delta v_{l}$ based on conservation of momentum:
\begin{equation}
    \Delta v_l = -\frac{m_f}{m_l} \Delta v_f.
\end{equation}
After obtaining $\Delta v_l$ for SHRP2\_f and SHRP2\_b, we combined CISS\_f and SHRP2\_f into one dataset (referred to as COM\_f).
We also combined SHRP2\_b and PCM\_b into another dataset (COM\_b).

\textbf{\textit{Sub-step 3 (Obtain REF\_f):}} The dataset COM\_f is biased towards severe crashes because some of the data come from CISS.
Thus, this sub-step developed case weights that adjust the COM\_f representation to match the reference distribution of $\Delta v_l$ from REF\_l using the \textit{k-nearest neighbors (KNN) sample weighting method} to assign weights to the samples in COM\_f.
(See Appendix \ref{appendix:KNN} for further information regarding this method.)
The new weighted dataset is called REF\_f.

\begin{figure}[!t]
    \centering
    \includegraphics[width=0.35\textwidth]{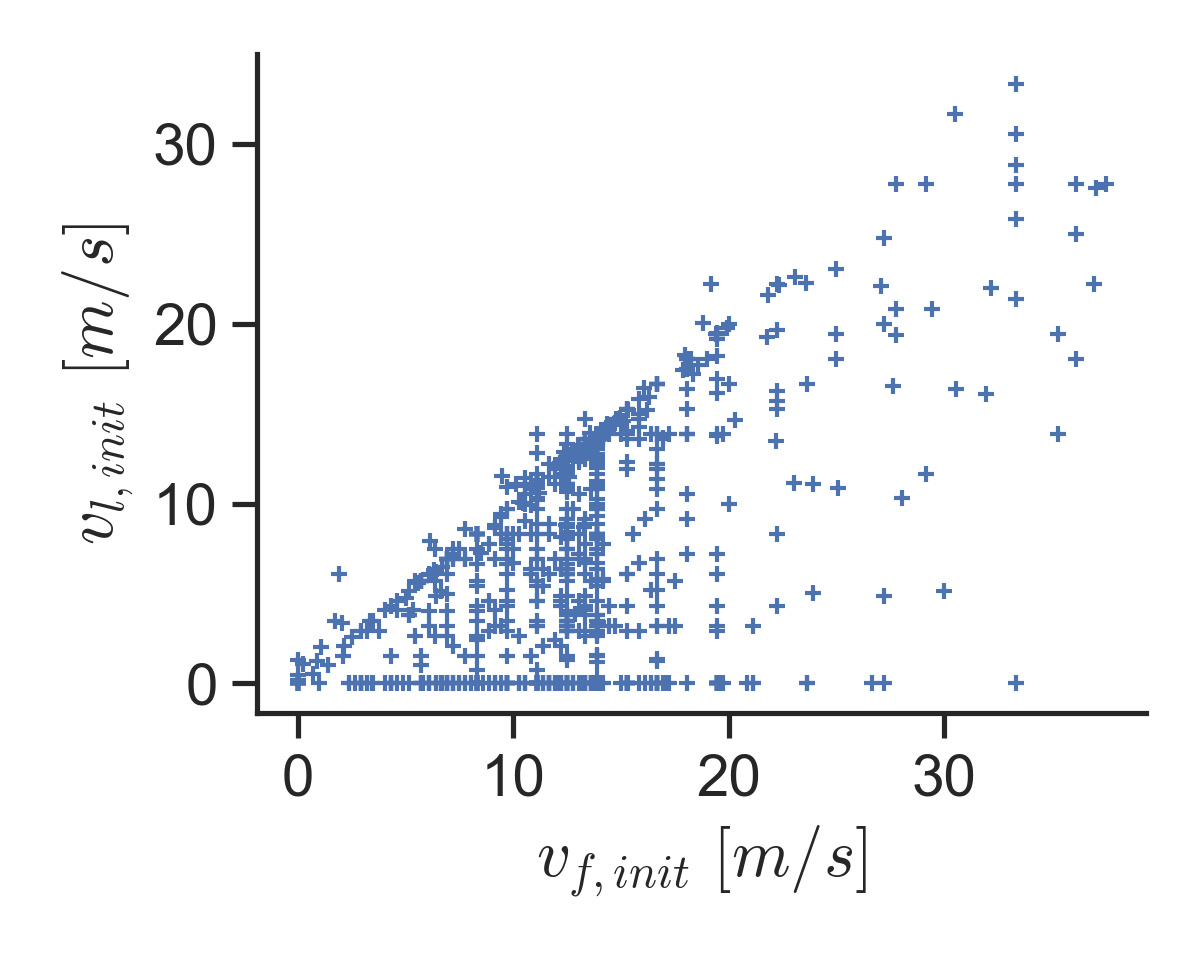}
    \caption{Scatter plot of $v_{f,init}$ and $v_{l,init}$ for COM\_b. In most cases, $v_{l,init}$ is no larger than $v_{f,init}$.}
    \label{fig:scatterplot}
\end{figure}

\textit{\textbf{Sub-step 4 (Obtain REF\_i):}} This sub-step created an intermediate reference dataset (REF\_i, as shown in Fig. \ref{fig:datacombination}) with as many parameters as possible, which was used for weighting COM\_b to reduce bias in those parameters.
REF\_i was created by combining two reference datasets: REF\_f, obtained in the previous sub-step, and the reference dataset of lead-vehicle kinematics (REF\_l).
The resulting dataset contains the Delta-v and minimum fitted acceleration of the lead vehicle, as well as the initial speeds of both vehicles ($\Delta v_l$, $a_{l,min}$, $v_{f,init}$, and $v_{l,init}$).

To obtain REF\_i, we randomly selected (with replacement) an equal number of samples from both reference datasets (REF\_f and REF\_l).
Then, we employed a \textit{pairing algorithm} to pair the selected samples from the two datasets one by one so that the pairs would preserve the correlation structure among the parameters ($\Delta v_l$, $a_{l,min}$, $v_{f,init}$, and $v_{l,init}$).

To design the pairing algorithm, we first investigated the correlation structure among the parameters.
We computed the correlations among the three parameters in REF\_l: the lead vehicle's Delta-v, initial speed, and minimum fitted acceleration ($\Delta v_l$, $v_{l,init}$, and $a_{l,min}$).
The results show that the initial speed of the lead vehicle ($v_{l,init}$) is highly correlated with the minimum fitted acceleration of the lead vehicle ($a_{l,min}$).
However, $\Delta v_l$ is only weakly correlated with the other two parameters.
(A Pearson correlation coefficient \cite{cohen2009pearson} with an absolute value smaller than 0.3 indicates a weak correlation; see Section \ref{section:results} for further information regarding the correlation assessment.)
In this study, for the sake of simplicity, we ignored the weak correlation and considered $\Delta v_l$ to be independent of the other two parameters ($v_{l,init}$ and $a_{l,min}$).
Since REF\_f also contains $\Delta v_l$, we can simply sample $v_{l,init}$ and $a_{l,min}$ (instead of all three parameters) from REF\_l.
Then, the pairing algorithm must pair those samples of the two parameters with samples from REF\_f (i.e., the reference dataset of the following vehicle's initial speed and Delta-v of the lead vehicle), in order to preserve the correlation structure among those parameters.
Since the samples were randomly drawn from the two reference datasets, the correlations between parameters sampled jointly within each dataset (i.e., the correlations between $v_{f,init}$ and $\Delta v_l$ and between $v_{l,init}$ and $a_{l,min}$) should be preserved naturally.
Therefore, we only need a pairing process to preserve the correlations between $v_{f,init}$ and $v_{l,init}$ and between $v_{f,init}$ and $a_{l,min}$. 

The pairing process starts with computing the correlation values.
Since no reference dataset containing $v_{f,init}$, $v_{l,init}$, and $a_{l,min}$ was available, we gathered information on both vehicles from the COM\_b dataset, the only dataset that contains the three parameters and $\Delta v_l$.
However, we observed that COM\_b has substantial biases in the Delta-v of the lead vehicle ($\Delta v_l$) and the initial speed of the following vehicle ($v_{f,init}$).
For $\Delta v_l$, the bias is inherited as COM\_b contains the dataset PCM\_b.
For $v_{f,init}$, 15\% of the cases in COM\_b (mostly cases from PCM\_b) have a $v_{f,init}$ of exactly 50 km/h.
This may be because the following vehicle was assumed to be driven at the speed limit (i.e., 50 km/h) during the reconstruction, as no detailed information was available.
To mitigate the biases in these two parameters, the samples in COM\_b were weighted using the KNN sample weighting method to match REF\_f.
We then used the ”weights” package in R \cite{weightsrpackage} to compute the two (Pearson) correlation coefficients for the weighted data: $\Tilde{r}(v_{f,init}, v_{l,init})$ = 0.78 and $\Tilde{r}(v_{f,init}, a_{l,min})$ = -0.54.
(Note that, before weighting, those two correlations were 0.53 and -0.20, respectively.)
In addition to the strong correlation between the initial speeds of both vehicles, we observed that $v_{l,init}$ is no larger than $v_{f,init}$ in most cases, as shown in Fig. \ref{fig:scatterplot}.

Based on these observations, Algorithm \ref{alg:alg2} was developed to pair samples from two datasets, preserving the two correlations, and ensuring that $v_{l,init}$ is no larger than $v_{f,init}$ in most cases.
(Algorithm \ref{alg:alg2} is described in detail in Appendix \ref{section:pairingalgorithm}.)
The pairing results (i.e., REF\_i) were used as an approximation for the reference dataset of the four parameters ($\Tilde{\Phi}(v_{f,init}, \Delta v_l, v_{l,init}, a_{l,min})$).

\textit{\textbf{Sub-step 5 (Obtain REF\_b):}} At this point REF\_i and COM\_b were in place, so we created the target dataset REF\_b, the reference dataset of the initial states and the two minimum fitted accelerations ($d_{init}$, $v_{f,init}$ $a_{f,min}$, $v_{l,init}$, and $a_{l,min}$).
Again, we weighted samples in COM\_b with the KNN sample weighting method to match REF\_i, in order to mitigate biases in the four parameters ($v_{f,init}, \Delta v_l, v_{l,init}$, and $a_{l,min}$).

\subsection{Step 4: Distribution Modeling} \label{section:dataset8}
This step constructed a comprehensive distribution model for REF\_b and used the model to generate a synthetic dataset, which is used in the next step to generate representative rear-end crashes.

Because of the large number of parameters and the complexity of the distribution, REF\_b was divided into six sub-datasets (referred to as S1-6) that were modeled separately, using the multivariate distribution modeling method proposed in our previous study \cite{wu2024modeling}.
The six sub-datasets were categorized based on the relationship between the initial speeds of both vehicles, whether the following vehicle braked, and whether any vehicle was initially at a standstill (see Section \ref{section:results} for further information).
The overall distribution model for REF\_b, which can be seen as a mixture distribution model, was derived by combining the distribution models for all sub-datasets according to their relative proportions in REF\_b.
A synthetic dataset containing synthetic both-vehicle information (referred to as REF\_sb) with a sample size of 10,000 was then built with samples generated from the overall distribution model.

\subsection{Step 5: Simulation and Sample Weighting}
This step created a set of synthetic rear-end crashes representative of the population of such crashes with respect to severity based on REF\_sl, REF\_sb, and the reference marginal distributions for the three parameters $T$, $t_g$, and $t_a$ (obtained in the previous steps).
We first ran simulations of rear-end conflicts under different kinematic parameter settings drawn from the distribution(s) developed in the previous steps.
Second, we selected valid simulations (defined in the following sub-section) from the simulated set.
Finally, the selected crashes were weighted using Iterative proportional fitting (IPF) \cite{choupani2016population} so that the marginal distributions of parameters for the selected crashes matched the reference distributions.
We describe these three sub-steps (simulation setup, generation of synthetic rear-end crashes, and creation of a representative set of synthetic rear-end crashes) in more detail below.

\subsubsection{Simulation setup} \label{section:simulationsetting}
The simulation frequency is 20 Hz, and $t_{sim}$ is the simulation time (s).
At the start of each simulation ($t_{sim} = 0$ s), the initial states are set according to these three parameters: the initial following distance and the initial speeds of both vehicles ($d_{init}, v_{f,init}$, and $v_{l,init}$).
The lead vehicle follows its synthetic speed profile (the six-parameter model; $[v_{l,init}, a_1, a_2, \tau_s, \tau_1, \tau_2]$) until the simulation time reaches five seconds ($t_{sim} = 5$ s), after which it will keep its speed constant.
Meanwhile, the following vehicle follows the acceleration computed by the following-vehicle behavior model (see Section \ref{section:followingvehmodel}).
The simulation stops if a crash happens or a maximum simulation time is reached.

To be consistent with the input data of this study, a \textit{valid simulation} must fulfill two conditions: 1) a crash happens and 2) the crash moment $t_c$ is approximately five seconds after the start of the simulation (as the five seconds pre-crash data were extracted for all crashes in the original datasets).
That is, as noted earlier, not all simulations met the conditions; some produced crashes that did not belong to the final dataset.
However, it is unnecessarily stringent to have the crash occur exactly five seconds after starting the simulation.
To provide some margin for the crash timing, a crash moment error $t_e\ (=t_c - 5)$ was created.
The second condition then becomes $|t_e| \leq t_{e,thd}$, where $t_{e,thd}$ is a predefined threshold value, set to 0.2 s in this work.
(See Section \ref{section:results} for further information regarding the choice of $t_{e,thd}$.)
In addition, the maximum simulation time was set to six seconds.

\subsubsection{Generation of synthetic rear-end crashes}
This sub-step ran simulations and searched for the valid ones, in order to create a set of synthetic rear-end crashes. 
A matching algorithm was used to (first) create combinations of parameters among REF\_sl and REF\_sb (the synthetic datasets of lead-vehicle speed profiles, initial states, and minimum fitted accelerations) and marginal distributions of the three parameters ($T$, $t_g$, and $t_a$) for simulation and (second) to search for valid simulations (defined in the previous sub-step).
REF\_sl contains seven parameters: $v_{l,init}$, $a_{l,min}$, $a_1$, $a_2$, $\tau_s$, $\tau_1$, and $\tau_2$.
The parameter $a_{l,min}$ was computed for REF\_sl the same way it was for REF\_l.
REF\_sb contains five parameters: $d_{init}, v_{f,init}, a_{f,min}, v_{l,init}$, and $a_{l,min}$.
Because $v_{l,init}$ and $a_{l,min}$ are common to both, they were used to link samples from the two datasets. 
We could not simulate all potential parameter combinations due to the high computational load.
Therefore, for each sample drawn from REF\_sb, a \textit{matching algorithm} was designed to stop looping after obtaining a valid simulation or reaching a predefined maximum number of iterations.
The algorithm is briefly described below (with further information in Appendix \ref{section:matchingalgorithm}).
\begin{enumerate}
    \item Draw $N$ samples, with replacement, from REF\_sl and REF\_sb: $U = \{\textbf{U}_i |\ i \in [1, N]\}$ and $V = \{\textbf{V}_j |\ j \in [1, N]\}$.
    \item For $j = 1$ to $N$:
    \begin{enumerate}
        \item Compute the Euclidean distance (based on the common parameters $v_{l,init}$ and $a_{l,min}$) between $\textbf{V}_j$ and each sample in set $U$, select the samples with a Euclidean distance no larger than a predefined threshold $d_{e,thd}$ as the pairing candidates of $\textbf{V}_j$, and save the samples as set $W$.
        \item Update set $W$ according to the sub-dataset which $\textbf{V}_j$ belongs to.
        For instance, a $\textbf{V}_j$ from sub-dataset 1 (S1) requires that $v_{f,init} > v_{l,init} > 0$.
        \item Create sets of candidates for the three parameters $T$, $t_g$, and $t_a$: $\bar{T}^*$, $\bar{t}_g^*$, and $\bar{t}_a^*$.
        Note that, in those sets, the samples are ordered randomly.
        \item Loop through $W$, $\bar{T}^*$, $\bar{t}_g^*$, and $\bar{t}_a^*$ until a valid simulation is obtained or the predefined maximum number of iterations is reached.
        Save the log when there is a valid simulation.
    \end{enumerate}
\end{enumerate}

\subsubsection{Creation of a representative set of synthetic rear-end crashes} \label{section:ipf}
A synthetic rear-end crash dataset (REF\_ss) was created, which includes all valid simulations.
The next step was weighting samples in the synthetic dataset to match the two reference datasets (REF\_sl and REF\_sb) and the reference marginal distributions of the remaining three parameters ($T$, $t_g$, and $t_a$) in the following-vehicle behavior model.
It is important to note that when modeling the two reference datasets, each one was split into multiple sub-datasets so that simpler models could be used for each sub-dataset.
The overall distribution model was then derived by combining the distribution models for the sub-datasets according to the sub-dataset proportions.

The objectives of this sample weighting were to retain 1) the reference marginal distribution of each of the three parameters ($T$, $t_g$, and $t_a$), 2) the proportion of each sub-dataset, and 3) the marginal distribution of each parameter for each sub-dataset instead of the whole.
Iterative proportional fitting (IPF) \cite{choupani2016population} was used to achieve these objectives.
In the algorithm, the Kolmogorov–Smirnov (KS) statistic was used to measure the difference between each weighted marginal distribution and its corresponding reference distribution.
The implementation consisted of the following steps:

\begin{enumerate}
    \item Set initial weight for all samples to 1: $w_i^{(0)} = 0$, $\forall i \in [1,N]$.
    \item For iteration number $k$ = 1 to 100:
    \begin{enumerate}
        \item For each parameter other than the three parameters ($T$, $t_g$, and $t_a$) of each sub-dataset, update the weights for samples in the sub-dataset using IPF.
        \item For each of the three parameters ($T$, $t_g$, and $t_a$), update the weights for all samples using IPF.
        \item Scale the weights so that $\sum_{i=1}^{N} w_i^{(k)} = N$.
        \item Compute the KS statistic between the weighted synthetic crash dataset and corresponding reference distributions with the weighted two-sample KS tests using the ”Ecume” package in R \cite{ecumerpackage}: $\{s_j^{(k)} | \forall j \in [1,n]\}$.
        \begin{enumerate}
            \item For each sub-dataset, compute the KS statistic between the marginal distribution in the sub-dataset and the reference distribution for each parameter other than the three parameters ($T$, $t_g$, and $t_a$).
            \item For each of the three parameters ($T$, $t_g$, and $t_a$), compute the KS statistic between the weighted marginal distribution in the overall synthetic crash dataset and the reference dataset.
        \end{enumerate}
        \item Compute the loss: $L_w^{(k)} = \sum_{j=1}^n [(s_j^{(k)})^2 \sum w_{l_j}^{(k)}]$, where $w_{l_j}^{(k)}$ is the weight for the corresponding sample of $s_j^{(k)}$.
    \end{enumerate}
    \item Select the optimal weights with the minimum loss: $\{w_i^{(k^*)} |i \in [1,N]\}$, where $k^* = \arg \underset{k}{\min} (\{L_w^{(k)} | k \in [1,n]\})$. 
\end{enumerate}

\subsection{Step 6: Validation} \label{section:validation}
In terms of validation, firstly, all three objectives of the sample weighting in the last step must be achieved.
Consequently, the proportion of each sub-dataset was checked, and two-sample KS tests were conducted for each parameter in each sub-dataset to compare the marginal distributions in the weighted synthetic crash dataset with their corresponding reference distributions.

Secondly, for each sub-dataset containing multiple parameters, we needed to verify the similarity of the overall multivariate distributions between the synthetic and reference datasets.
To do this, we utilized the t-distributed stochastic neighbor embedding (t-SNE) technique \cite{van2008visualizing} to transform the multidimensional data into unidimensional data and then conducted a two-sample KS test on the transformed data.

Lastly, the crash severity levels between the synthetic and reference datasets also needed to be compared.
For each synthetic crash, the Delta-v of the lead vehicle ($\Delta v_l$) was computed using the Kudlich-Slibar rigid body impulse model \cite{kudlich1966beitrag}, in which the coefficient of restitution $e$ (the ratio of the post-impact vehicle-velocity-difference to the pre-impact vehicle-velocity-difference between the two vehicles) was computed as suggested in an existing study \cite{leifer2013supplemental}:
\begin{equation} \label{eq:e}
    \begin{array}{l}
    e = 0.47477 - 0.26139 \cdot log(\Delta v_{pre})\ ...\\
    \ \ \ \ \ + 0.03382 \cdot (log(\Delta v_{pre}))^2\ ...\\
    \ \ \ \ \ - 0.1139 \cdot (log(\Delta v_{pre}))^3,
    \end{array}
\end{equation}
where $\Delta v_{pre}$ is the pre-impact vehicle-velocity-difference (m/s). 
The weighted distribution of $\Delta v_l$ was then compared with the reference distribution of $\Delta v_l$ (obtained in REF\_l), using the weighted two-sample KS test.

\section{Results} \label{section:results}
\subsection{Data Combination}
\subsubsection{Fitting of speed profiles}
\begin{figure}[!t]
    \centering
    \subfloat[]{\includegraphics[width=0.24\textwidth]{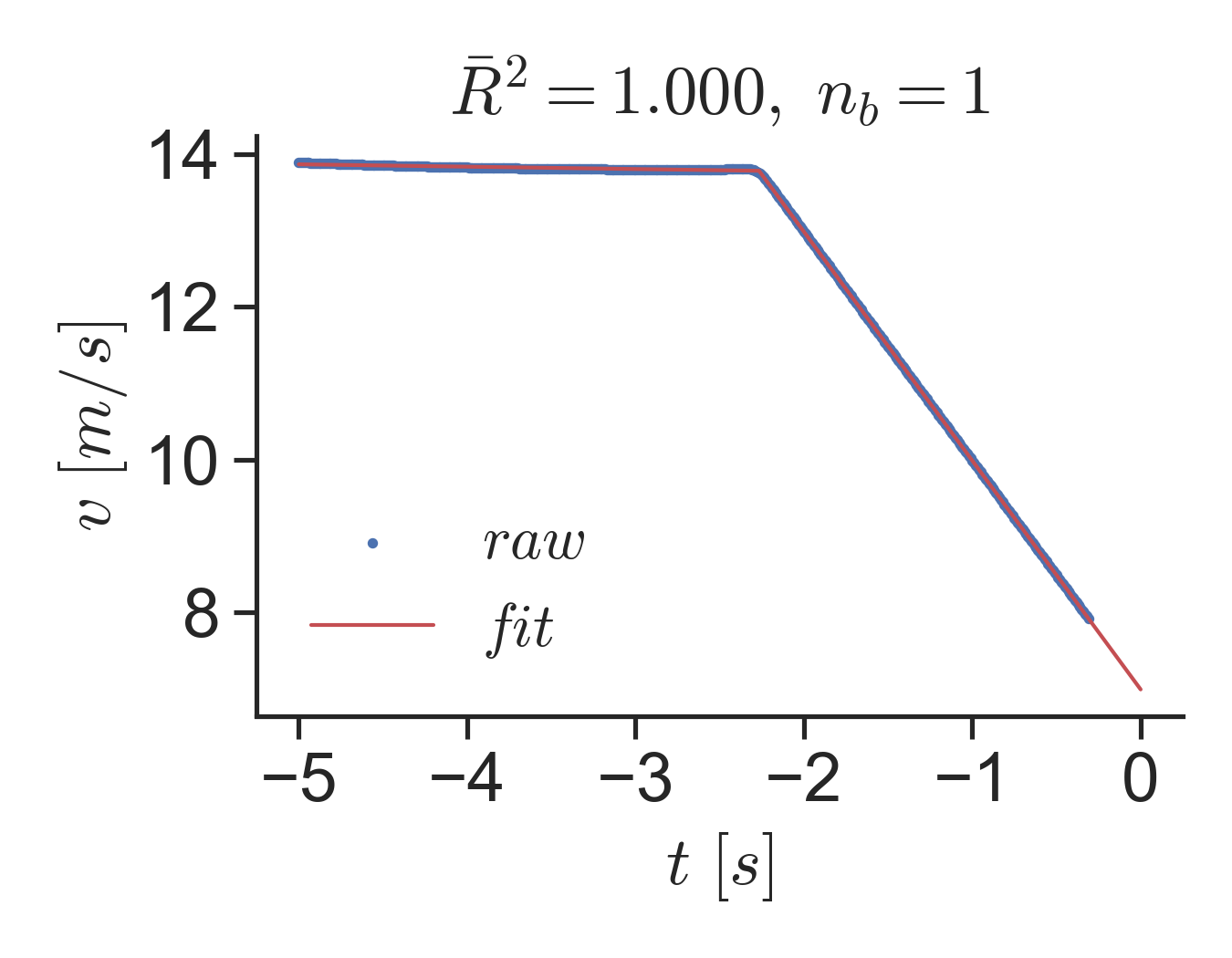}}
    \hfil
    \subfloat[]{\includegraphics[width=0.24\textwidth]{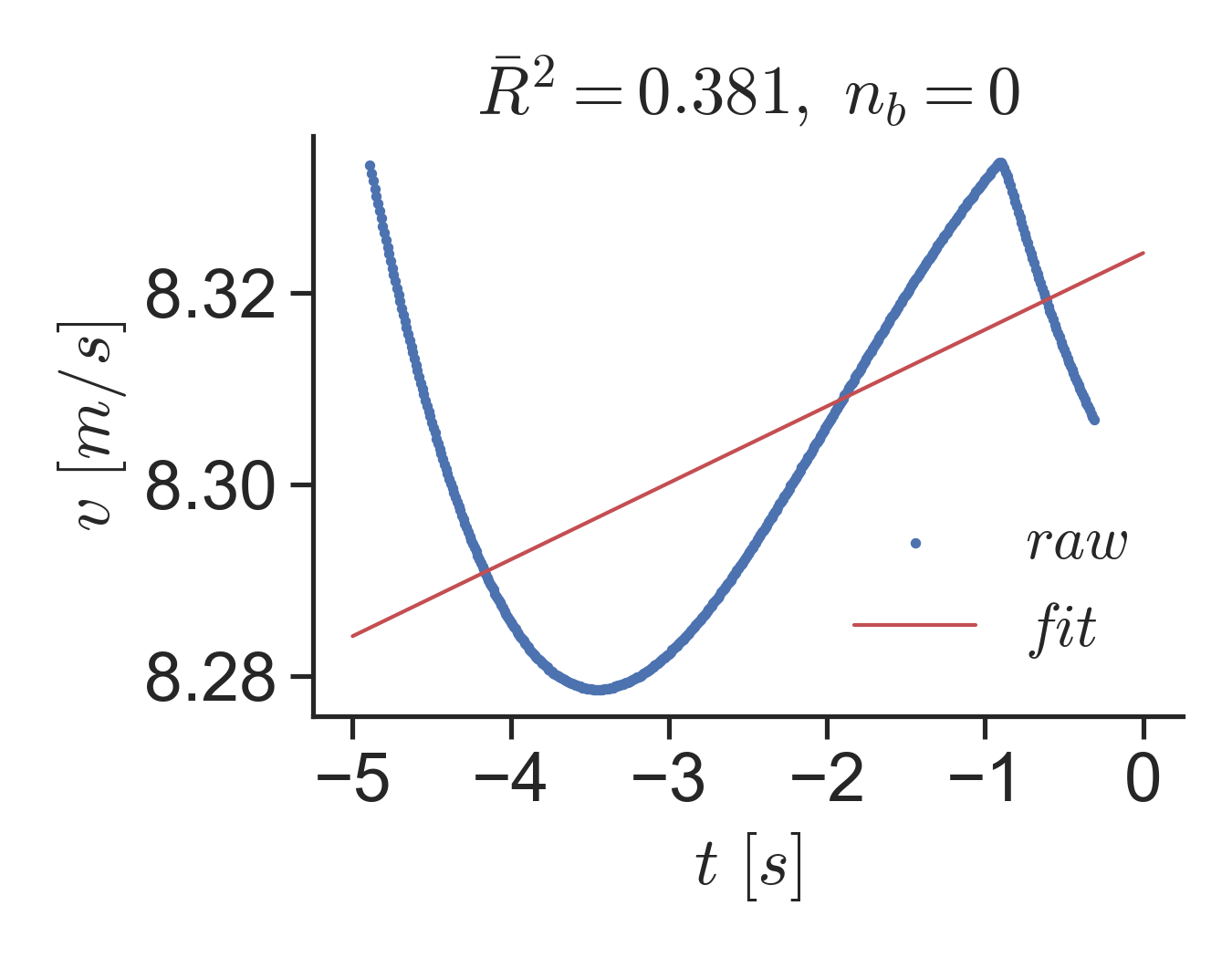}}
    \caption{Examples of speed profile fit results. $n_b$ is the number of breakpoints.}
    \label{fig:examplesoffit}
\end{figure}

The speed profiles of both vehicles in SHRP2\_b and PCM\_b were fitted into the piecewise linear model.
84.2\% (1540 out of 1828) of the speed profiles have an adjusted R-squared $\bar{R}^2$ greater than 0.9: see Fig. \ref{fig:examplesoffit}(a).
Of the remaining 15.8\% (288), 93.1\% (268) showed a speed change (i.e., the difference between the maximum and minimum speed) of less than 0.5 m/s.
For those 268 profiles with only minor speed changes, the piecewise linear model simplified the speed profile as a straight line, which led to a lower $\bar{R}^2$: see Fig. \ref{fig:examplesoffit}(b).
The remaining 6.9\% (20) of the cases have an adjusted R-squared $\bar{R}^2$ greater than 0.75.

\subsubsection{Mass ratio distribution}
\begin{figure}[!t]
    \centering
    \includegraphics[width=0.35\textwidth]{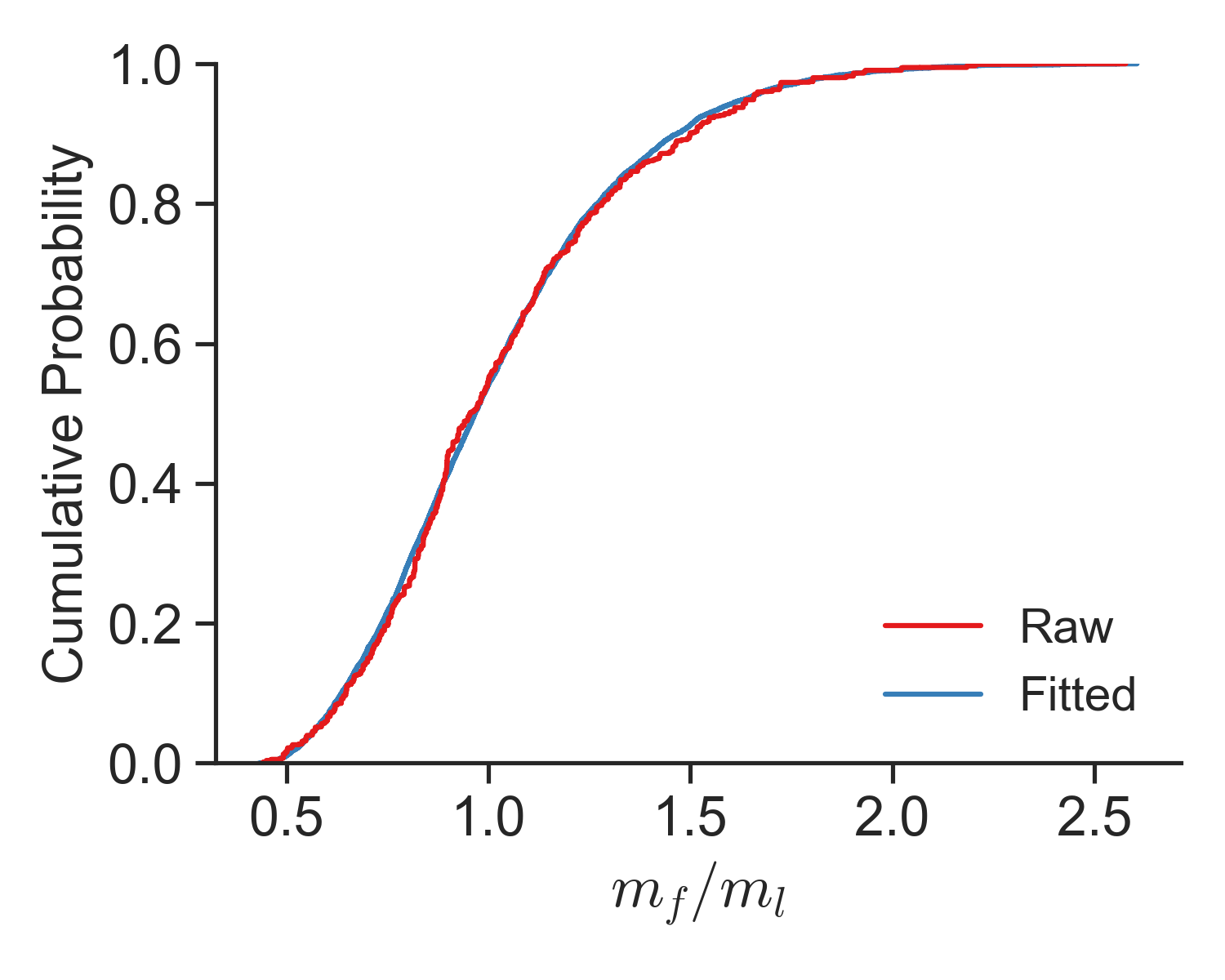}
    \caption{Fitting of the mass ratio distribution in CISS\_m.
    KS test results: sample size $n$ = 748, statistic = 0.04, p-value = 0.79.}
    \label{fig:massratio}
\end{figure}
The generalized gamma distribution was selected for fitting the mass ratio data in CISS\_m.
A two-sample KS test was conducted between the raw and the fitted distributions.
The results do not indicate any significant difference (see Fig. \ref{fig:massratio}).

\subsubsection{Reference datasets}
\begin{figure}[!t]
    \centering
    \subfloat[]{\includegraphics[width=0.24\textwidth]{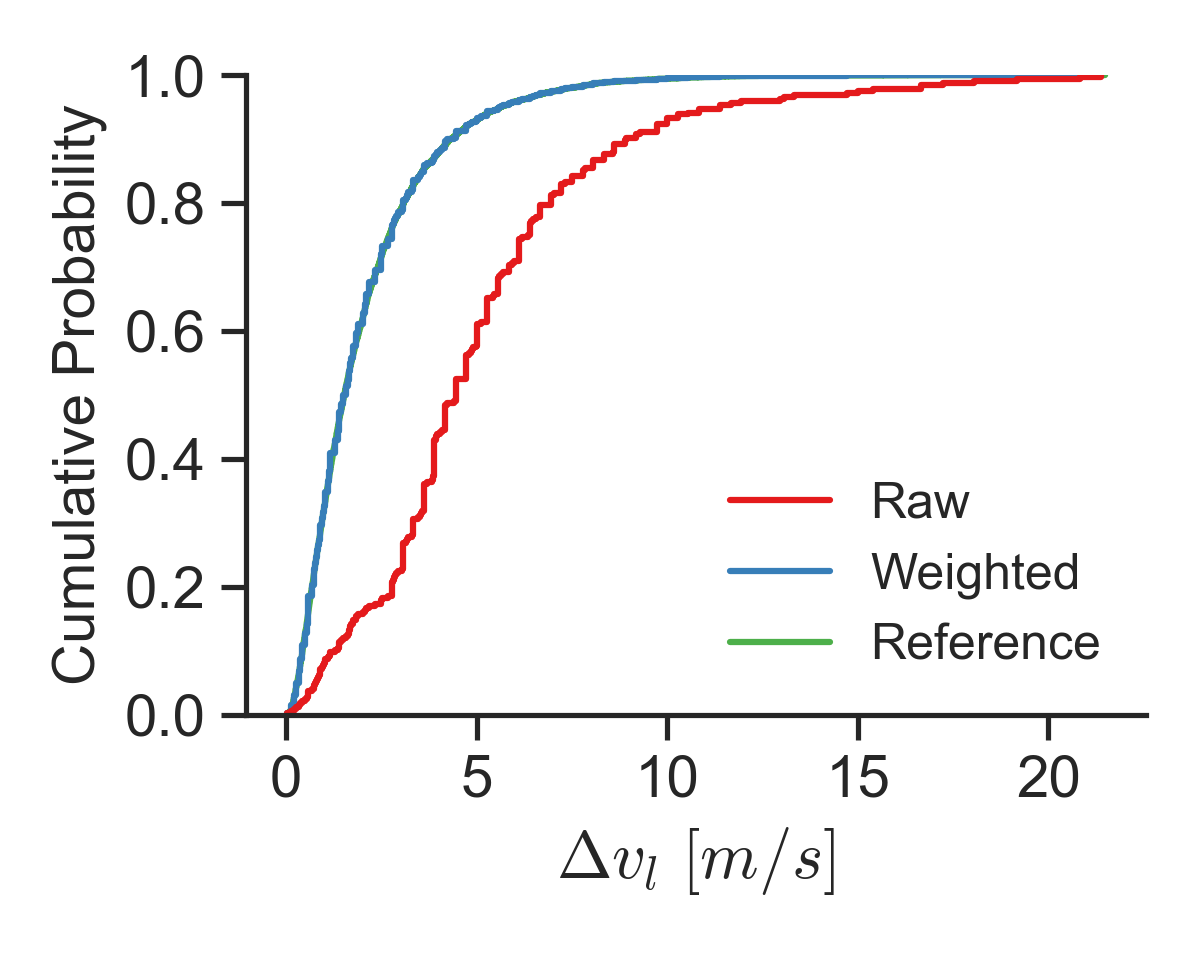}}
    \hfil
    \subfloat[]{\includegraphics[width=0.24\textwidth]{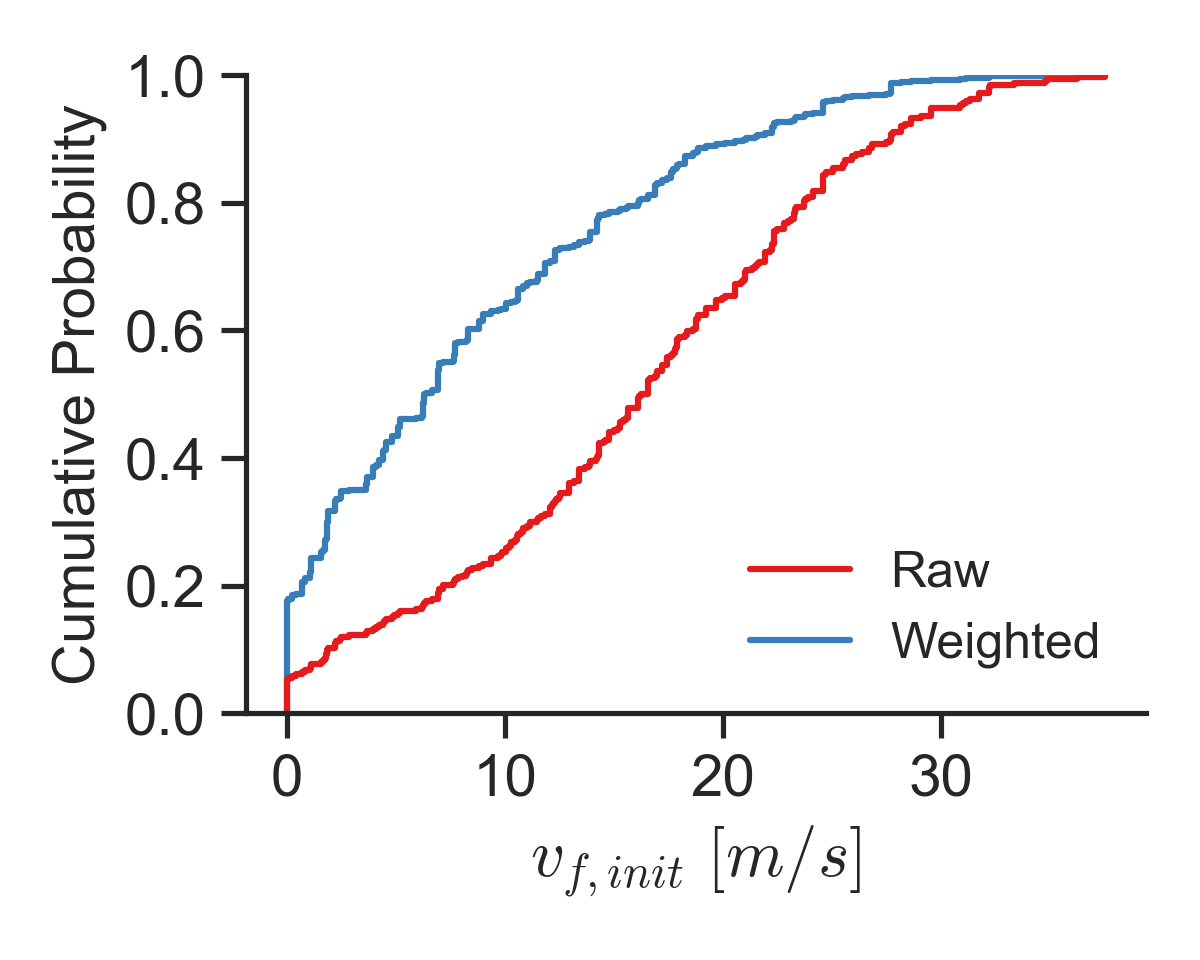}}
    \caption{KNN sample weighting results of the raw dataset COM\_f according to the reference distribution of $\Delta v_l$: (a) $\Delta v_l$, and (b) $v_{f,init}$.
    The blue and green lines are almost identical in (a). 
    The weighted two-sample KS test results between the weighted samples and reference distribution of $\Delta v_l$: valid sample size (i.e., the sum of weights) of the weighted distribution $n$ = 324, sample size of the reference distribution $n_{r}$ = 10,000, statistic = 0.03, and p-value = 1.00.}
    \label{fig:weightedcombinedfollowing}
\end{figure}

The samples in the raw dataset COM\_f (n=524) were weighted using the KNN sample weighting method according to $\Tilde{\Phi}(\Delta v_l)$ from REF\_l.
There were 324 samples with a weight value larger than zero.
The cumulative distribution functions (CDFs) of $\Delta v_l$ and $v_{f,init}$ are shown in Figs. \ref{fig:weightedcombinedfollowing} (a)-(b), respectively.
The weighted $v_{f,init}$ distribution was then used as the reference distribution of $v_{f,init}$.
The weighted two-sample KS test between the weighted samples and the reference distribution of $\Delta v_l$ shows no significant difference.
Compared with the raw distribution, the weighted distribution $\Tilde{\Phi}(\Delta v_l, v_{f,init})$ has a higher proportion of low-severity (i.e., small $\Delta v_l$) and low-speed (i.e., small $v_{f,init}$) crashes.

\begin{figure}[!t]
    \centering
    \subfloat[]{\includegraphics[width=0.24\textwidth]{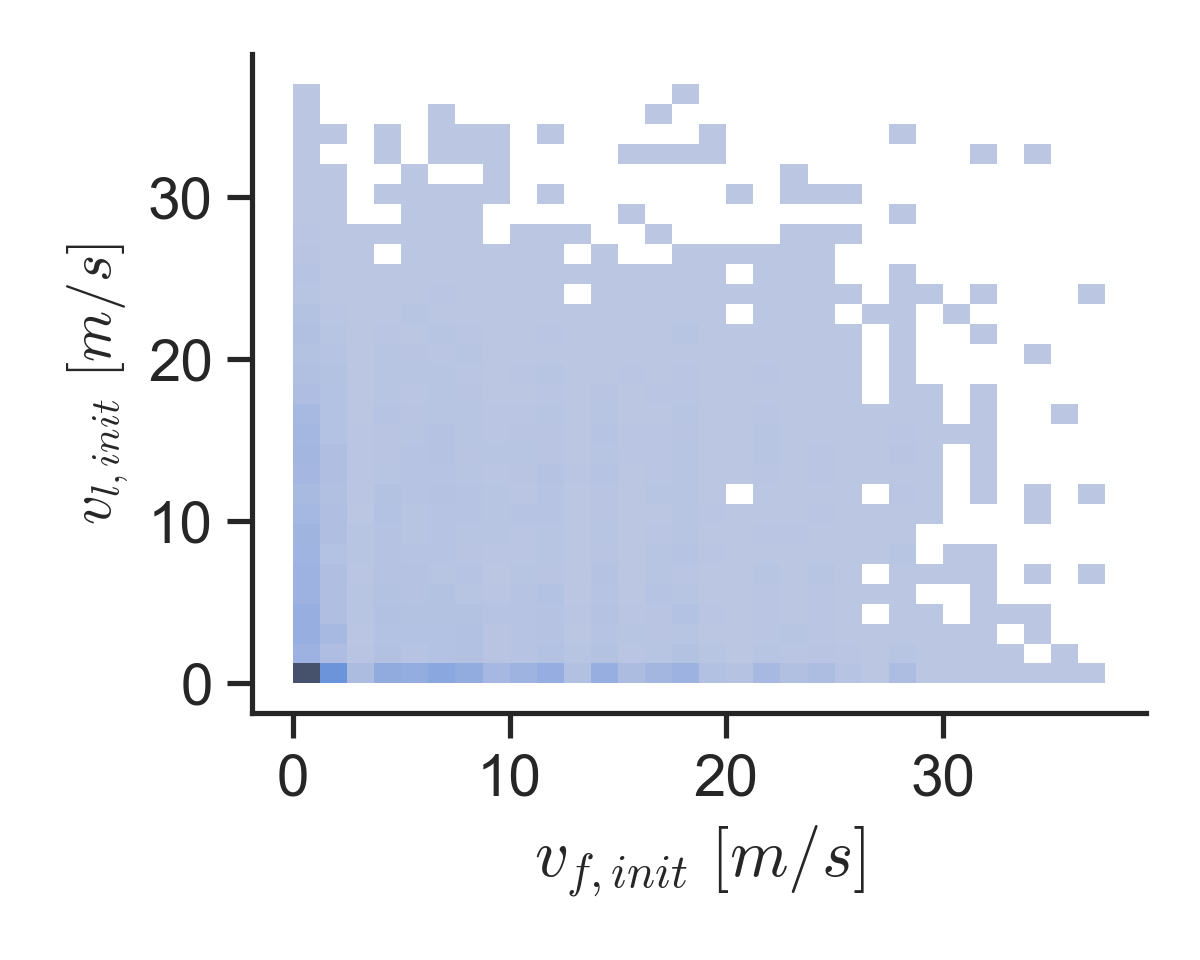}}
    \hfil
    \subfloat[]{\includegraphics[width=0.24\textwidth]{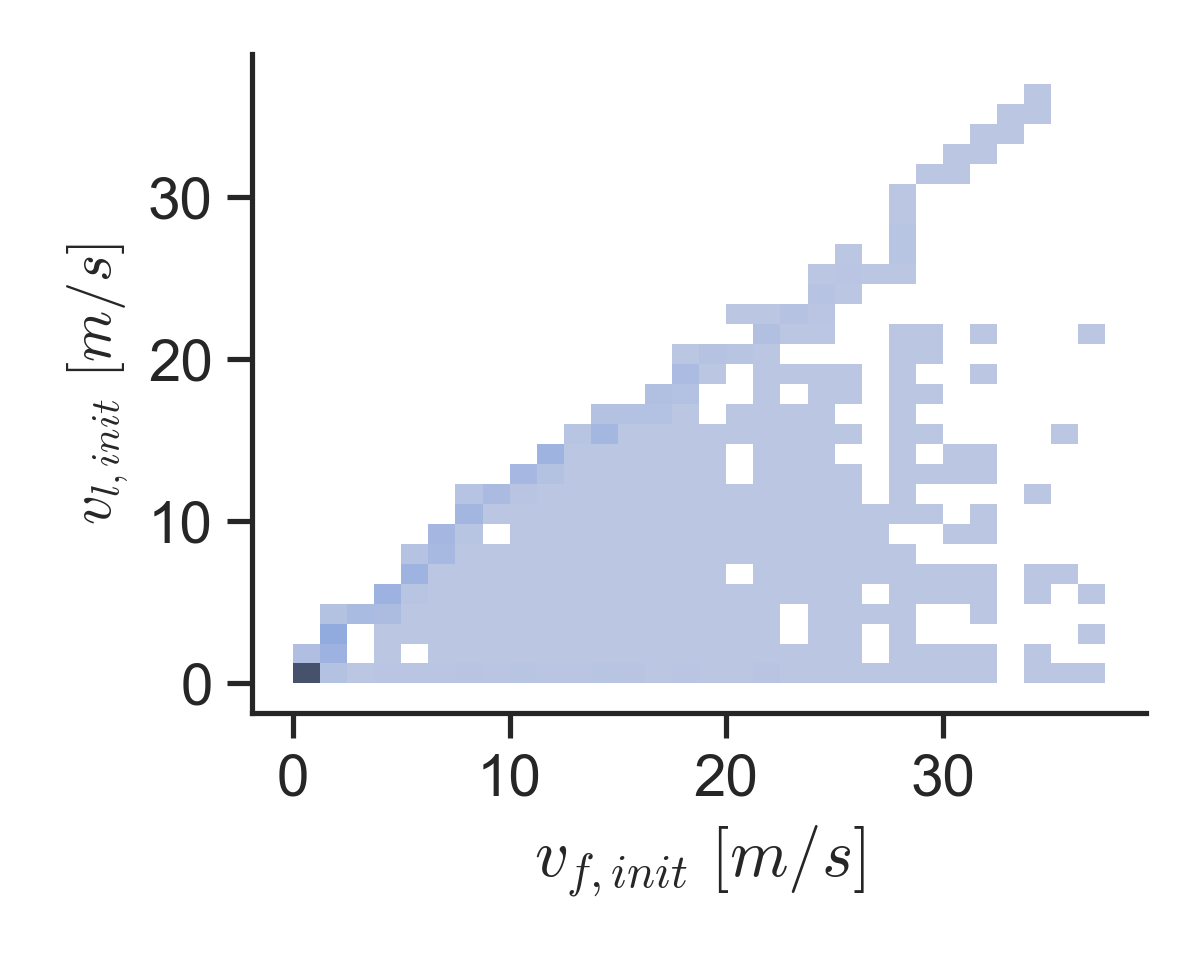}}
    \caption{Joint distribution of $v_{f,init}$ and $v_{l,init}$ before (a) and after (b) pairing samples from $\Tilde{\Phi}(v_{l,init}$, $a_{l,min})$ and $\Tilde{\Phi}(v_{f,init}$, $\Delta v_l)$.}
    \label{fig:pair}
\end{figure}

The reference dataset REF\_i was created by pairing an equal number of samples randomly selected (with replacement) from two reference datasets using the pairing algorithm described in Appendix \ref{section:pairingalgorithm}.
Fig. \ref{fig:pair} shows the joint distribution of the initial speeds of both vehicles before and after pairing the selected samples.
The target correlations to preserve were: $\Tilde{r}(v_{f,init}$, $v_{l,init})$ = 0.78 and $\Tilde{r}(v_{f,init}$, $a_{l,min})$ = -0.54.
The pairing algorithm effectively retained the correlations among relevant parameters in the pairing results (i.e., REF\_i): $r(v_{f,init}$, $v_{l,init})$ = 0.78 and $r(v_{f,init}$, $a_{l,min})$ = -0.54.


\begin{table}[!t] 
\caption{Comparison between the weighted and reference marginal distributions for four weighting parameters\label{tab:comparisonfourparameters}} 
\centering
\begin{threeparttable}
\begin{tabular}{ccccc}
\hline
Parameter & $n^*$ & $n_r^{**}$ & KS statistic & p-value\\
\hline
$v_{f,init}$ & \multirow{4}{*}{852} & \multirow{4}{*}{10,000} & 0.08 & 0.54\\
$\Delta v_l$ & & & 0.10 & 0.32\\
$v_{l,init}$ & & & 0.10 & 0.29\\
$a_{l,min}$ & & & 0.09 & 0.43\\
\hline
\end{tabular}
\begin{tablenotes}
\RaggedRight
\item $^*$ Valid sample size = sum of sample weights.
\item $^{**}$ Sample size of the reference data.
\end{tablenotes}
\end{threeparttable}
\end{table}

\begin{figure}[!t]
    \centering
    \subfloat[]{\includegraphics[width=0.24\textwidth]{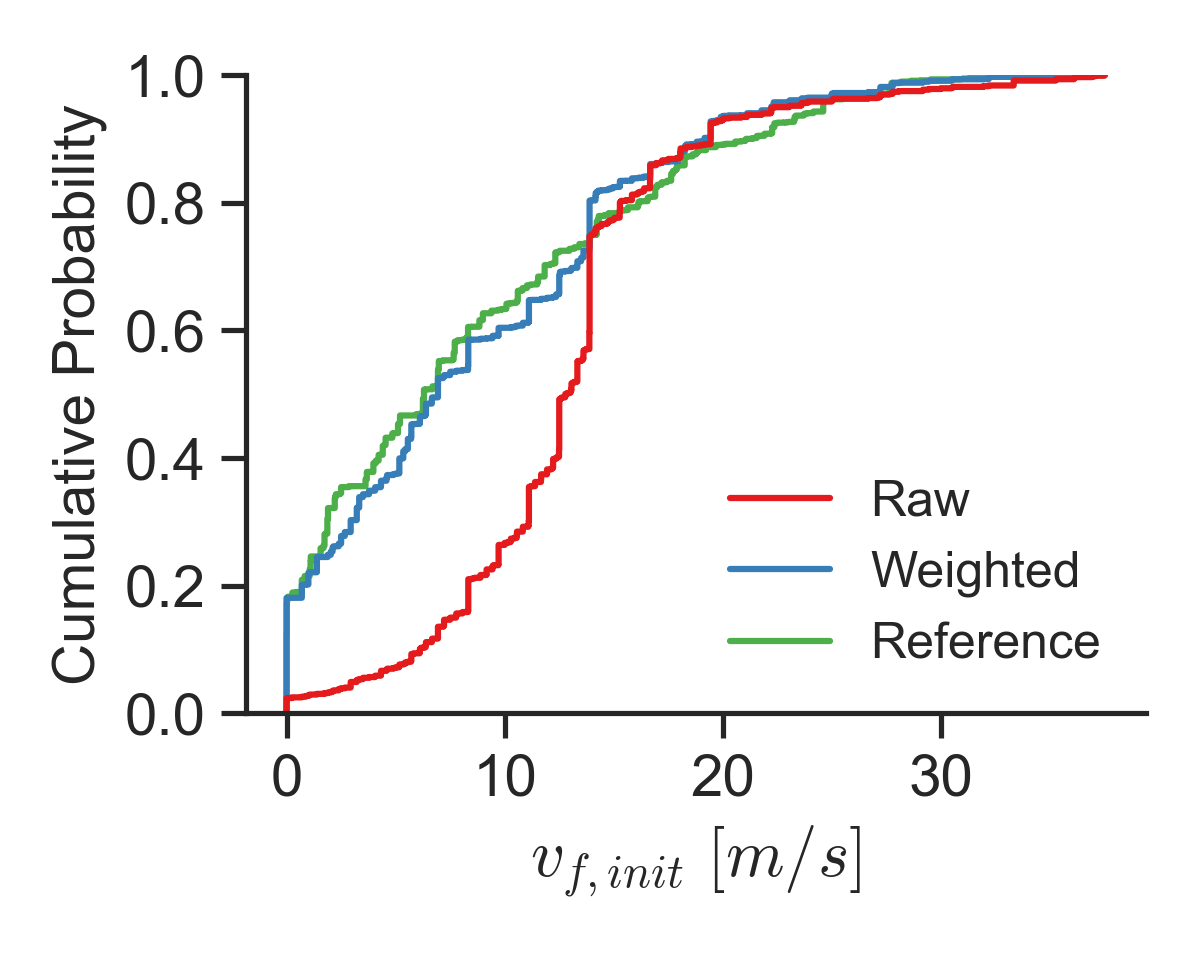}}
    \hfil
    \subfloat[]{\includegraphics[width=0.24\textwidth]{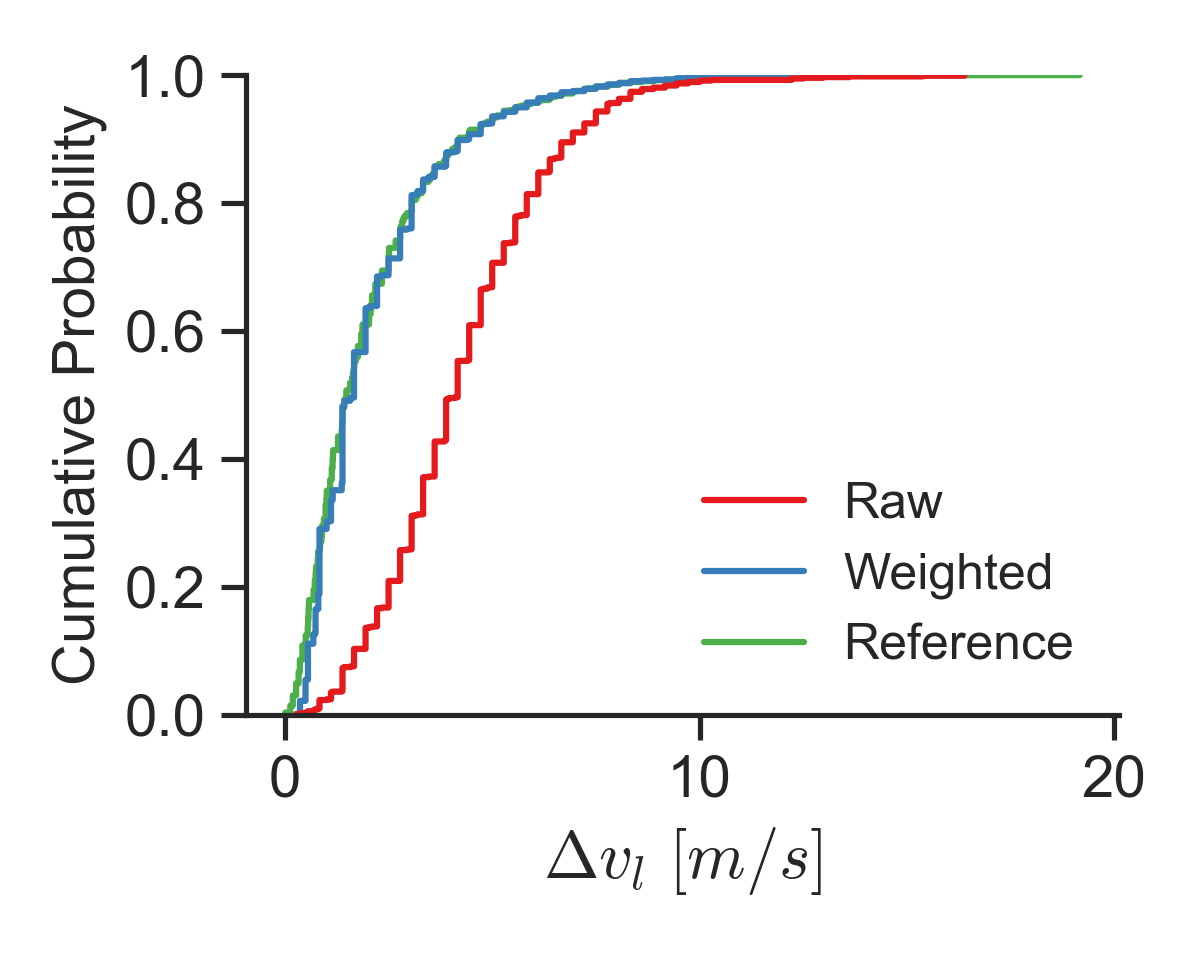}}
    \vfil
    \subfloat[]{\includegraphics[width=0.24\textwidth]{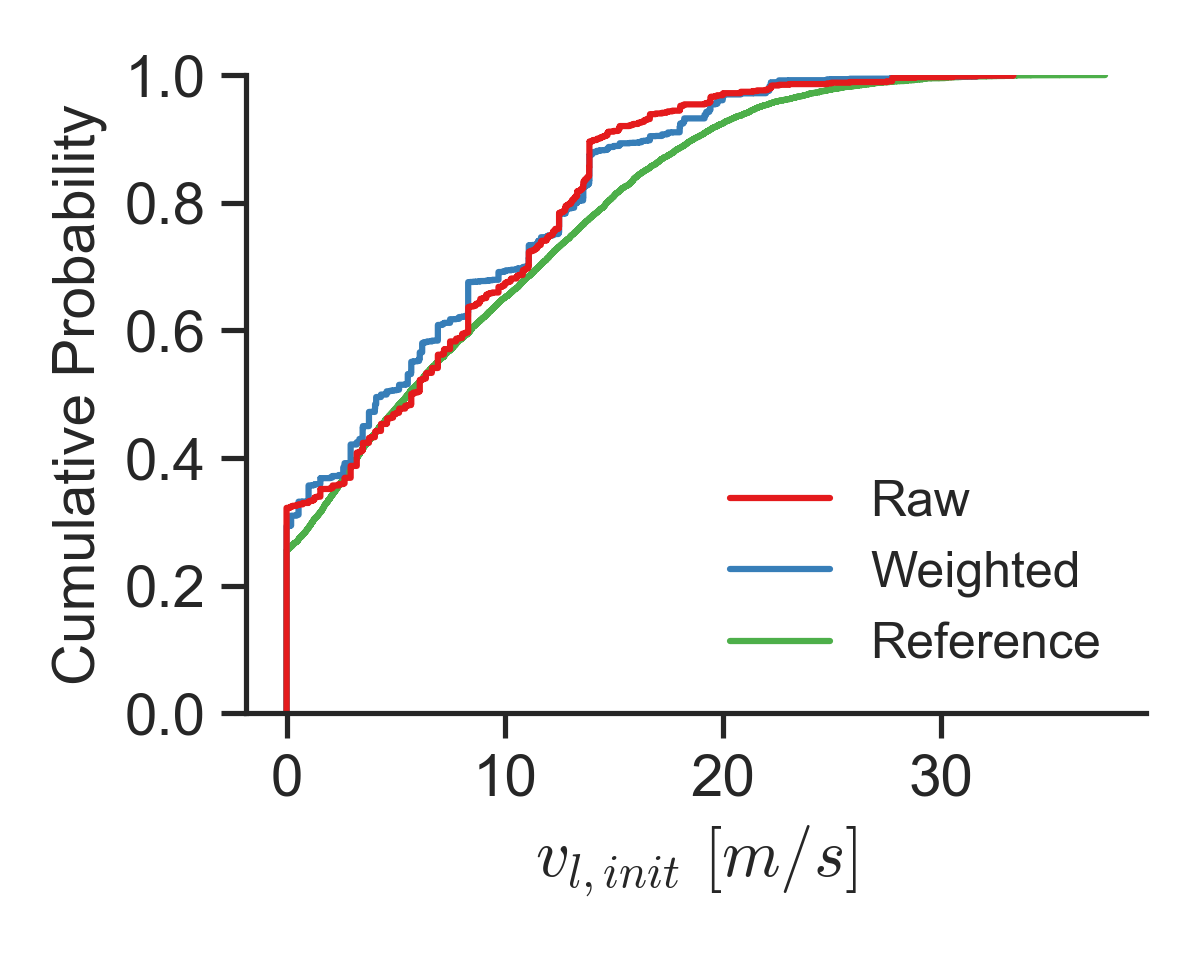}}
    \hfil
    \subfloat[]{\includegraphics[width=0.24\textwidth]{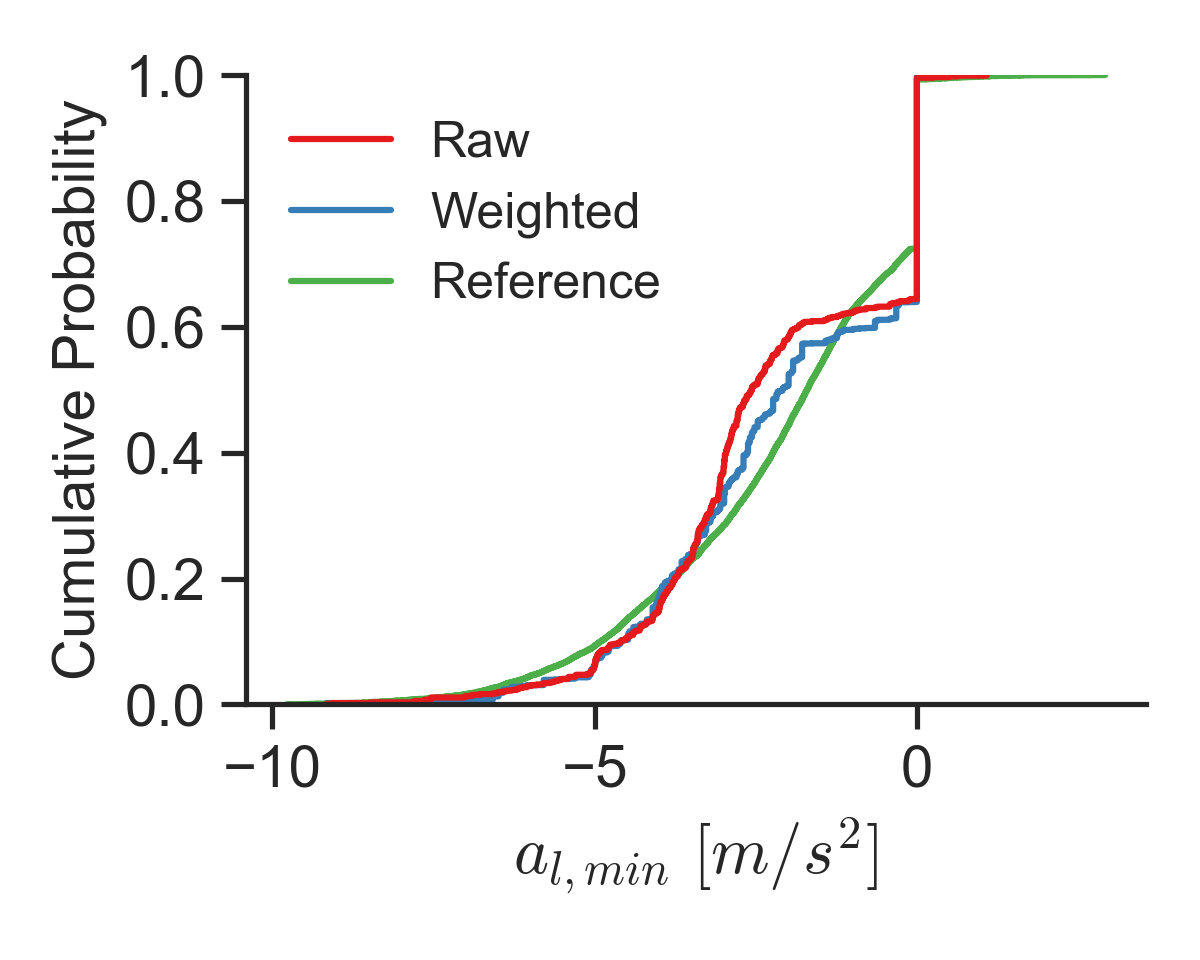}}
    \caption{KNN sample weighting results of the raw combined dataset COM\_b (obtained by combining SHRP2\_b and COM\_b) according to REF\_i.}
    \label{fig:weightedinitialstates}
\end{figure}

REF\_b (i.e., the reference dataset of the initial states and minimum fitted accelerations of both vehicles) was created by weighting samples in COM\_b (the combined dataset of both vehicles) using the KNN sample weighting method according to the intermediate reference dataset REF\_i.
There were 852 samples (out of 913) with a weight value larger than zero.
The weighted two-sample KS tests were conducted to test whether the weighted and the reference data are from the same distribution.
The results in Table \ref{tab:comparisonfourparameters} do not indicate any significant difference.
However, it is worth noting that the lack of significance does not necessarily imply that the datasets are from the same distribution.
Nonetheless, a visual comparison of the well-aligned weighted CDFs for each of the four parameters ($v_{f,init}$, $\Delta v_l$, $v_{l,init}$, and $a_{l,min}$) in the weighted and reference distributions indicates substantial similarities (see Fig. \ref{fig:weightedinitialstates}). 

\subsection{Modeling of REF\_b}
\subsubsection{Data categorization} \label{section:categorization}
REF\_b, the reference dataset of the initial speeds and minimum fitted accelerations of both vehicles and the initial distance, was divided into six sub-datasets (based on the relationship between the initial speeds of both vehicles, whether the following vehicle braked, and whether either vehicle was initially stationary).

\begin{table}[!t]
    \centering
    \caption{Six sub-datasets} \label{tab:category}
    \begin{threeparttable}
        \begin{tabular}{cccc}
            \hline
            \multirow{2}{*}{Sub-dataset} & \multicolumn{2}{c}{Conditions} & \multirow{2}{*}{Proportion}\\
            \cmidrule(rl){2-3}
            & Initial speeds [m/s] & Acceleration$^*$ [m/s$^2$]&\\
            \hline
            S1 & $v_{init,f} > v_{init,l} > 0$ & $a_{f,min} \geq 0$ & 9.6\%\\
            S2 & $v_{init,f} > v_{init,l} > 0$ & $a_{f,min} < 0$ & 30.9\%\\
            S3 & $v_{init,f} > v_{init,l} = 0$ & \textemdash & 13.1\%\\
            S4 & $v_{init,f} = v_{init,l} = 0$ & \textemdash & 16.3\%\\
            S5 & $0 < v_{init,f} \leq v_{init,l}$ & $a_{f,min} \geq 0$ & 12.6\%\\
            S6 & $0 < v_{init,f} \leq v_{init,l}$ & $a_{f,min} < 0$ & 17.5\%\\
            \hline
        \end{tabular}
        \begin{tablenotes}
            \RaggedRight
            \item $^*$ $a_{f,min} \geq$ 0 m/s$^2$ means the following vehicle did not brake in the event, while $a_{f,min} < $ 0 m/s$^2$ indicates the following vehicle braked in the event.
        \end{tablenotes}
    \end{threeparttable}
\end{table}

Table \ref{tab:category} shows the six sub-datasets (S1–S6), including their corresponding proportions.

In both S1 and S2, the following vehicle initially approached the moving lead vehicle (i.e., $v_{init,f} > v_{init,l} >$ 0 m/s).
The following vehicle braked in S2 (i.e., $a_{f,min} < 0$ m/s$^2$), but not in S1 (i.e., $a_{f, min} \leq 0$ m/s$^2$).

In S3, the following vehicle initially approached the stationary lead vehicle – while in S4, both vehicles were initially stationary.
Abnormal acceleration behaviors are present in 56.2\% of the cases in S4 (9.2\% of all cases).

In S5 and S6, the lead vehicle initially moved away from the moving following vehicle.
In S6, the following vehicle braked (i.e., $a_{f, min} < 0$ m/s$^2$), while in S5 it did not (i.e., $a_{f, min} \leq 0$ m/s$^2$).

It is important to note that some of the five parameters in certain sub-datasets can be constant.
For instance, the following vehicle's minimum fitted acceleration ($a_{f, min}$) is zero for all cases in S1 since the following vehicle did not brake.
Only the non-constant parameters in each sub-dataset were modeled.

\subsubsection{Comparison between the synthetic and reference datasets of selected parameters}
\begin{table*}[!t] 
\caption{Comparison between the synthetic and reference datasets of selected parameters \label{tab:comparisonrefandsyn}} 
\centering
\begin{threeparttable}
\begin{tabular}{cccccccc}
\hline
\multirow{3}{*}{Parameter} & \multirow{3}{*}{Unit} & \multicolumn{2}{c}{Reference} & \multicolumn{2}{c}{Synthetic} & \multirow{3}{*}{KS statistic} & \multirow{3}{*}{p-value}\\
& & \multicolumn{2}{c}{($n=852^*$)} & \multicolumn{2}{c}{($n=10,000$)}\\
\cmidrule(rl){3-4} \cmidrule(rl){5-6}
& & Mean & SD & Mean & SD & \\
\hline
$d_{init}$ & $m$ & 17.62 & 21.83 & 18.05 & 21.10 & 0.05 & 0.95\\
$v_{f,init}$ & $m/s$ & 8.56 & 7.49 & 8.77 & 7.31 & 0.07 & 0.78\\
$a_{f,min}$ & $m/s^2$ & -2.40 & 2.83 & -2.43 & 2.72 & 0.06 & 0.93\\
$v_{l,init}$ & $m/s$ & 6.58 & 6.75 & 6.68 & 6.82 & 0.05 & 0.94\\
$a_{l,min}$ & $m/s^2$ & -2.03 & 1.93 & -2.06 & 1.92 & 0.04 & 0.99\\
\hline
\end{tabular}
\begin{tablenotes}
\RaggedRight
\item $^*$ Valid sample size = sum of sample weights.
\end{tablenotes}
\end{threeparttable}
\end{table*}

\begin{figure}[!t]
    \centering
    \subfloat[]{\includegraphics[width=0.24\textwidth]{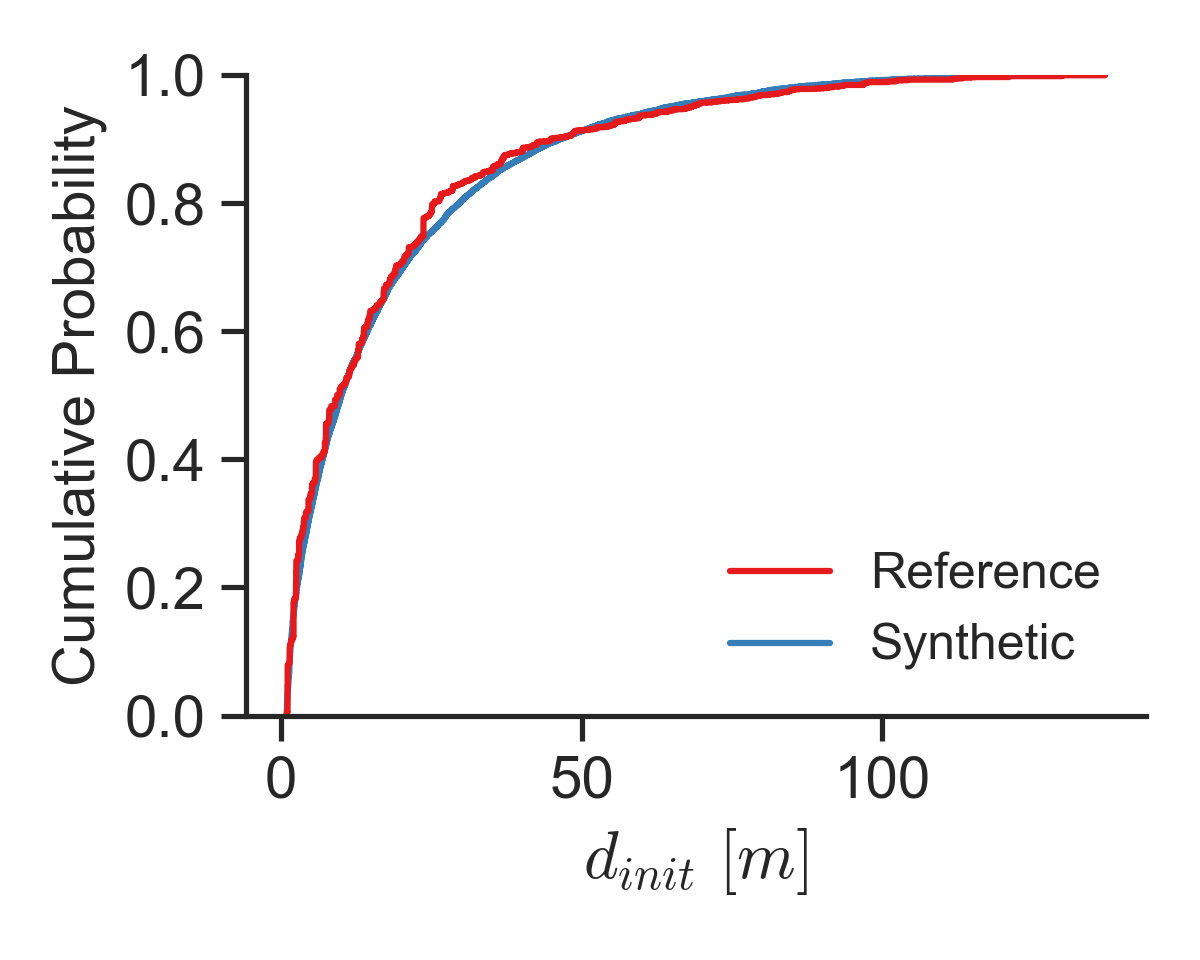}}
    \hfil
    \subfloat[]{\includegraphics[width=0.24\textwidth]{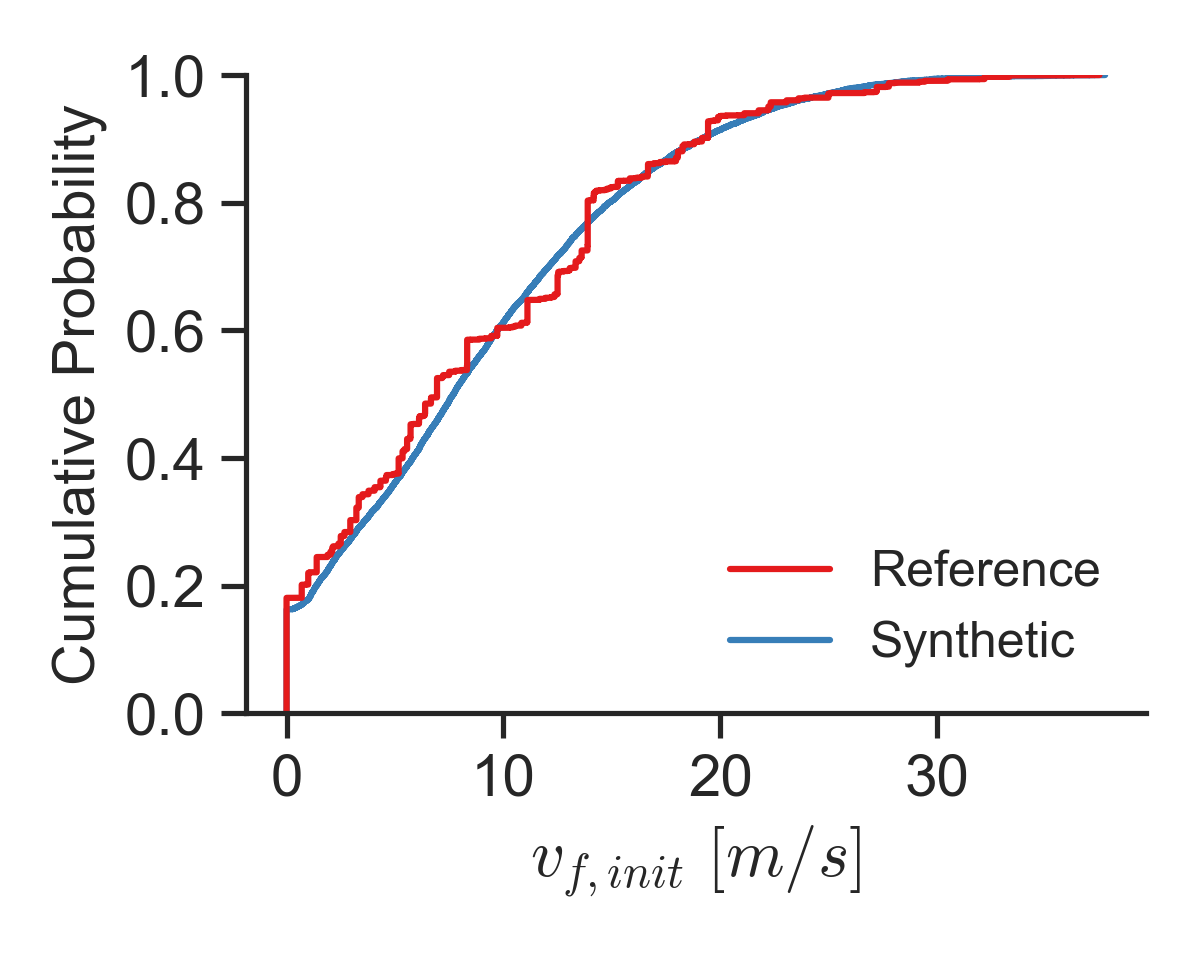}}
    \vfil
    \subfloat[]{\includegraphics[width=0.24\textwidth]{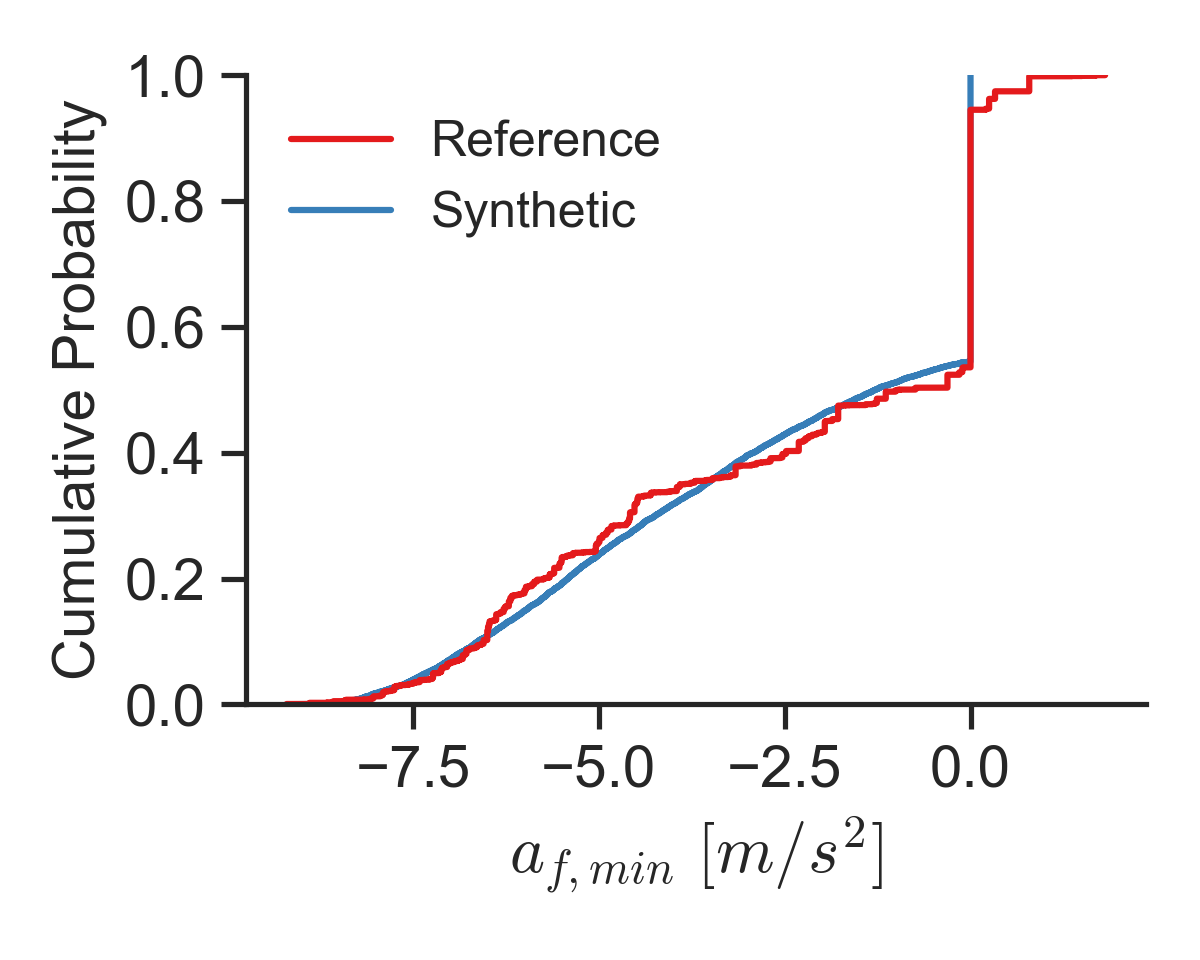}}
    \hfil
    \subfloat[]{\includegraphics[width=0.24\textwidth]{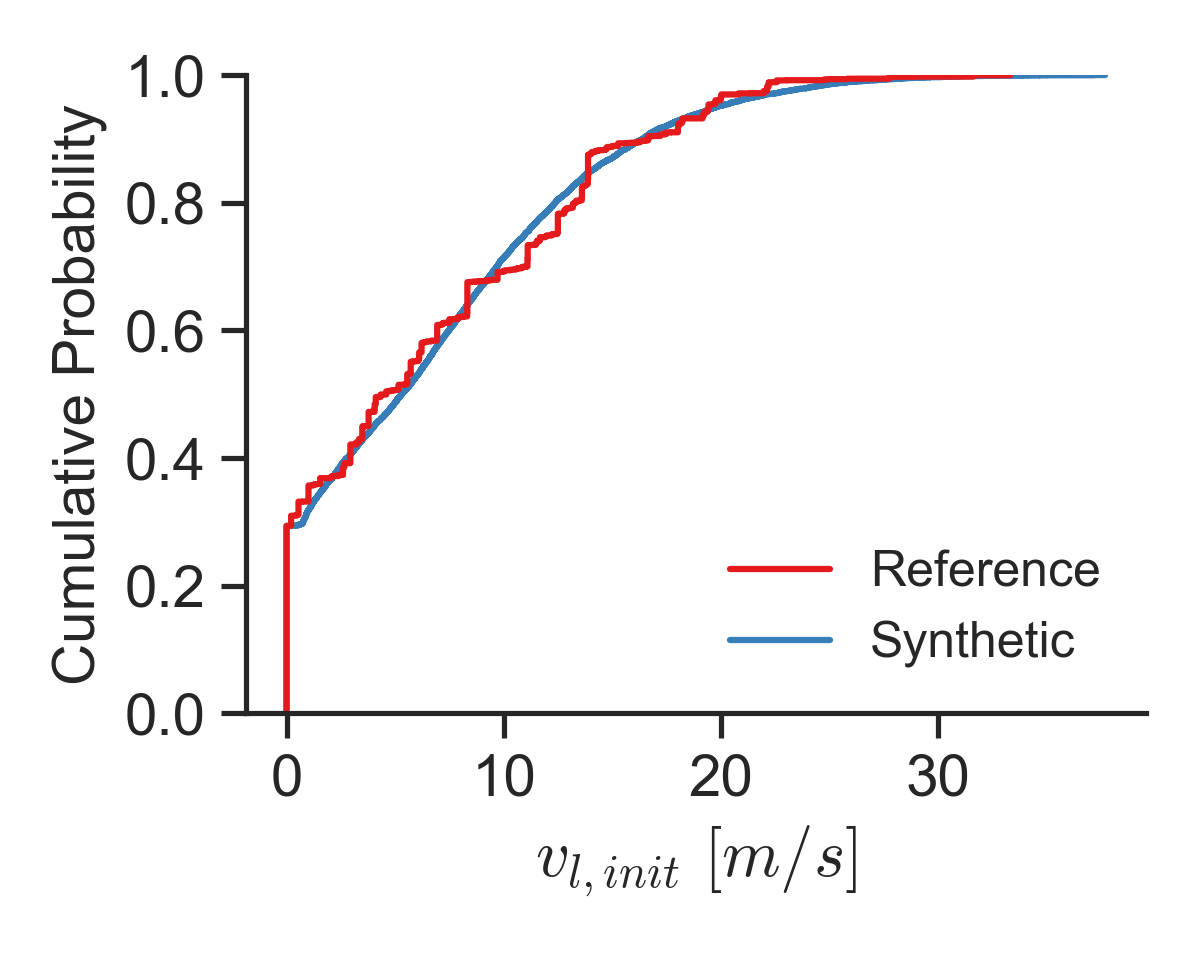}}
    \vfil
    \subfloat[]{\includegraphics[width=0.24\textwidth]{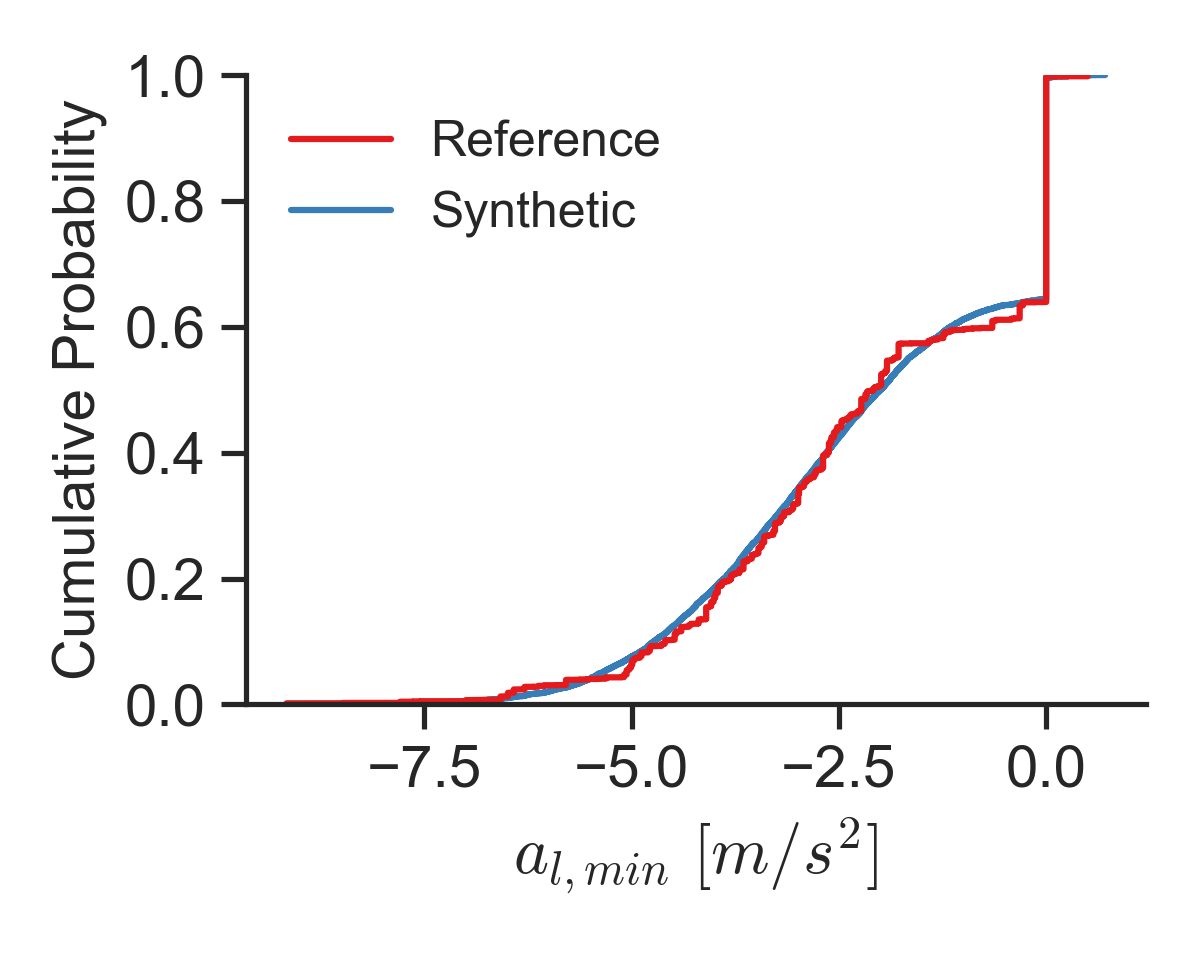}}
    \caption{Weighted CDFs for each of the five parameters in the reference and synthetic datasets: (a) $d_{init}$, (b) $v_{f,init}$, (c) $a_{f,min}$, (d) $v_{l,init}$, and (e) $a_{l,min}$.}
    \label{fig:referencevssynthetic}
\end{figure}

A mixture distribution model was constructed for REF\_b by combining all distribution models built for each sub-dataset according to sub-dataset proportions.
The model was used to create a synthetic dataset containing 10,000 samples of initial states and minimum fitted accelerations of both vehicles (REF\_sb).
Table \ref{tab:comparisonrefandsyn} compares the five parameters for the synthetic and reference datasets (REF\_sb and REF\_b).
Although there are minor differences in the weighted mean and standard deviation (SD) of each parameter between the reference and synthetic datasets, the two datasets underwent a weighted two-sample KS test to test if there are significant differences in each of the five parameters.
The p-values in Table \ref{tab:comparisonrefandsyn} indicate no significant differences between the compared datasets.
Additionally, Fig. \ref{fig:referencevssynthetic} shows that the weighted CDFs for each of the five parameters in the two datasets are well aligned, demonstrating substantial similarities between the two datasets.

\subsection{Simulation} \label{section:simulation}
\begin{table*}[!t] 
\caption{Comparison between the synthetic crash dataset and reference distributions} \label{tab:comparisoncrash}
\centering
\begin{threeparttable}
\begin{tabular}{ccccccc}
\hline
\multirow{2}{*}{Parameter group} & \multirow{2}{*}{Sub-dataset} & \multicolumn{2}{c}{Proportion} & \multirow{2}{*}{Parameter} & \multirow{2}{*}{KS statistic} & \multirow{2}{*}{p-value}\\
\cmidrule(rl){3-4}
& & Reference & Synthetic \\
\hline
\multirow{27}{*}{Initial states and two minimum accelerations} & \multirow{5}{*}{1} & \multirow{5}{*}{0.096} & \multirow{5}{*}{0.092 (-0.004)} & $d_{init}$ & 0.07 & 0.91\\
& & & & $v_{f,init}$ & 0.10 & 0.55\\
& & & & $v_{l,init}$ & 0.09 & 0.66\\
& & & & $a_{l,min}$ & 0.12 & 0.30\\
& & & & Overall & 0.12 & 0.29\\
\cmidrule(rl){2-7}
& \multirow{6}{*}{2} & \multirow{6}{*}{0.309} & \multirow{6}{*}{0.291 (-0.018)} & $d_{init}$ & 0.04 & 0.67\\
& & & & $v_{f,init}$ & 0.06 & 0.22\\
& & & & $a_{f,min}$ & 0.03 & 0.91\\
& & & & $v_{l,init}$ & 0.05 & 0.47\\
& & & & $a_{l,min}$ & 0.03 & 0.82\\
& & & & Overall & 0.07 & 0.08\\
\cmidrule(rl){2-7}
& \multirow{4}{*}{3} & \multirow{4}{*}{0.131} & \multirow{4}{*}{0.139 (+0.008)} & $d_{init}$ & 0.07 & 0.47\\
& & & & $v_{f,init}$ & 0.04 & 1.00\\
& & & & $a_{f,min}$ & 0.05 & 0.90\\
& & & & Overall & 0.08 & 0.37\\
\cmidrule(rl){2-7}
& 4 & 0.163 & 0.163 (+0.000) & $d_{init}$ & 0.50 & 0.00$^*$\\
\cmidrule(rl){2-7}
& \multirow{4}{*}{5} & \multirow{4}{*}{0.126} & \multirow{4}{*}{0.134 (+0.008)} & $d_{init}$ & 0.07 & 0.59\\
& & & & $v_{f,init}$ & 0.09 & 0.25\\
& & & & $v_{l,init}$ & 0.08 & 0.37\\
& & & & $a_{l,min}$ & 0.16 & 0.00$^*$\\
& & & & Overall & 0.10 & 0.11\\
\cmidrule(rl){2-7}
& \multirow{6}{*}{6} & \multirow{6}{*}{0.175} & \multirow{6}{*}{0.182 (+0.007)} & $d_{init}$ & 0.09 & 0.10\\
& & & & $v_{f,init}$ & 0.08 & 0.12\\
& & & & $a_{f,min}$ & 0.04 & 0.95\\
& & & & $v_{l,init}$ & 0.07 & 0.25\\
& & & & $a_{l,min}$ & 0.08 & 0.14\\
& & & & Overall & 0.08 & 0.14\\
\cmidrule(rl){1-7}
\multirow{32}{*}{Lead-vehicle speed profile} & 1 & 0.255 & 0.301 (+0.046) & \textemdash & \textemdash & \textemdash\\
\cmidrule(rl){2-7}
& \multirow{3}{*}{2} & \multirow{3}{*}{0.105} & \multirow{3}{*}{0.090 (-0.015)} & $v_{l,init}$ & 0.11 & 0.31\\
& & & & $a_1$ & 0.12 & 0.17\\
& & & & Overall & 0.13 & 0.13\\
\cmidrule(rl){2-7}
& \multirow{4}{*}{3} & \multirow{4}{*}{0.102} & \multirow{4}{*}{0.080 (-0.022)} & $v_{l,init}$ & 0.05 & 0.97\\
& & & & $a_1$ & 0.07 & 0.64\\
& & & & $\tau_s$ & 0.05 & 0.98\\
& & & & Overall & 0.05 & 0.93\\
\cmidrule(rl){2-7}
& \multirow{7}{*}{4} & \multirow{7}{*}{0.157} & \multirow{7}{*}{0.146 (-0.011)} & $v_{l,init}$ & 0.08 & 0.13\\
& & & & $a_1$ & 0.03 & 0.98\\
& & & & $a_2$ & 0.07 & 0.34\\
& & & & $\tau_s$ & 0.04 & 0.95\\
& & & & $\tau_1$ & 0.07 & 0.37\\
& & & & $\tau_2$ & 0.04 & 0.95\\
& & & & Overall & 0.08 & 0.13\\
\cmidrule(rl){2-7}
& \multirow{5}{*}{5} & \multirow{5}{*}{0.046} & \multirow{5}{*}{0.048 (+0.002)} & $v_{l,init}$ & 0.07 & 0.98\\
& & & & $a_1$ & 0.17 & 0.12\\
& & & & $a_2$ & 0.16 & 0.16\\
& & & & $\tau_1$ & 0.15 & 0.27\\
& & & & Overall & 0.09 & 0.80\\
\cmidrule(rl){2-7}
& \multirow{5}{*}{6} & \multirow{5}{*}{0.133} & \multirow{5}{*}{0.134 (+0.001)} & $v_{l,init}$ & 0.08 & 0.23\\
& & & & $a_1$ & 0.04 & 0.87\\
& & & & $a_2$ & 0.04 & 0.93\\
& & & & $\tau_1$ & 0.03 & 0.98\\
& & & & Overall & 0.08 & 0.20\\
\cmidrule(rl){2-7}
& \multirow{7}{*}{7} & \multirow{7}{*}{0.202} & \multirow{7}{*}{0.201 (-0.001)} & $v_{l,init}$ & 0.09 & 0.11\\
& & & & $a_1$ & 0.10 & 0.07\\
& & & & $a_2$ & 0.05 & 0.83\\
& & & & $\tau_s$ & 0.10 & 0.10\\
& & & & $\tau_1$ & 0.04 & 0.95\\
& & & & $\tau_2$ & 0.07 & 0.35\\
& & & & Overall & 0.10 & 0.07\\
\cmidrule(rl){1-7}
\multirow{3}{*}{The remaining three parameters} & \multirow{3}{*}{\textemdash} & \multirow{2}{*}{\textemdash} & \multirow{2}{*}{\textemdash} & $T$ & 0.05 & 0.86\\
& & & & $t_g$ & 0.05 & 0.75\\
& & 0.092 & 0.096 (+0.004) & $t_a$ & 0.07 & 0.82\\
\hline
\end{tabular}
\begin{tablenotes}
\RaggedRight
\item $^*$ The difference is significant at the 0.05 significance level.
\end{tablenotes}
\end{threeparttable}
\end{table*}

\begin{figure}[!t]
    \centering
    \subfloat[]{\includegraphics[width=0.24\textwidth]{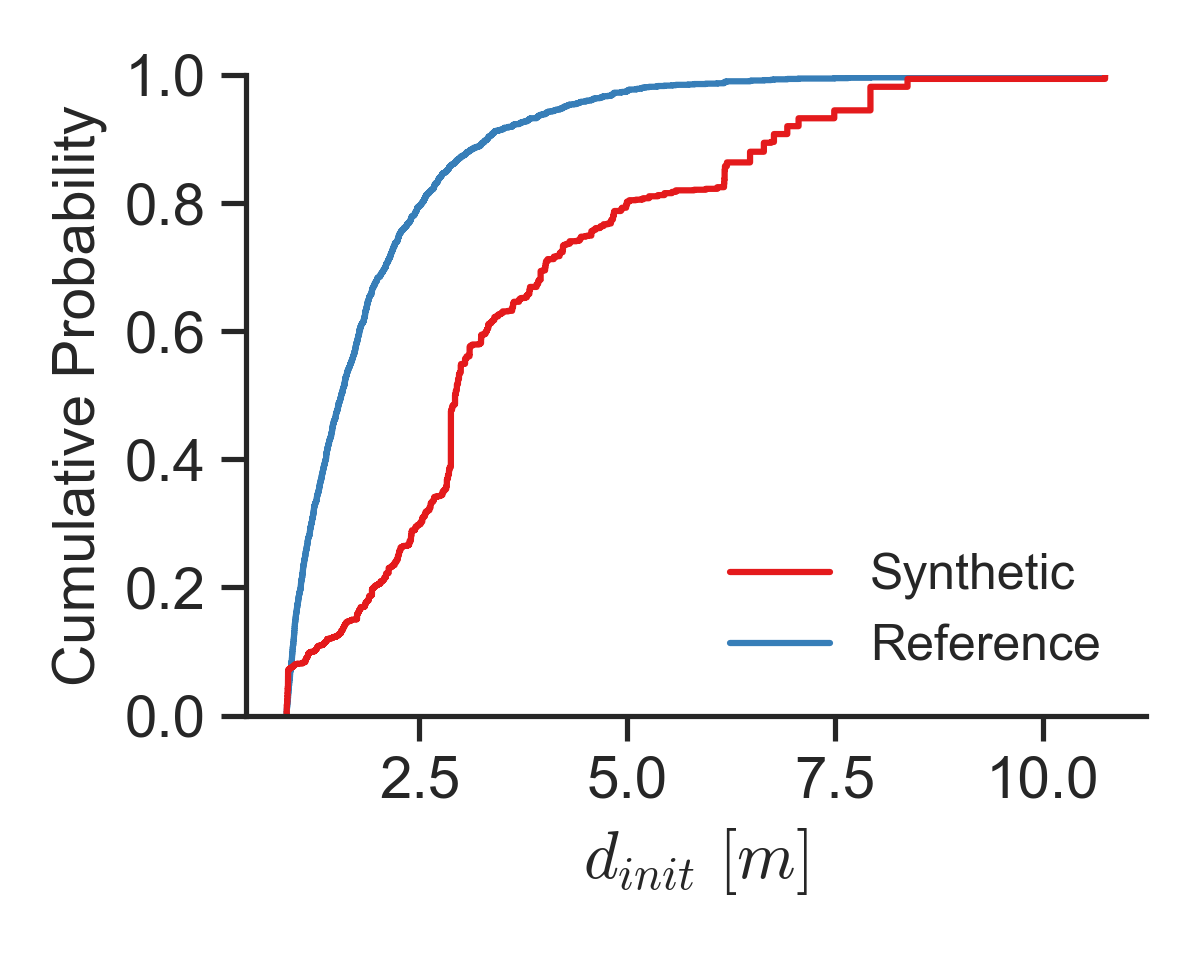}}
    \hfil
    \subfloat[]{\includegraphics[width=0.24\textwidth]{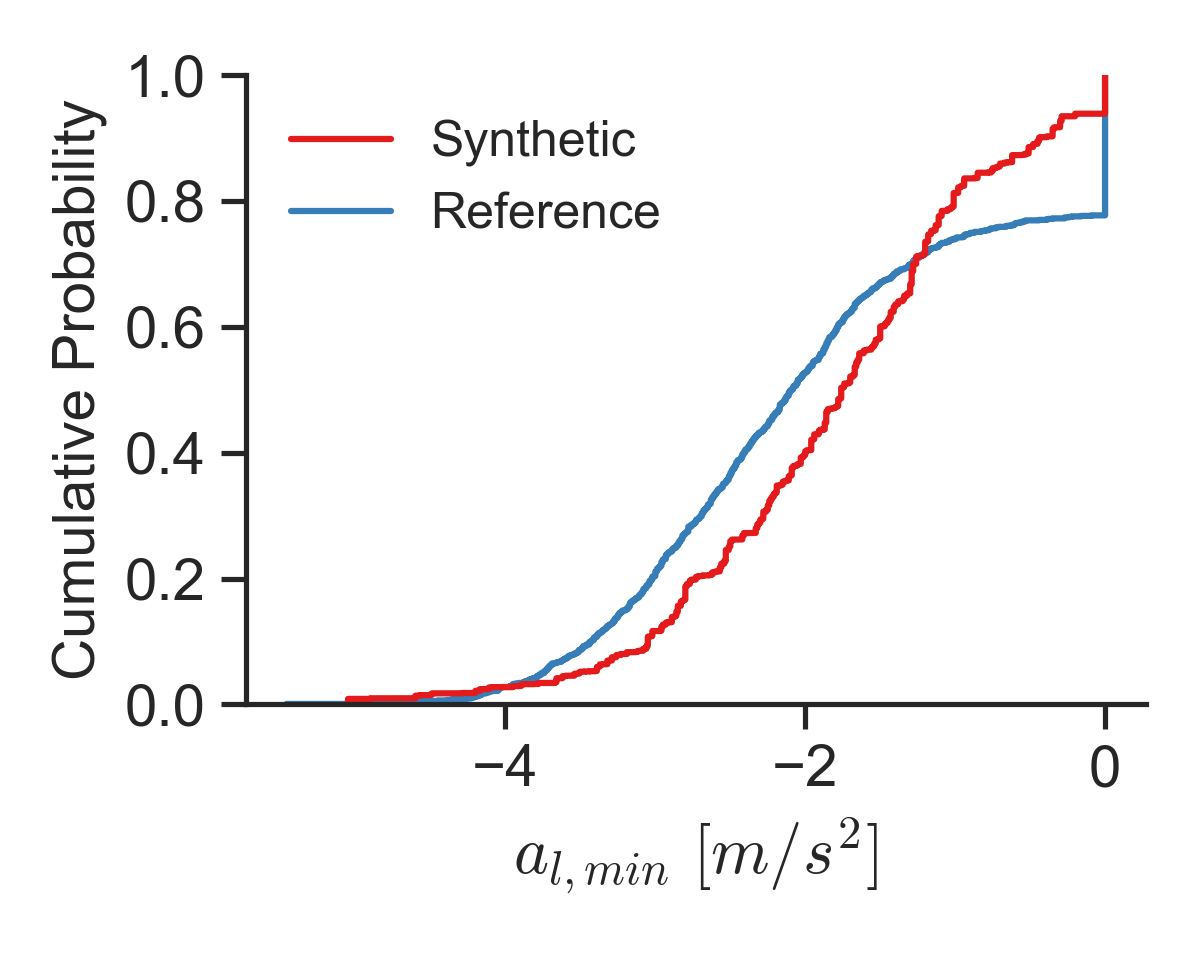}}
    \caption{CDFs of two parameters for the weighted synthetic crash dataset and the corresponding reference distributions: a) $d_{init}$ in REF\_b sub-dataset 4, and b) $a_{l,min}$ in REF\_b sub-dataset 5.}
    \label{fig:two_sig_tests}
\end{figure}

\begin{figure}[!t]
    \centering
    \subfloat[]{\includegraphics[width=0.24\textwidth]{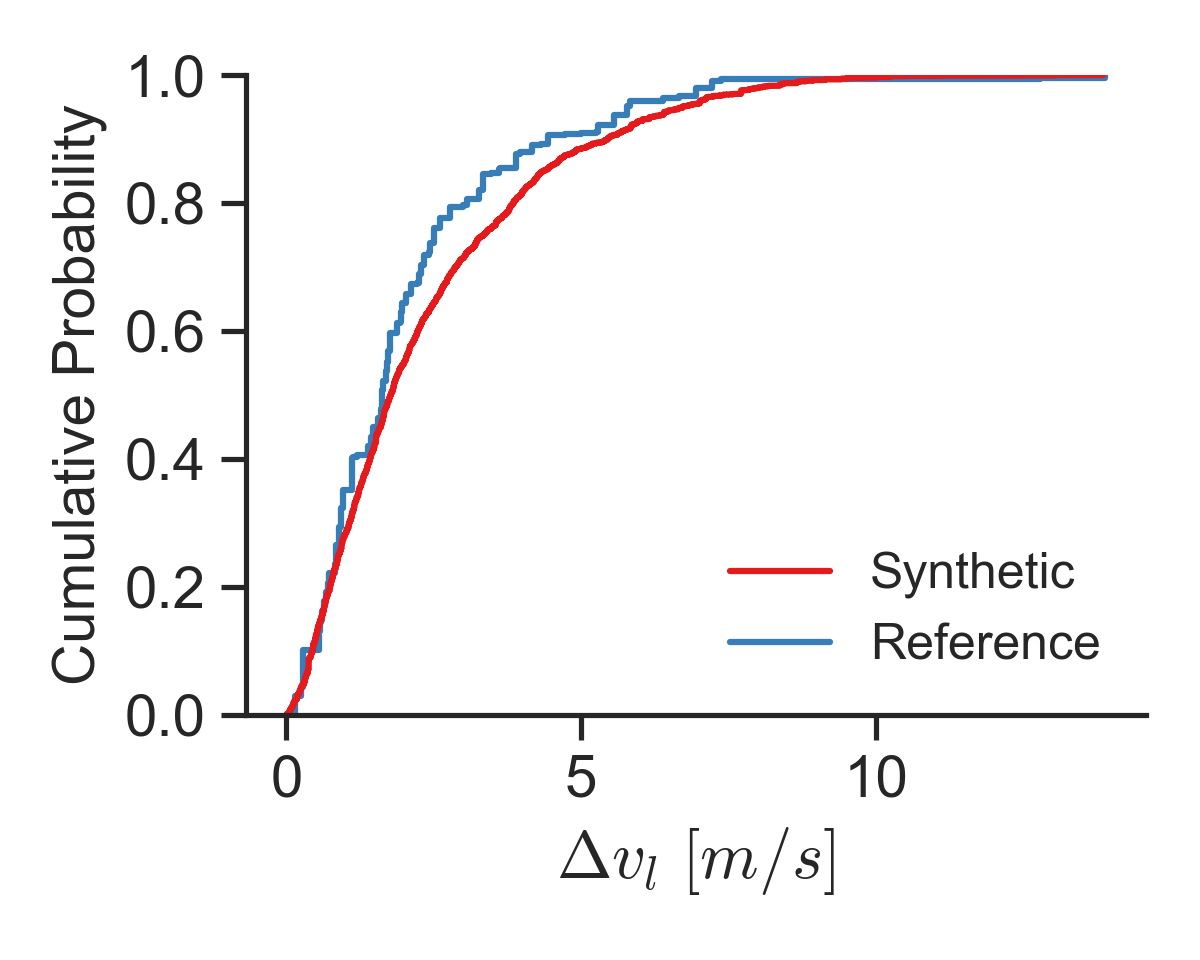}}
    \hfil
    \subfloat[]{\includegraphics[width=0.24\textwidth]{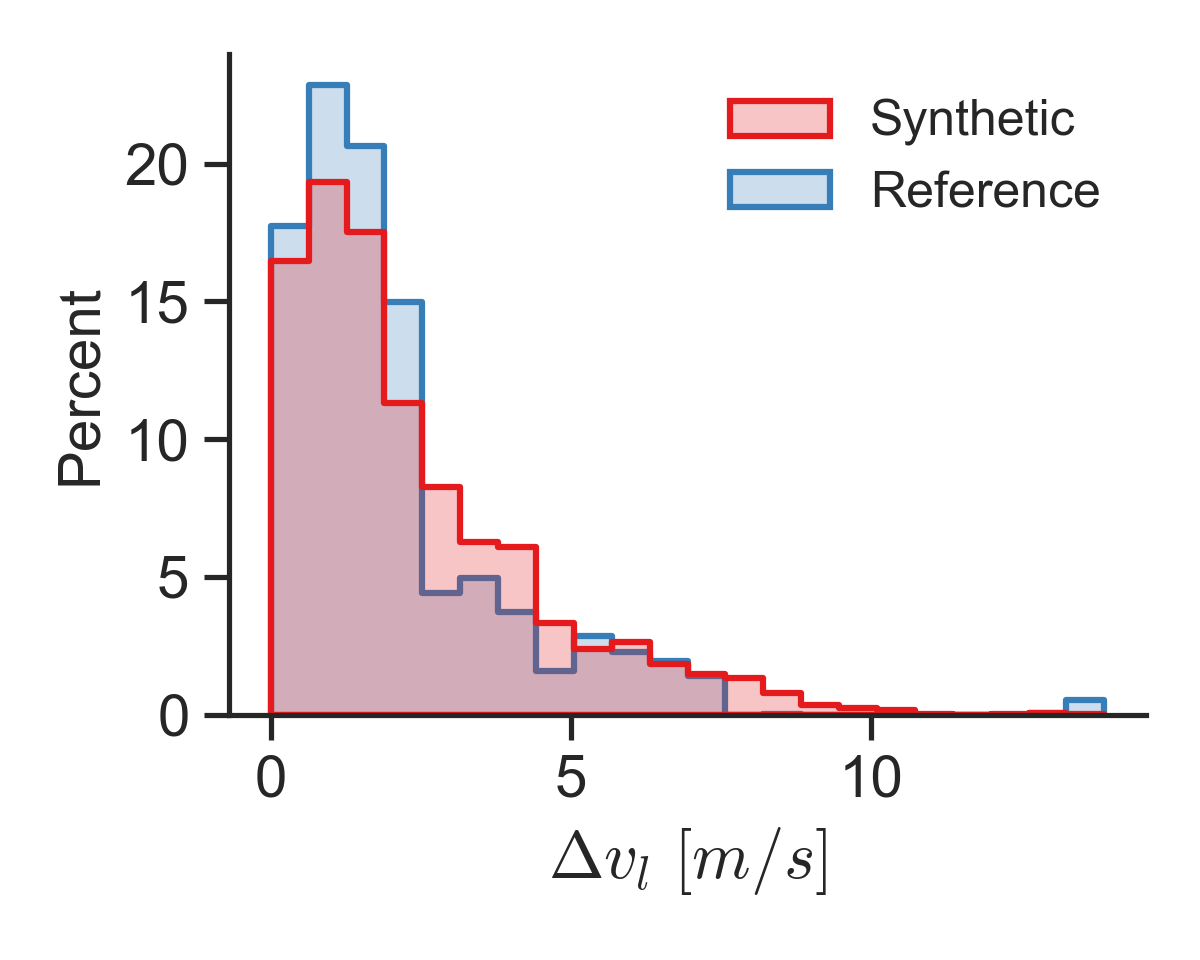}}
    \caption{Comparison between the synthetic crash dataset and the reference dataset for $\Delta v_l$: a) CDFs, and b) histograms. The weighted two-sample KS test results between the distributions of $\Delta v_l$ in the synthetic and reference datasets: sample size of the synthetic dataset $n$ = 5,000; sample size of the reference dataset $n_{r}$ = 130; statistic = 0.12, p-value = 0.27.}
    \label{fig:lead_delta_v}
\end{figure}

We ran the matching algorithm for the two reference datasets (REF\_sl and REF\_sb) and selected 5,000 valid simulations to create a synthetic rear-end crash dataset.
The sample weighting method outlined in Section \ref{section:ipf} was used to assign a weight for each sample in the synthetic crash dataset.

Table \ref{tab:comparisoncrash} shows the results of the weighted two-sample KS tests between the synthetic crash dataset and the reference datasets (REF\_sl, REF\_sb, and the reference marginal distributions of the remaining three parameters $T$, $t_g$ and $t_a$).
There were, in total, 61 tests.
At a significance level of 0.05, approximately three ($\approx 61 \times 0.05$) tests are likely to be incorrectly tested to be significant when they should not be.
In our situation, two (less than three) tests show a significant difference at the 0.05 significance level.
We looked into those two tests (i.e., the marginal distribution test for $d_{init}$ in REF\_b S4 and the marginal distribution test for $a_{l, min}$ in REF\_b S5) and analyzed the possible cause.

Fig. \ref{fig:two_sig_tests} shows that the CDFs for the two tests have significant differences.
Specifically, Fig. \ref{fig:two_sig_tests}(a) shows that fewer valid simulations have a small initial distance ($d_{init}$), and Fig. \ref{fig:two_sig_tests}(b) shows that there are fewer valid simulations in which the lead vehicle did not brake at all during the event (i.e., in which the minimum acceleration $a_{l, min} =$ 0 m/s$^2$).
These differences can be explained by a limitation in the following vehicle's acceleration model, the modified IDM; it cannot imitate accelerations as aggressive as those in the real world.
For instance, in the cases in Fig. \ref{fig:two_sig_tests}(a), the following vehicle was stationary and would not start to move forward unless the distance to the lead vehicle was large enough.
(See Section \ref{section:idm} for more discussion.)

Fig. \ref{fig:lead_delta_v} compares the weighted synthetic crash and reference datasets for $\Delta v_l$.
The results of the weighted two-sample KS test indicate that there is no significant difference between the two datasets.
Although the CDFs in Fig. \ref{fig:lead_delta_v}(a) are substantially similar, the histograms in Fig. \ref{fig:lead_delta_v}(b) illustrate a higher proportion of low values in the reference distribution of $\Delta v_l$.
This discrepancy could be due to the way that $\Delta v_l$ was estimated for SHRP2 crashes (in REF\_l).
During the impact in a rear-end crash, the lead vehicle has a rapid speed increase followed by a swift speed decrease.
For these SHRP2 crashes, $\Delta v_l$ was calculated as the difference between the post- and pre-impact lead-vehicle speed.
The frequency of the lead-vehicle speed signal was 10 Hz; at such a low frequency, the speed signal is unlikely to accurately capture the true post-impact (peak) speed, thereby resulting in an underestimation of $\Delta v_l$.

\section{Discussion and Conclusions} \label{section:discussion}
Unlike other studies focusing mainly on injury-involved or policed-reported rear-end crashes \cite{bareiss2019crash, gambi2019generating, wang2022autonomous}, this study created a representative synthetic rear-end crash dataset encompassing the full severity range, from physical contact to high-severity levels.

The process of generating synthetic rear-end crash scenarios consists of three main steps: 1) parameterizing the rear-end crashes through modeling the following and lead vehicles, 2) building reference datasets from the parameterized crash data, and 3) generating representative synthetic crash scenarios.

In the first step, a following-vehicle behavior model was developed by combining two existing driver models.
The model also included the potential for generating 'abnormal' driver acceleration behavior, a phenomenon observed in 9.2\% of all crashes.
Using this model and the lead-vehicle kinematics model (created in a previous study \cite{wu2024modeling}), we sought to emulate vehicle behaviors that are as similar as possible to those in real-world rear-end crash scenarios.
Combining the two vehicle models and the initial states of rear-end crash scenarios created a twelve-dimensional vector representing a rear-end crash.

In the second step, parameterized crash data from multiple crash datasets were combined and weighted to create a reference dataset of the initial states (and minimum fitted accelerations of both vehicles) (REF\_b).
A synthetic dataset containing these data (REF\_sb) was then created by sampling from the distribution model built for REF\_b.

At last, simulations were conducted using the following-vehicle behavior model and the two synthetic datasets, REF\_sb and REF\_sl (a representative synthetic rear-end crash lead-vehicle speed profile dataset created in a previous study \cite{wu2024modeling}).
valid simulations were gathered and weighted using an IPF-based weighting algorithm to create a representative synthetic rear-end crash dataset.

In terms of validation, a more comprehensive validation process than in other studies was conducted.
Non-parametric statistic tests were implemented for the marginal distributions—not only for the crash outcomes (e.g., Delta-v of the lead vehicle) but also for each of the twelve parameters.

\subsection{Contributions} \label{section:application}
\begin{figure}[!t]
    \centering
    \includegraphics[width=0.35\textwidth]{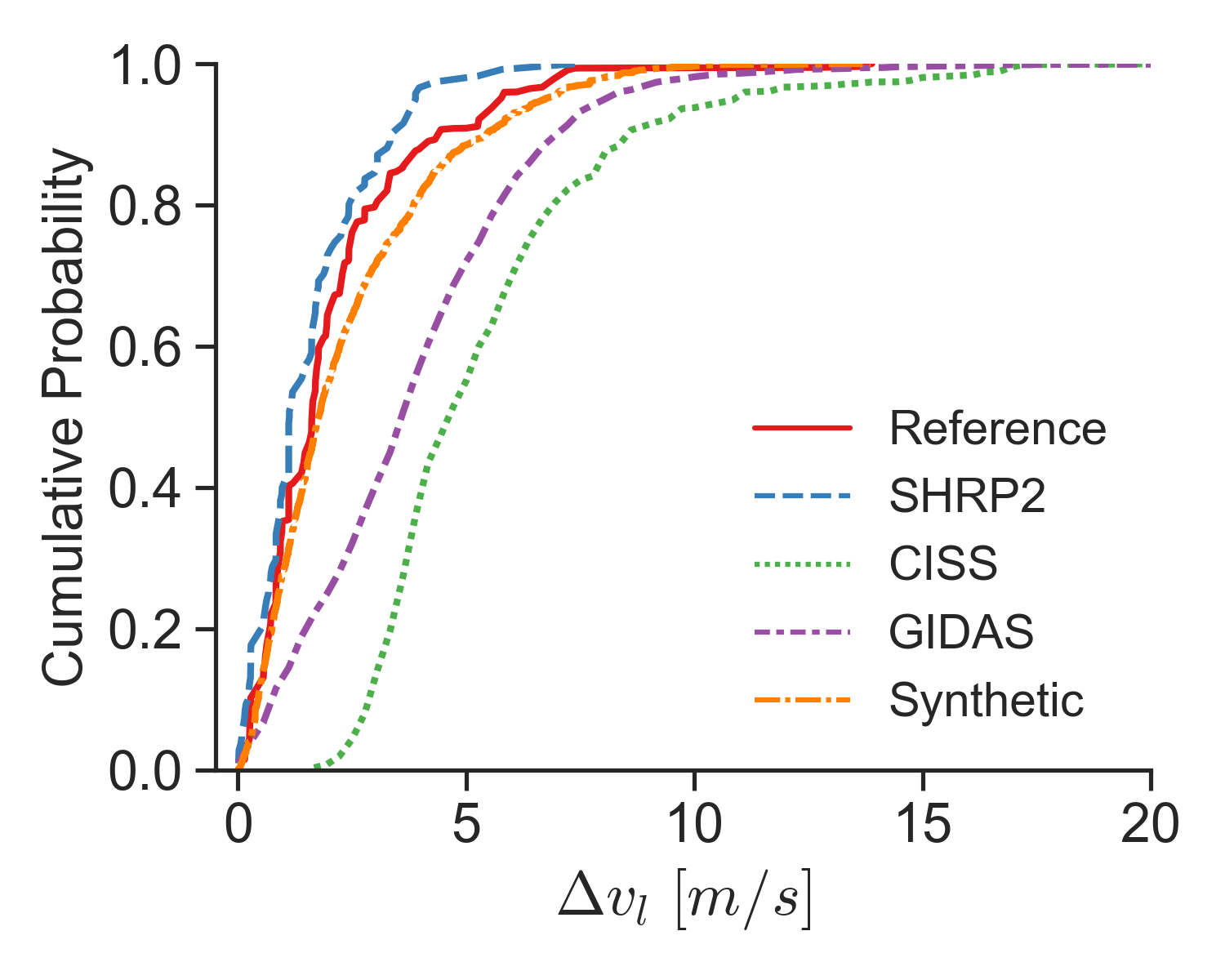}
    \caption{CDF curves for $\Delta v_l$ in rear-end crashes among various datasets.}
    \label{fig:comp_delta_v_all}
\end{figure}

This study created a representative dataset of synthetic rear-end crashes covering the full range of severity levels.
This dataset can be used for safety assessments of ADAS and ADS and as a benchmark when evaluating the representativeness of scenarios generated through other methods (such as traffic-simulation-based and machine-learning-based).
Fig. \ref{fig:comp_delta_v_all} compares the CDF curves of the lead vehicle's Delta-v in rear-end crashes from various datasets.
Compared to the reference dataset, the CISS and GIDAS datasets are biased towards severe crashes.
Although the SHRP2 dataset is similar to the reference dataset, it lacks high-severity cases (the maximum $\Delta v_l$ in the SHRP2 dataset is 7.4 m/s).
In contrast, the synthetic crash dataset, mirroring the reference dataset, encompasses crashes with $\Delta v_l$ reaching 13.8 m/s.
As mentioned in Section \ref{section:simulation}, the synthetic dataset exhibits a strong resemblance to the reference dataset regarding the distribution of Delta-v of the lead vehicle.

The methodological contributions of this study mainly lie in the data combination (including sample weighting) methods.
In the study, none of the available crash datasets contain all necessary signals without substantial bias; this shortcoming is common in data-driven studies.
We used a set of methods to combine and weight data from multiple crash datasets to mitigate the biases.
Among these methods, the KNN sample weighting method is particularly noteworthy because, unlike conventional post-stratification methods, it can be used to weight biased data to match a reference dataset even when omitted strata exist.
In the future, these data combination methods can be applied when creating a multivariate joint distribution is needed and the only datasets available contain biased data or incomplete signals.

\subsection{Limitations and future work} \label{section:idm}
Our previous study \cite{wu2024modeling} relied on pre-crash data from the United States to establish the reference dataset of lead-vehicle kinematics.
However, we faced a shortage of crash data involving both vehicles: SHRP2\_b contained only 37 samples.
Therefore, we had to use the available data from the GIDAS-PCM dataset (PCM\_b), even though it was from Germany.
We assumed that rear-end crashes in the US and Germany have similar mechanisms, although their distributions may differ.
Moreover, during data combination, the KNN sample weighting method was applied to reduce bias in the merged raw data (SHRP2\_b and PCM\_b) so that the weighted data could match the reference dataset created using the US crash data.
In addition, in Sub-step 4 of the data combination step, we used the optimal pairing results of samples from the two reference datasets REF\_l and REF\_f as an approximation for the reference dataset of the four parameters ($v_{f,init}$).
This compromise was necessary because no such reference dataset is available (at least not to us).

The modified IDM was used to simulate the acceleration behavior of the following vehicle.
However, the model was designed and calibrated to replicate naturalistic car-following behaviors rather than crashes.
As a result, this model cannot accurately mimic highly aggressive accelerations, which often occur in real-world situations.
Thus, as mentioned in Section \ref{section:simulation}, a subset of crash scenarios in REF\_b sub-datasets S4-5 was missing.
To address this limitation, future research should try calibrating the modified IDM acceleration model using pre-crash data for scenario generation.

As mentioned in Section \ref{section:simulation}, the low-frequency speed signal used to estimate the Delta-v of the lead vehicle in the SHRP2 dataset might underestimate the true value.
Future research should either find a better estimation method or use data with a higher frequency.

\section*{Acknowledgments}
This research was funded by FFI Vinnova, a Swedish governmental agency for innovation, as part of the project Improved quantitative driver behavior models and safety assessment methods for ADAS and AD (QUADRIS: nr. 2020-05156).
The SHRP2 data used in this study has the identifier DOI SHRP2-DUL-16-172 and was made available to us by the Virginia Tech Transportation Institute (VTTI) under a Data License Agreement.
The findings and conclusions of this paper are those of the authors and do not necessarily represent the views of VTTI, the Transportation Research Board (TRB), or the National Academies.
The authors wish to thank Mikael Ljung Aust at Volvo Cars Safety Centre for reviewing the manuscript.

\appendices
\section{KNN sample weighting algorithm} \label{appendix:KNN}
Since raw distributions from datasets can often be biased with respect to one or more parameters, the sample weighting process aims to assign weights to samples in raw distributions $\{\textbf{X}_i | i \in [1,n]\}$ (where $\textbf{X}_{i} = [x_1^{(i)},...,x_K^{(i)}]^T$) so that the weighted data matches the known reference distribution of a subset of parameters $\Tilde{\Phi}(x_1,...,x_m)\ (m < K)$.

Post-stratification weighting \cite{holt1979post} is one possibility. It is a statistical technique commonly used in survey research to reduce bias and improve the accuracy of population estimates.
It involves dividing the target population into strata based on certain characteristics or variables, collecting data within each stratum, and then assigning weights to the observations based on the target population distribution within each stratum.
Typically, binning is used to create strata when the variables are continuous. The weight for observations in each stratum is the target population total divided by the number of observations in the stratum.
In our situation, the raw samples (i.e., observations) $\{\textbf{X}_i | i \in [1,n]\}$ should be grouped into discrete bins designed based on the known reference distribution (i.e., target population) $\Tilde{\Phi}(x_1,...,x_m)$.
However, this method assumes that no strata are omitted.
In other words, observations within all bins must correspond to the reference distribution.
In our case, omitted strata did exist in the combined data.
One possible cause could be the bias in the combined data.
For instance, datasets sourced from CISS and GIDAS-PCM contain only severe crashes.
Therefore, a novel method, the k-nearest neighbors (KNN) sample weighting method, was proposed to handle this issue.

The KNN sample weighting method can be seen as a post-stratification weighting method with a dynamic binning strategy.
Each sample extracted from the known reference distribution carries a weight of one.
For each extracted sample, the k-nearest raw samples are grouped into one bin to share the weight (see Step 3c in the following algorithm).
It is also worth mentioning that samples that have never been selected as the nearest neighbors of any extracted sample will have a weight of zero.

\begin{algorithm}
\caption{KNN sample weighting algorithm.} \label{alg:alg1}
\begin{algorithmic}
\STATE 
\STATE Set $w_{i} = 0\ \forall\ i \in [1,n]$
\STATE Generate $N$ samples from $\Tilde{\Phi}(x_1,...,x_m)$: $\{[\Tilde{x}_1^{(j)},...,\Tilde{x}_m^{(j)}]^T |\ j \in [1,N]\}$
\STATE For $j = 1$ to $N$:
\STATE \hspace{0.5cm} $d_j^{(i)} = \sqrt{\sum_{p=1}^m (\Tilde{x}_p^{\prime(j)} - x_p^{\prime(i)})^2}\ \forall\ i \in [1,n]$
\STATE \hspace{0.5cm} $\omega_j^{(i)} = 1/d_j^{(i)}$ if all$(d_j^{(i)} > 0)$ else $I_{\{d_j^{(i)} = 0\}}\ \forall\ i \in [1,n]$
\STATE \hspace{0.5cm} $H = \arg \underset{i}{\max}(\{\omega_j^{(i)} |\ i \in [1,n]\}, k)$
\STATE \hspace{0.5cm} $w_{h_l} \gets w_{h_l} + \frac{\omega_j^{(h_l)}}{\sum_{l=1}^k \omega_j^{(h_l)}}$ $\forall\ h_{l} \in H$
\STATE $w_{i} \gets \frac{w_{i}}{\sum_{i=1}^n w_{i}} \sum_{i=1}^n I_{\{w_{i} > 0\}}\ \forall\ i \in [1,n]$
\end{algorithmic}
\end{algorithm}

As shown in Algorithm \ref{alg:alg1}, the KNN sample weighting method contains four main steps.
\begin{enumerate}
    \item Set the initial sample weight for each raw sample to zero: $w_i = 0\ \forall\ i \in [1,n]$.
    \item Sample $N$ samples from the known reference distribution $\Tilde{\Phi}(x_1,...,x_m)$.
    \item For any generated sample $\Tilde{\textbf{X}}_j$:
    \begin{enumerate}
        \item Compute the Euclidean distance between $\Tilde{\textbf{X}}_j$ and $\textbf{X}_i$, $d_j^{(i)}$, for all $i \in [1,n]$. ($\Tilde{x}_p^{\prime(j)}$ and $x_p^{\prime(i)}$ are the standardized values of $\Tilde{x}_p^{(j)}$ and $x_p^{(i)}$, respectively.)
        \item Compute the distributing weight of the raw sample $\textbf{X}_i$ for $\Tilde{\textbf{X}}_j$, $\omega_j^{(i)}$, for all $i \in [1,n]$. (A smaller Euclidean distance correlates to a higher distributing weight.)
        \item Distribute a weight value of one among the top k raw samples with the highest distributing weights ($\{\textbf{X}_{h_l} | h_l \in H\}$).
    \end{enumerate}
    \item Scale the weights so that $\sum_{i=1}^{n} w_i = n$.
\end{enumerate}
The value of k is determined by minimizing the loss, $\sum_{l = 1}^m s_l^{(k)}$, where $s_l^{(k)}$ is the KS statistic for $x_l$ conditioned on k computed with the weighted two-sample KS tests between the weighted $x_l$ data and the reference data $\{\Tilde{x}_l^{(j)} | j \in [1,N]\}$.

\section{Pairing algorithm} \label{section:pairingalgorithm}

\begin{figure}[!t]
    \centering
    \subfloat[]{\includegraphics[width=0.24\textwidth]{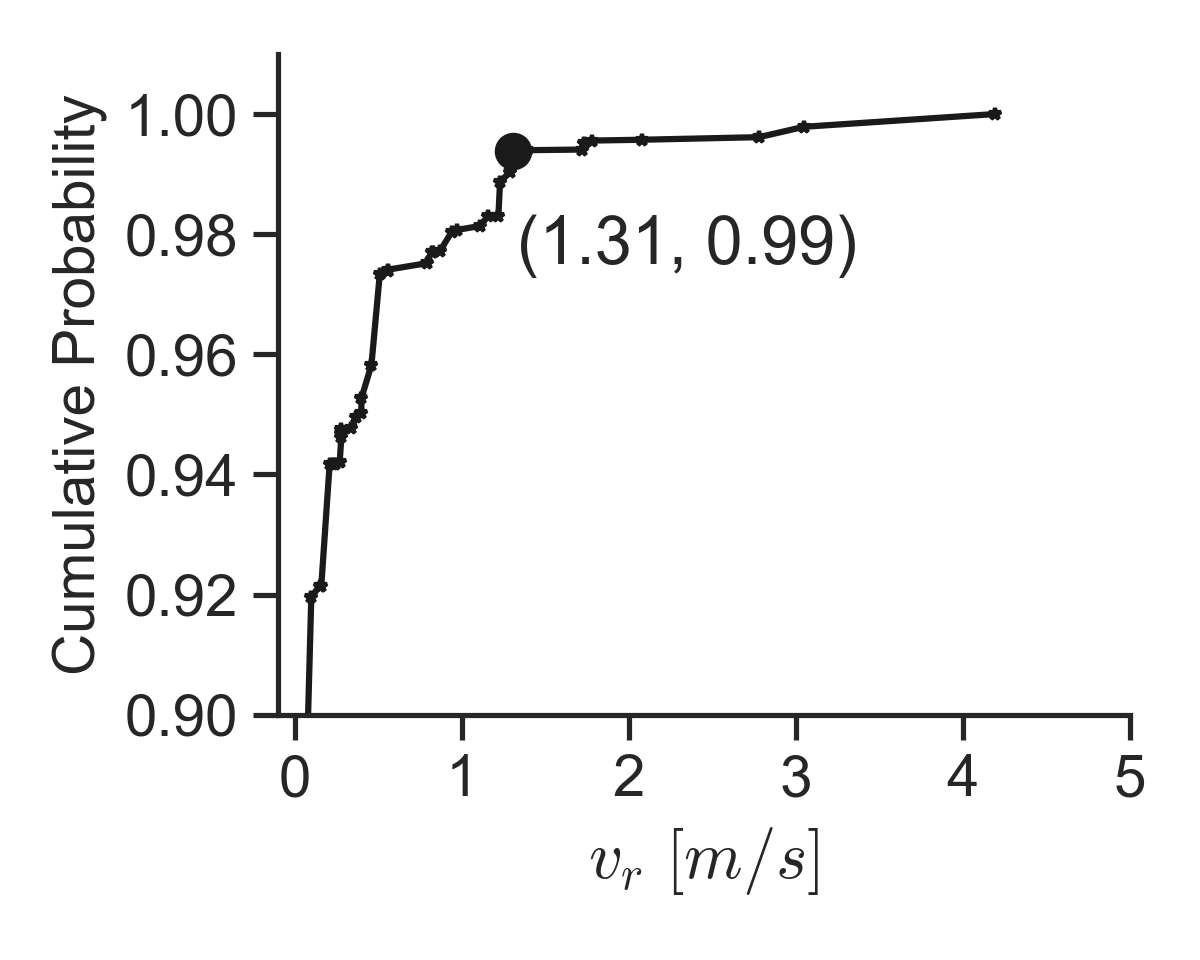}}
    \hfil
    \subfloat[]{\includegraphics[width=0.24\textwidth]{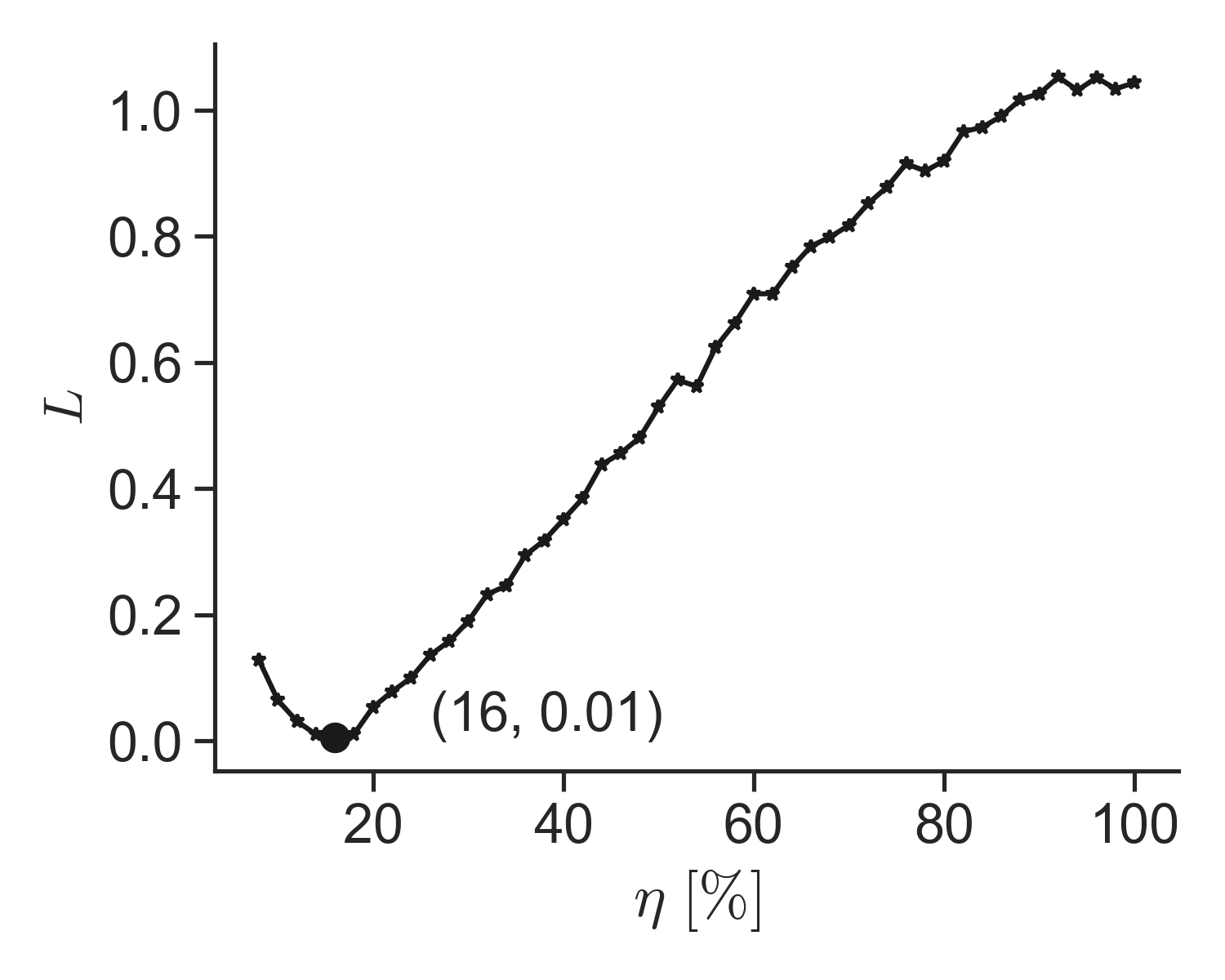}}
    \caption{Pairing algorithm parameter setting: a) CDF of the lead vehicle's initial relative speed ($v_{l,init} - v_{f,init}$), and b) Loss of the pairing as a function of $\eta$.}
    \label{fig:pairing_parameter_setting}
\end{figure}

The pairing algorithm was mainly based on the relationship between the initial speeds of both vehicles ($v_{f,init}$ and $v_{l,init}$) because these two parameters have a stronger correlation than the one between the following vehicle's initial speed ($v_{f,init}$) and the lead vehicle's minimum fitted acceleration ($a_{l,min}$).
We also observed that $v_{l,init}$ is no larger than $v_{f,init}$ in most cases.

As shown in Algorithm \ref{alg:alg2}, $\eta$ is the percentage of randomly selected samples from set $A$, $v_{r,thd} (= v_{l,init} - v_{f,init})$ is the threshold of the lead vehicle's initial relative speed (m/s), $f_{v_{l,init}}$ is the probability density function of $v_{l,init}$ (estimated using the marginal distribution of $v_{l,init}$ from REF\_l), $L$ is the loss, and $\eta^*$ is the optimal $\eta$ with minimum loss.
A smaller $\eta$ leads to a stronger correlation between $v_{f,init}$ and $v_{l,init}$ (as for $v_{f,init}$ and $a_{l,min}$).
$v_{r,thd}$ was set to 1.31 m/s, the elbow point in its CDF curve: see Fig. \ref{fig:pairing_parameter_setting}(a).
$\eta^*$ was set to 0.16 with a minimum loss of 0.01: see Fig. \ref{fig:pairing_parameter_setting}(b).

\begin{algorithm*}
\caption{Pairing algorithm.}\label{alg:alg2}
\begin{algorithmic}
\STATE 
\STATE Sample with the replacement of $N$ samples from $\Tilde{\Phi}(v_{f,init}, \Delta v_l)$ and $\Tilde{\Phi}(v_{l,init}, a_{l,min})$, respectively:
\STATE \hspace{0.5cm} $A = \{\textbf{A}_i |\ i \in [1, N]\}$, $B = \{\textbf{B}_j |\ j \in [1, N]\}$, where $\textbf{A}_i = [\Tilde{v}_{f,init}^{(i)}, \Tilde{\Delta v}_{l}^{(i)}]^T$ and $\textbf{B}_j = [\Tilde{v}_{l,init}^{(j)}, \Tilde{a}_{l,min}^{(j)}]^T$
\STATE For $\eta = 0:2:100\ [\%]$:
\STATE \hspace{0.5cm} Select randomly $\eta$ samples out of $A$
\STATE \hspace{0.5cm} Create a copy of $B$: $\hat{B} = B$
\STATE \hspace{0.5cm} For each selected sample $\textbf{A}_i$, select its corresponding $\hat{\textbf{B}}_i$ from $\hat{B}$:
\STATE \hspace{1cm} If $\Tilde{v}_{l,init}^{(j)} > \Tilde{v}_{f,init}^{(i)} + v_{r,thd}\ \forall\ \textbf{B}_j \in \hat{B}$:
\STATE \hspace{1.5cm} Select the one with the minimum $\Tilde{v}_{l,init}$
\STATE \hspace{1cm} Else:
\STATE \hspace{1.5cm} Select based on $f_{v_{l,init}}(v_{l,init}|v_{l,init} \leq \Tilde{v}_{f,init}^{(i)} + v_{r,thd})$
\STATE \hspace{1cm} Drop $\hat{\textbf{B}}_i$ from $\hat{B}$
\STATE \hspace{0.5cm} Sort the remaining samples in $A$ in ascending order of $\Tilde{v}_{f,init}$: $\{\textbf{A}_p |\ p \in P\}$
\STATE \hspace{0.5cm} For each $\textbf{A}_p$, select its corresponding $\hat{\textbf{B}}_p$ from $B$ (same as for $\textbf{A}_i$)
\STATE \hspace{0.5cm} Compute correlations for paired samples:
\STATE \hspace{1cm} $r(v_{f,init}, v_{l,init})$, $r(v_{f,init}, a_{l,min})$, $r(\Delta v_l, v_{l,init})$, and $r(\Delta v_l, a_{l,min})$
\STATE \hspace{0.5cm} Compute the loss for paired samples:
\STATE \hspace{1cm} If $|r(\Delta v_l, v_{l,init})| < 0.3\ \&\ |r(\Delta v_l, a_{l,min})| < 0.3$:
\STATE \hspace{1.5cm} $L(\eta) = |r(v_{f,init}, v_{l,init}) - \Tilde{r}(v_{f,init}, v_{l,init})| + |r(v_{f,init}, a_{l,min}) - \Tilde{r}(v_{f,init}, a_{l,min})|$
\STATE \hspace{1cm} Else:
\STATE \hspace{1.5cm} $L(\eta) = +\infty$
\STATE $\eta^* = \arg \underset{x}{\min} (L)$
\end{algorithmic}
\end{algorithm*}

\section{Matching algorithm} \label{section:matchingalgorithm}
The algorithm is shown in Algorithm \ref{alg:alg3}.
$n_{iter}$ is the current number of iterations in terms of lead-speed profile, $n_{iter,max}$ is the maximum number of iterations (set to 10 in this study), $d_j^{(i)}$ is the Euclidean distance between $\textbf{V}_j$ and $\textbf{U}_i$, $W$ is the set of candidates in $U$ that can pair with $\textbf{V}_j$.
$\bar{T}$ and $\bar{T}_g$ are the sets of percentiles $\{\pi_p |\ p \in \{0.01, 0.02, ..., 0.99\}\}$ from their marginal distributions.
$\bar{T}^*$, $\bar{t}_g^*$, and $\bar{t}_a^*$ are sets containing corresponding parameter candidates.
The function update\_candidates\_$W$ updates $W$ according to the sub-dataset that $\textbf{V}_j$ belongs to.
(For instance, a $\textbf{V}_j$ from S1 requires that $v_{f,init} > v_{l,init} > 0$.)
The functions update\_candidates\_$T$, update\_candidates\_$t_g$ and update\_candidates\_$t_a$ update $\bar{T}^*$, $\bar{t}_g^*$ and $\bar{t}_a^*$, respectively.
These functions were designed based on the monotonous correlation between the parameter ($T$, $t_g$, or $t_a$) and the crash moment $t_c$.
(For instance, a larger $T$ would result in the vehicle maintaining a longer following distance, delaying any potential crash.)
Therefore, if the current $t_c$ is less than five seconds, the $T$ candidates for the next iteration must be larger than the current value of $T$.
The simulation function sim is described in Section \ref{section:simulationsetting}.

\begin{figure}[!t]
    \centering
    \includegraphics[width=0.4\textwidth]{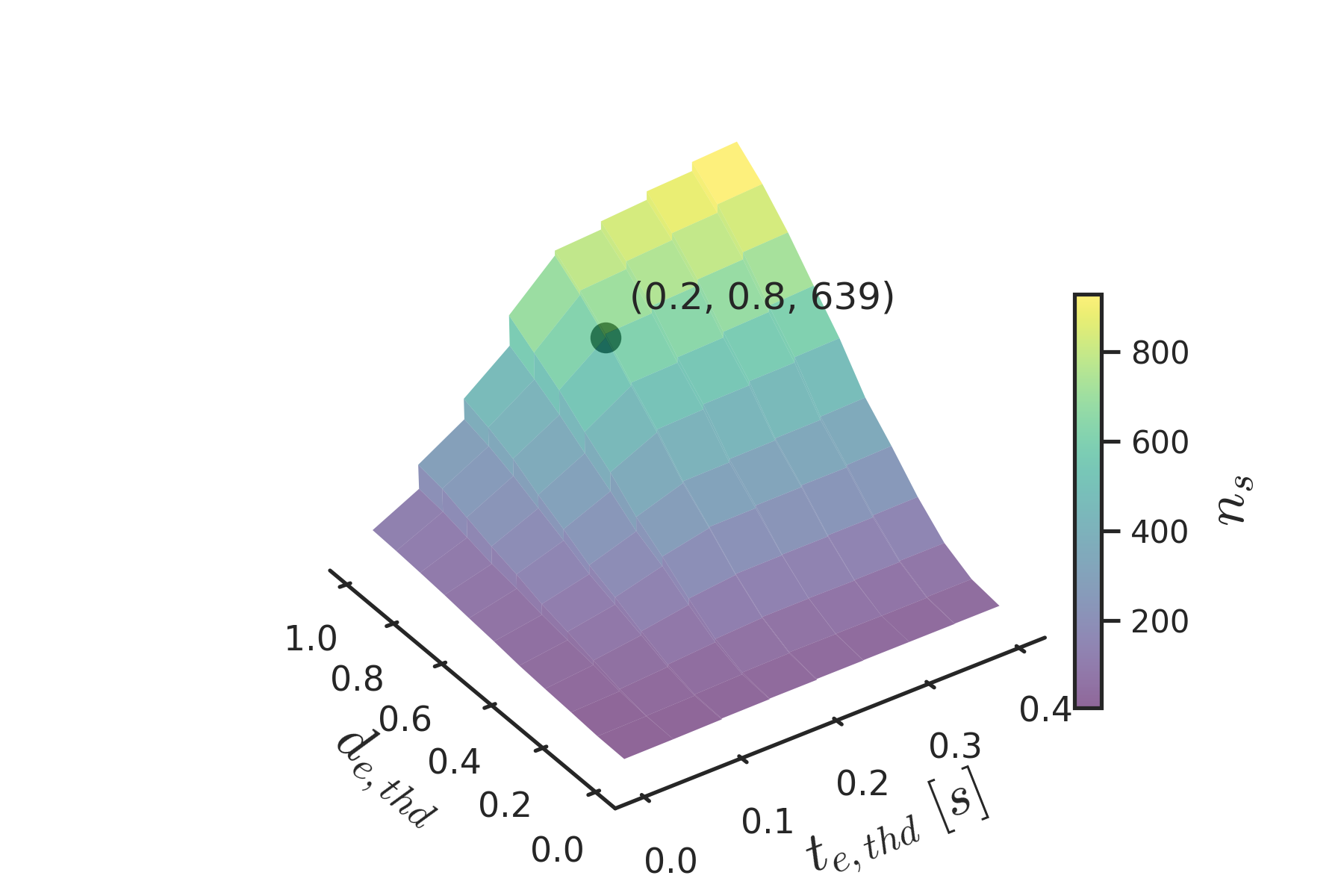}
    \caption{Selection of $t_{e,thd}$ and $d_{e,thd}$.}
    \label{fig:t_e_and_d_e}
\end{figure}

To determine the two threshold values $t_{e,thd}$ and $d_{e,thd}$, a subset with a sample size of 200 was randomly extracted from REF\_sl and REF\_sb, respectively ($U$ and $V$).
For each $\textbf{U}_i \in U$ and each $\textbf{V}_j \in V$, we looped through $\bar{T}^*$ and $\bar{T}_g^*$ to find the event with the minimum crash moment error $t_e$.
The total number of valid simulations $n_s$ is a function of $t_{e,thd}$ and $d_{e,thd}$ (see Fig. \ref{fig:t_e_and_d_e}).
The elbow point in the surface was selected: $d_{e,thd}$ = 1.0, $t_{e,thd}$ = 0.2 s.

\begin{algorithm*}
\caption{Matching algorithm.}\label{alg:alg3}
\begin{algorithmic}
\STATE 
\STATE Sample with the replacement of $N$ samples from REF\_sl and REF\_sb, respectively:
\STATE \hspace{0.5cm} $U = \{\textbf{U}_i |\ i \in [1, N]\}$, where $\textbf{U}_i = [\Tilde{v}_{l,init}^{(i)}, \Tilde{a}_{l,min}^{(i)}, \Tilde{a}_{1}^{(i)}, \Tilde{a}_{2}^{(i)}, \Tilde{\tau}_{s}^{(i)}, \Tilde{\tau}_{1}^{(i)}, \Tilde{\tau}_{2}^{(i)}]^T$
\STATE \hspace{0.5cm} $V = \{\textbf{V}_j |\ j \in [1, N]\}$, where $\textbf{V}_j = [\Tilde{d}_{init}^{(j)}, \Tilde{v}_{f,init}^{(j)}, \Tilde{a}_{f,min}^{(j)}, \Tilde{v}_{l,init}^{(*j)}, \Tilde{a}_{l,min}^{(*j)}]^T$
\STATE For $j = 1$ to $N$:
\STATE \hspace{0.5cm} Set default values: $valid = \text{False}$, $log$ = None, $n_{iter} = 0$
\STATE \hspace{0.5cm} $d_j^{(i)} = \sqrt{(\Tilde{v}_{l,init}^{'(i)} - \Tilde{v}_{l,init}^{'(*j)})^2 + (\Tilde{a}_{l,min}^{'(i)} - \Tilde{a}_{l,min}^{'(*j)})^2}\ \forall\ i \in [1,N]$
\STATE \hspace{0.5cm} $W = \{\textbf{U}_i |\ d_j^{(i)} \leq d_{e,thd}, i \in [1, N]\}$
\STATE \hspace{0.5cm} $W \gets$ update\_candidates\_$W(W)$
\STATE \hspace{0.5cm} While (not $valid$) and ($n(W) > 0$) and ($n_{iter} < n_{iter,max}$):
\STATE \hspace{1cm} $n_{iter} \gets n_{iter} + 1$
\STATE \hspace{1cm} Create candidates for the three parameters: $\bar{T}^*$, $\bar{t}_g^*$, $\bar{t}_a^*$
\STATE \hspace{1cm} Select randomly $\textbf{W}_l$ from $W$ and drop it from $W$
\STATE \hspace{1cm} While (not $valid$) and ($n(\bar{T}) > 0$):
\STATE \hspace{1.5cm} If $log$ is not None:
\STATE \hspace{2cm} $\bar{T}^* \gets$ update\_candidates\_$T$($log, \bar{T}^*$)
\STATE \hspace{2cm} If $n(\bar{T}^*) = 0$: break
\STATE \hspace{1.5cm} Select randomly $T^{(m)}$ from $\bar{T}^*$ and drop it from $\bar{T}^*$
\STATE \hspace{1.5cm} While (not $valid$) and ($n(\bar{t}_g^*) > 0$):
\STATE \hspace{2cm} If $log$ is not None:
\STATE \hspace{2.5cm} $\bar{t}_g^* \gets$ update\_candidates\_$t_g$($log, \bar{t}_g^*$)
\STATE \hspace{2.5cm} If $n(\bar{t}_g^*) = 0$: break
\STATE \hspace{2cm} Select randomly $t_g^{(h)}$ from $\bar{t}_g^*$ and drop it from $\bar{t}_g^*$
\STATE \hspace{2cm} If abnormal\_acceleration($\textbf{V}_j$):
\STATE \hspace{2.5cm} While (not $valid$) and ($n(\bar{t}_a^*) > 0$):
\STATE \hspace{3cm} If $log$ is not None:
\STATE \hspace{3.5cm} $\bar{t}_a^* \gets$ update\_candidates\_$t_a$($log, \bar{t}_a^*$)
\STATE \hspace{3.5cm} If $n(\bar{t}_a^*) = 0$: break
\STATE \hspace{3cm} Select randomly $t_a^{(q)}$ from $\bar{t}_a^*$ and drop it from $\bar{t}_a^*$
\STATE \hspace{3cm} $valid, log = \text{sim}(\textbf{V}_i, \textbf{W}_l, T^{(m)}, t_g^{(h)}, t_a^{(q)})$
\STATE \hspace{2cm} Else:
\STATE \hspace{2.5cm} $valid, log = \text{sim}(\textbf{V}_i, \textbf{W}_l, T^{(m)}, t_g^{(h)}, +\infty)$
\STATE \hspace{0.5cm} If $valid$: Save $log$
\end{algorithmic}
\end{algorithm*}

\bibliographystyle{IEEEtran}
\bibliography{references}

\begin{thebibliography}{10}
\providecommand{\url}[1]{#1}
\csname url@samestyle\endcsname
\providecommand{\newblock}{\relax}
\providecommand{\bibinfo}[2]{#2}
\providecommand{\BIBentrySTDinterwordspacing}{\spaceskip=0pt\relax}
\providecommand{\BIBentryALTinterwordstretchfactor}{4}
\providecommand{\BIBentryALTinterwordspacing}{\spaceskip=\fontdimen2\font plus
\BIBentryALTinterwordstretchfactor\fontdimen3\font minus
  \fontdimen4\font\relax}
\providecommand{\BIBforeignlanguage}[2]{{%
\expandafter\ifx\csname l@#1\endcsname\relax
\typeout{** WARNING: IEEEtran.bst: No hyphenation pattern has been}%
\typeout{** loaded for the language `#1'. Using the pattern for}%
\typeout{** the default language instead.}%
\else
\language=\csname l@#1\endcsname
\fi
#2}}
\providecommand{\BIBdecl}{\relax}
\BIBdecl

\bibitem{pradhan2022impact}
A.~K. Pradhan, A.~Hungund, D.~E. Sullivan \emph{et~al.}, ``Impact of advanced
  driver assistance systems (adas) on road safety and implications for
  education, licensing, registration, and enforcement,'' Massachusetts. Dept.
  of Transportation. Office of Transportation Planning, Tech. Rep., 2022.

\bibitem{feng2020safety}
S.~Feng, Y.~Feng, X.~Yan, S.~Shen, S.~Xu, and H.~X. Liu, ``Safety assessment of
  highly automated driving systems in test tracks: A new framework,''
  \emph{Accident Analysis \& Prevention}, vol. 144, p. 105664, 2020.

\bibitem{dona2022virtual}
R.~Don{\`a} and B.~Ciuffo, ``Virtual testing of automated driving systems. a
  survey on validation methods,'' \emph{IEEE Access}, vol.~10, pp.
  24\,349--24\,367, 2022.

\bibitem{cai2022survey}
J.~Cai, W.~Deng, H.~Guang, Y.~Wang, J.~Li, and J.~Ding, ``A survey on
  data-driven scenario generation for automated vehicle testing,''
  \emph{Machines}, vol.~10, no.~11, p. 1101, 2022.

\bibitem{szalay2023critical}
Z.~Szalay, ``Critical scenario identification concept: the role of the
  scenario-in-the-loop approach in future automotive testing,'' \emph{IEEE
  Access}, 2023.

\bibitem{wimmer2023harmonized}
P.~Wimmer, O.~Op\_Den\_Camp, H.~Weber, H.~Chajmowicz, M.~Wagner, J.~L. Mallada,
  F.~Fahrenkrog, and F.~Denk, ``Harmonized approaches for baseline creation in
  prospective safety performance assessment of driving automation systems,'' in
  \emph{27th International Technical Conference on the Enhanced Safety of
  Vehicles (ESV), Yokohama, Japan}, 2023, pp. 3--6.

\bibitem{feng2021intelligent}
S.~Feng, X.~Yan, H.~Sun, Y.~Feng, and H.~X. Liu, ``Intelligent driving
  intelligence test for autonomous vehicles with naturalistic and adversarial
  environment,'' \emph{Nature communications}, vol.~12, no.~1, p. 748, 2021.

\bibitem{baron2020repeatable}
W.~Baron, C.~Sippl, K.-S. Hielscher, and R.~German, ``Repeatable simulation for
  highly automated driving development and testing,'' in \emph{2020 IEEE 91st
  Vehicular Technology Conference (VTC2020-Spring)}.\hskip 1em plus 0.5em minus
  0.4em\relax IEEE, 2020, pp. 1--7.

\bibitem{shah2018airsim}
S.~Shah, D.~Dey, C.~Lovett, and A.~Kapoor, ``Airsim: High-fidelity visual and
  physical simulation for autonomous vehicles,'' in \emph{Field and Service
  Robotics: Results of the 11th International Conference}.\hskip 1em plus 0.5em
  minus 0.4em\relax Springer, 2018, pp. 621--635.

\bibitem{scanlon2021waymo}
J.~M. Scanlon, K.~D. Kusano, T.~Daniel, C.~Alderson, A.~Ogle, and T.~Victor,
  ``Waymo simulated driving behavior in reconstructed fatal crashes within an
  autonomous vehicle operating domain,'' \emph{Accident Analysis \&
  Prevention}, vol. 163, p. 106454, 2021.

\bibitem{bareiss2019crash}
M.~Bareiss, J.~Scanlon, R.~Sherony, and H.~C. Gabler, ``Crash and injury
  prevention estimates for intersection driver assistance systems in left turn
  across path/opposite direction crashes in the united states,'' \emph{Traffic
  injury prevention}, vol.~20, no. sup1, pp. S133--S138, 2019.

\bibitem{hamdane2015issues}
H.~Hamdane, T.~Serre, C.~Masson, and R.~Anderson, ``Issues and challenges for
  pedestrian active safety systems based on real world accidents,''
  \emph{Accident Analysis \& Prevention}, vol.~82, pp. 53--60, 2015.

\bibitem{haus2019potential}
S.~H. Haus and H.~C. Gabler, ``The potential for active safety mitigation of us
  vehicle-bicycle crashes,'' \emph{Future Active Safety Technology Towards Zero
  Traffic Accidents (FAST-Zero)}, 2019.

\bibitem{bargman2017counterfactual}
J.~B{\"a}rgman, C.-N. Boda, and M.~Dozza, ``Counterfactual simulations applied
  to shrp2 crashes: The effect of driver behavior models on safety benefit
  estimations of intelligent safety systems,'' \emph{Accident Analysis \&
  Prevention}, vol. 102, pp. 165--180, 2017.

\bibitem{ma2022verification}
Y.~Ma, C.~Sun, J.~Chen, D.~Cao, and L.~Xiong, ``Verification and validation
  methods for decision-making and planning of automated vehicles: A review,''
  \emph{IEEE Transactions on Intelligent Vehicles}, vol.~7, no.~3, pp.
  480--498, 2022.

\bibitem{olleja2022can}
P.~Olleja, J.~B{\"a}rgman, and N.~Lubbe, ``Can non-crash naturalistic driving
  data be an alternative to crash data for use in virtual assessment of the
  safety performance of automated emergency braking systems?'' \emph{Journal of
  safety research}, vol.~83, pp. 139--151, 2022.

\bibitem{arun2021systematic}
A.~Arun, M.~M. Haque, A.~Bhaskar, S.~Washington, and T.~Sayed, ``A systematic
  mapping review of surrogate safety assessment using traffic conflict
  techniques,'' \emph{Accident Analysis \& Prevention}, vol. 153, p. 106016,
  2021.

\bibitem{leledakis2021method}
A.~Leledakis, M.~Lindman, J.~{\"O}sth, L.~W{\aa}gstr{\"o}m, J.~Davidsson, and
  L.~Jakobsson, ``A method for predicting crash configurations using
  counterfactual simulations and real-world data,'' \emph{Accident Analysis \&
  Prevention}, vol. 150, p. 105932, 2021.

\bibitem{gambi2019generating}
A.~Gambi, T.~Huynh, and G.~Fraser, ``Generating effective test cases for
  self-driving cars from police reports,'' in \emph{Proceedings of the 2019
  27th ACM Joint Meeting on European Software Engineering Conference and
  Symposium on the Foundations of Software Engineering}, 2019, pp. 257--267.

\bibitem{wang2022autonomous}
X.~Wang, Y.~Peng, T.~Xu, Q.~Xu, X.~Wu, G.~Xiang, S.~Yi, and H.~Wang,
  ``Autonomous driving testing scenario generation based on in-depth
  vehicle-to-powered two-wheeler crash data in china,'' \emph{Accident Analysis
  \& Prevention}, vol. 176, p. 106812, 2022.

\bibitem{wu2024modeling}
J.~Wu, C.~Flannagan, U.~Sander, and J.~B{\"a}rgman, ``Modeling lead-vehicle
  kinematics for rear-end crash scenario generation,'' \emph{IEEE Transactions
  on Intelligent Transportation Systems}, 2024.

\bibitem{derbel2013modified}
O.~Derbel, T.~Peter, H.~Zebiri, B.~Mourllion, and M.~Basset, ``Modified
  intelligent driver model for driver safety and traffic stability
  improvement,'' \emph{IFAC Proceedings Volumes}, vol.~46, no.~21, pp.
  744--749, 2013.

\bibitem{svard2021computational}
M.~Sv{\"a}rd, G.~Markkula, J.~B{\"a}rgman, and T.~Victor, ``Computational
  modeling of driver pre-crash brake response, with and without off-road
  glances: Parameterization using real-world crashes and near-crashes,''
  \emph{Accident Analysis \& Prevention}, vol. 163, p. 106433, 2021.

\bibitem{zhang2019crash}
F.~Zhang, E.~Y. Noh, R.~Subramanian, and C.-L. Chen, ``Crash investigation
  sampling system: Sample design and weighting,'' Tech. Rep., 2019.

\bibitem{subramanian2020crash}
R.~Subramanian and E.~Acevedo-D{\'\i}az, ``Crash investigation sampling system
  2019 data manual,'' Tech. Rep., 2020.

\bibitem{hankey2016description}
J.~M. Hankey, M.~A. Perez, and J.~A. McClafferty, ``Description of the shrp 2
  naturalistic database and the crash, near-crash, and baseline data sets,''
  Virginia Tech Transportation Institute, Tech. Rep., 2016.

\bibitem{schubert2017gidas}
A.~Schubert, H.~Liers, and M.~Petzold, ``The gidas pre-crash-matrix 2016:
  Innovations for standardized pre-crash-scenarios on the basis of the vufo
  simulation model vast,'' 2017.

\bibitem{victor2015analysis}
T.~Victor, M.~Dozza, J.~B{\"a}rgman, C.-N. Boda, J.~Engstr{\"o}m, C.~Flannagan,
  J.~D. Lee, and G.~Markkula, ``Analysis of naturalistic driving study data:
  Safer glances, driver inattention, and crash risk,'' Tech. Rep., 2015.

\bibitem{svard2017quantitative}
M.~Sv{\"a}rd, G.~Markkula, J.~Engstr{\"o}m, F.~Granum, and J.~B{\"a}rgman, ``A
  quantitative driver model of pre-crash brake onset and control,'' in
  \emph{Proceedings of the Human Factors and Ergonomics Society Annual
  Meeting}, vol.~61, no.~1.\hskip 1em plus 0.5em minus 0.4em\relax SAGE
  Publications Sage CA: Los Angeles, CA, 2017, pp. 339--343.

\bibitem{lee1976theory}
D.~N. Lee, ``A theory of visual control of braking based on information about
  time-to-collision,'' \emph{Perception}, vol.~5, no.~4, pp. 437--459, 1976.

\bibitem{bargman2015does}
J.~B{\"a}rgman, V.~Lisovskaja, T.~Victor, C.~Flannagan, and M.~Dozza, ``How
  does glance behavior influence crash and injury risk? a
  ‘what-if’counterfactual simulation using crashes and near-crashes from
  shrp2,'' \emph{Transportation Research Part F: Traffic Psychology and
  Behaviour}, vol.~35, pp. 152--169, 2015.

\bibitem{markkula2016farewell}
G.~Markkula, J.~Engstr{\"o}m, J.~Lodin, J.~B{\"a}rgman, and T.~Victor, ``A
  farewell to brake reaction times? kinematics-dependent brake response in
  naturalistic rear-end emergencies,'' \emph{Accident Analysis \& Prevention},
  vol.~95, pp. 209--226, 2016.

\bibitem{cohen2009pearson}
I.~Cohen, Y.~Huang, J.~Chen, J.~Benesty, J.~Benesty, J.~Chen, Y.~Huang, and
  I.~Cohen, ``Pearson correlation coefficient,'' \emph{Noise reduction in
  speech processing}, pp. 1--4, 2009.

\bibitem{weightsrpackage}
\BIBentryALTinterwordspacing
J.~Pasek, A.~Tahk, G.~Culter, and M.~Schwemmle, \emph{weights: Weighting and
  Weighted Statistics}, 2021, r package version 1.0.4. [Online]. Available:
  \url{https://CRAN.R-project.org/package=weights}
\BIBentrySTDinterwordspacing

\bibitem{choupani2016population}
A.-A. Choupani and A.~R. Mamdoohi, ``Population synthesis using iterative
  proportional fitting (ipf): A review and future research,''
  \emph{Transportation Research Procedia}, vol.~17, pp. 223--233, 2016.

\bibitem{ecumerpackage}
\BIBentryALTinterwordspacing
H.~{Roux de Bezieux}, \emph{Ecume: Equality of 2 (or k) Continuous Univariate
  and Multivariate Distributions}, 2021, r package version 0.9.1. [Online].
  Available: \url{https://CRAN.R-project.org/package=Ecume}
\BIBentrySTDinterwordspacing

\bibitem{van2008visualizing}
L.~Van~der Maaten and G.~Hinton, ``Visualizing data using t-sne.''
  \emph{Journal of machine learning research}, vol.~9, no.~11, 2008.

\bibitem{kudlich1966beitrag}
H.~Kudlich, \emph{Beitrag zur Mechanik des
  Kraftfahrzeug-Verkehrsunfalls}.\hskip 1em plus 0.5em minus 0.4em\relax na,
  1966.

\bibitem{leifer2013supplemental}
J.~Leifer, ``A supplemental analysis of selected two-vehicle front-to-rear
  collisions from the nass/cds,'' \emph{Society of Automotive Engineering
  (SAE)}, 2013.

\bibitem{holt1979post}
D.~Holt and T.~F. Smith, ``Post stratification,'' \emph{Journal of the Royal
  Statistical Society Series A: Statistics in Society}, vol. 142, no.~1, pp.
  33--46, 1979.

\end{thebibliography}
\begin{IEEEbiography}[{\includegraphics[width=1in,height=1.25in,clip,keepaspectratio]{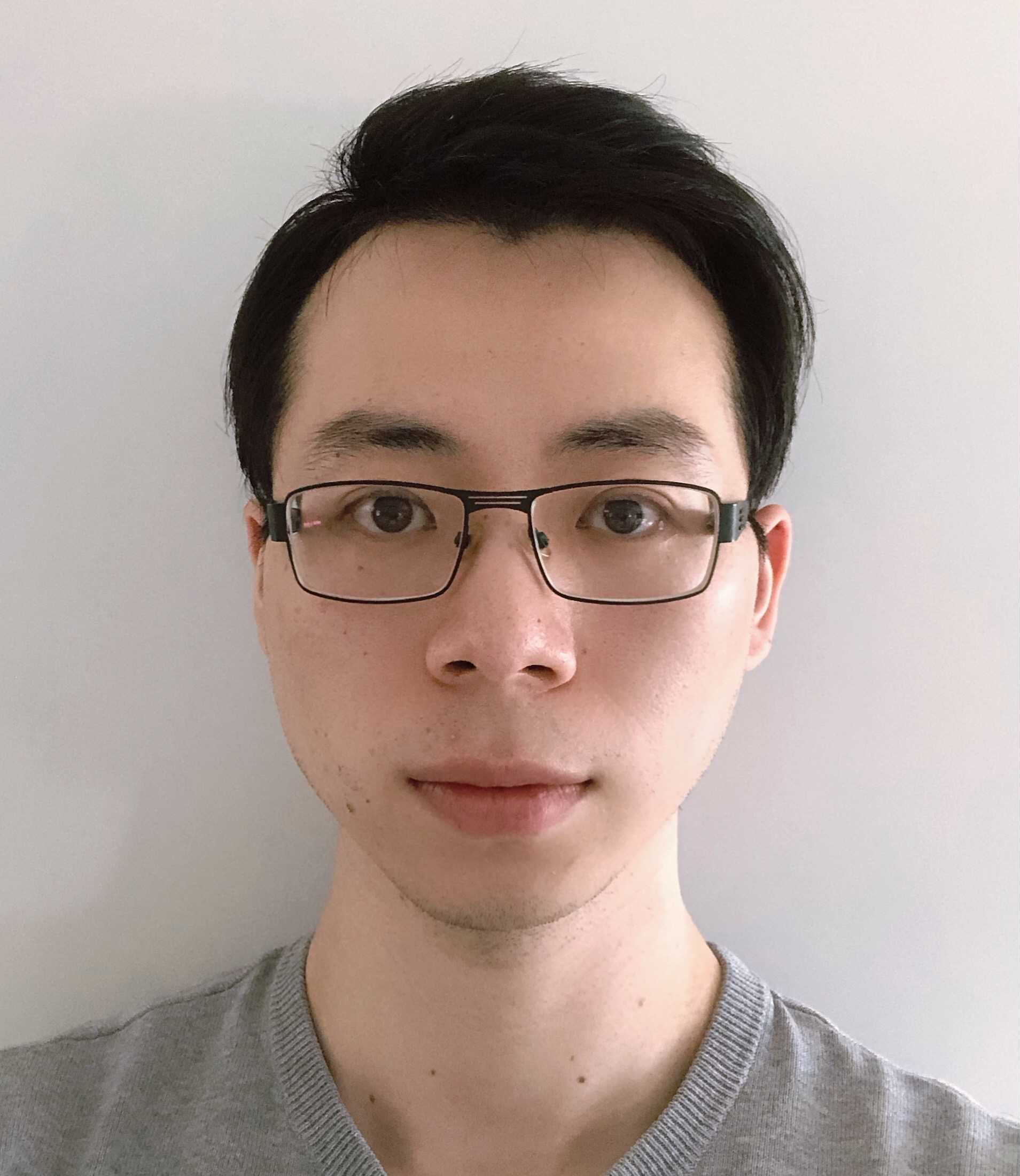}}]{Jian Wu}
received his B.S. and M.S. degrees in automotive engineering from Tsinghua University, Beijing, China, in 2013 and 2016. He is now an industrial Ph.D. candidate with Volvo Cars Safety Center and the Department of Mechanics and Maritime Sciences, Chalmers University of Technology, Göteborg, Sweden. He is the author or coauthor of four journal papers and two conference papers. His current research interests include driver behavior modeling, crash data synthesis, and the safety assessments of vehicle safety technologies.
\end{IEEEbiography}

\begin{IEEEbiography}[{\includegraphics[width=1in,height=1.25in,clip,keepaspectratio]{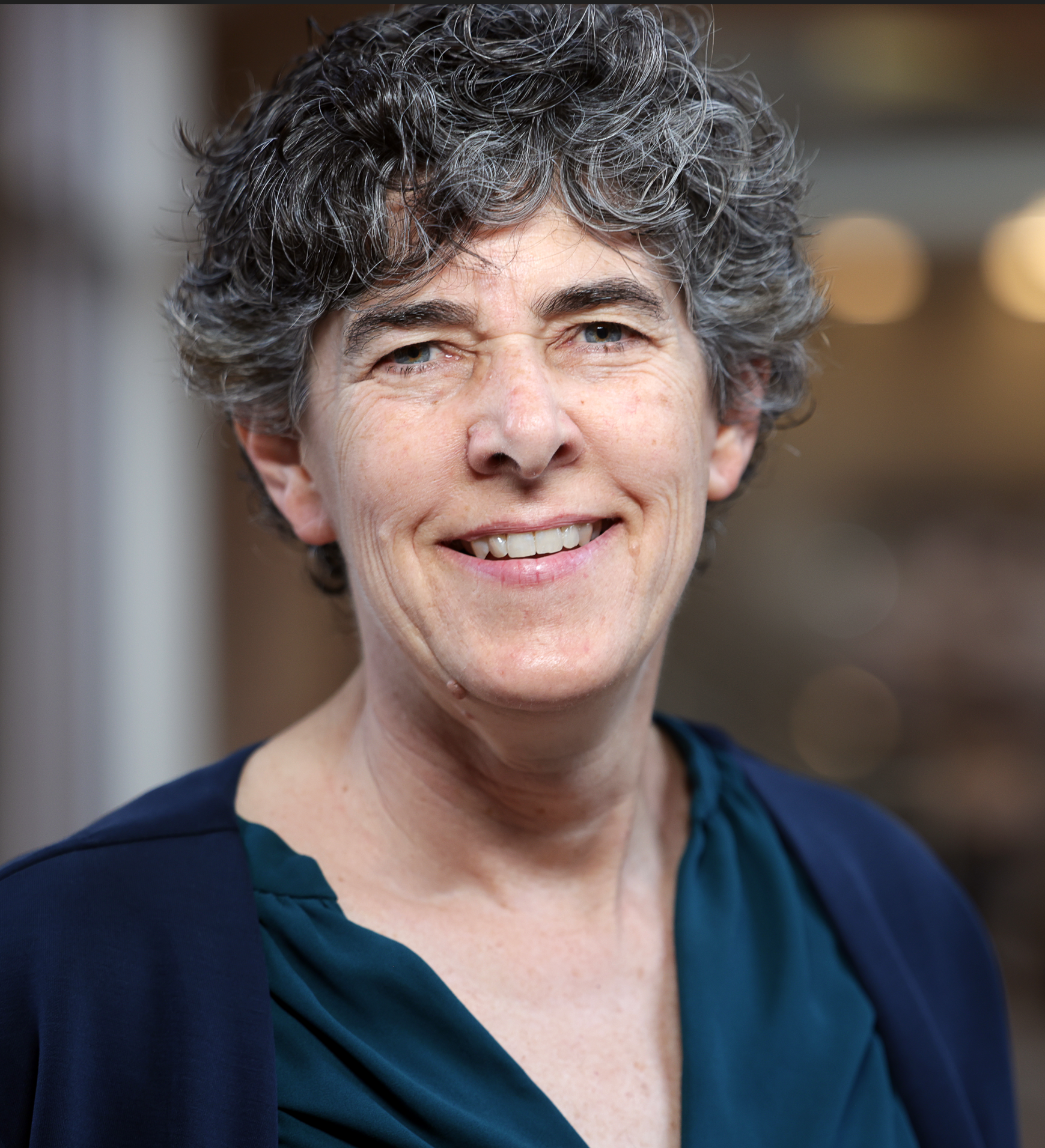}}]{Carol Flannagan}
received an M.A. in Statistics and a Ph.D. in Mathematical Psychology from the University of Michigan. She is a Research Professor at the University of Michigan Transportation Research Institute (UMTRI) in Ann Arbor, Michigan, USA, and an Affiliated Associate Professor at Chalmers University, Göteborg, Sweden. Her work in transportation research encompasses the analysis of a wide variety of transportation-related data and the development of innovative statistical methods for transportation research. She is currently working on a number of projects related to safety assessment and benefits assessment for advanced technologies, including ADS.
\end{IEEEbiography}

\begin{IEEEbiography}[{\includegraphics[width=1in,height=1.25in,clip,keepaspectratio]{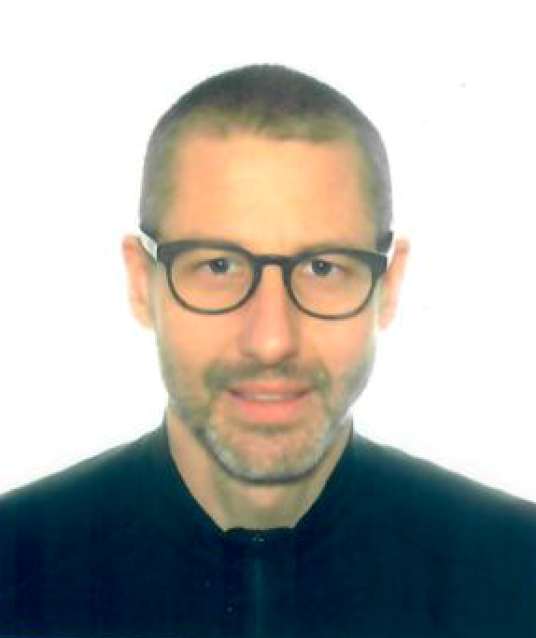}}]{Ulrich Scander}
received his B.S. degree in Biomedical Engineering from the University of Aachen, Germany, in 1997 and his M.S. degree in Accident Research from Graz University of Technology, Austria, in 2008. He received his Ph.D. in Machine and Vehicle Systems from Chalmers University of Technology, Gothenburg, Sweden. He has worked for over 20 years in different positions, such as data analyst and senior principal researcher at Autoliv Research in Germany and Sweden. Since 2022, he has been a Technical Expert at the Safety Centre of Volvo Cars, leading the analysis of field data with a focus on crashes and their consequences.
\end{IEEEbiography}

\begin{IEEEbiography}[{\includegraphics[width=1in,height=1.25in,clip,keepaspectratio]{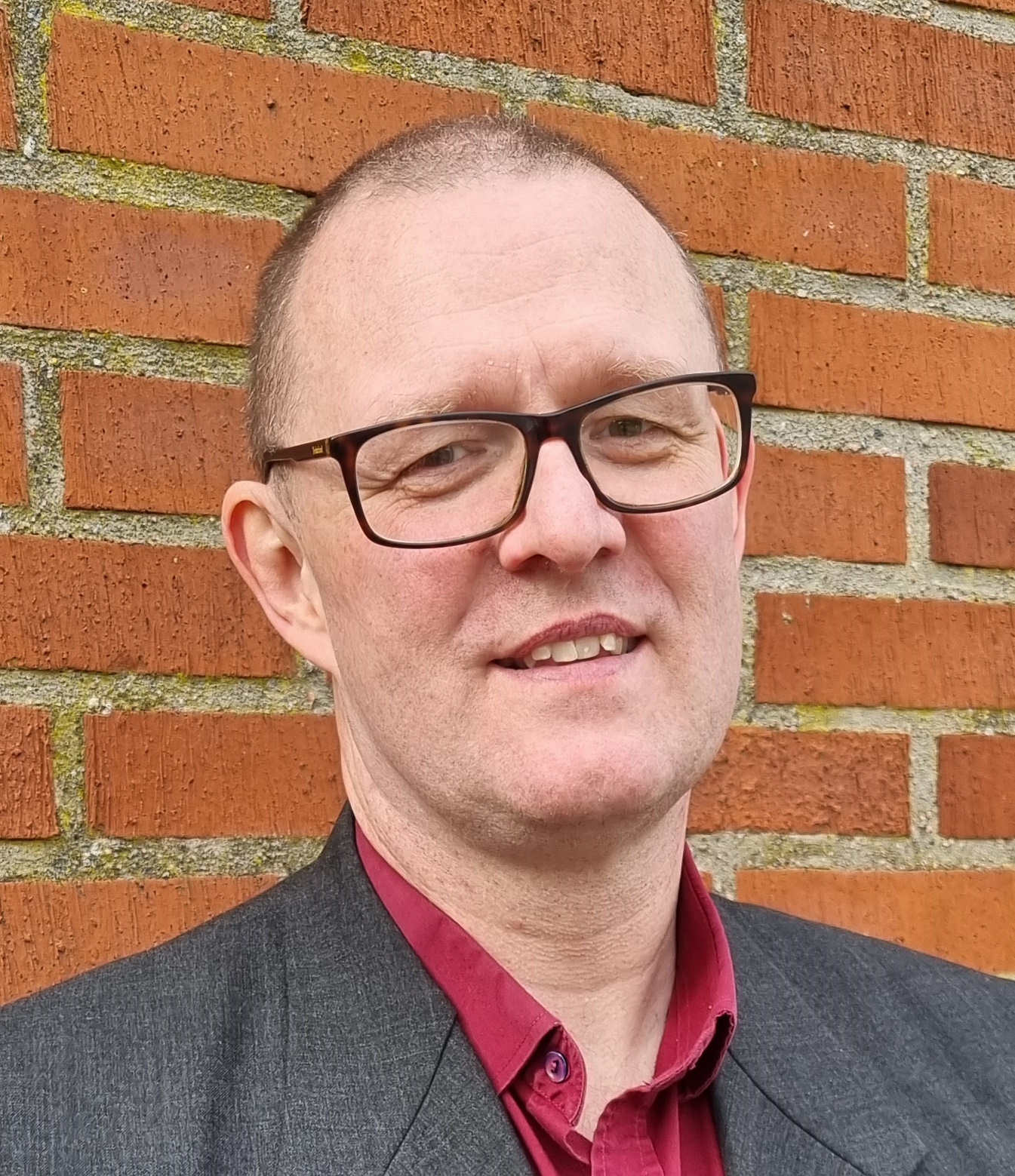}}]{Jonas Bärgman}
received his M.Sc. degree in Mechanical Engineering at Chalmers University of Technology, Gothenburg, Sweden in 1997. After his degree, he worked as an industry researcher in in-crash safety at Autoliv Research for three years and as a software developer at AB Volvo for two years. At this point, he continued his career at Autoliv Research (again) in the domain of pre-crash safety, focusing on human factors and driver behavior. In 2009, he started working at Chalmers University of Technology to build a research group on Active Safety. He received his Ph.D. in Machine and Vehicle Systems from the same university in 2016 and is currently a Professor. He is also the examiner for the course “Vehicle and Traffic Safety” in the master’s degree program in Mobility Engineering. His main research interests are virtual safety assessment and its components, including driver behavior modeling, scenario generation, and statistical methods.
\end{IEEEbiography}


\vfill

\end{document}